%% file: paper.tex
\documentclass[lettersize,journal]{IEEEtran}
\usepackage{amsmath,amsfonts,amssymb}
\usepackage{algorithmic}
\usepackage{algorithm}
\usepackage{array}
\usepackage{textcomp}
\usepackage{stfloats}
\usepackage{url}
\usepackage{verbatim}
\usepackage{graphicx}
\usepackage{cite}
\usepackage{setspace}
\usepackage{caption}
\usepackage{subcaption}
\usepackage{etoolbox}
\usepackage{tabularx}
\usepackage{url}
\usepackage{titlesec}
\usepackage{adjustbox}
\usepackage{tikz}
\usepackage{comment}
\usepackage{colortbl}
\usepackage{color}
\usepackage{hyperref}
\usepackage{multirow}
\usepackage{booktabs}
\usepackage{xcolor}
\usepackage{pifont}
\usepackage{bbding}
\usepackage{fontawesome}
\usepackage{xspace}
\usepackage{float}
\usepackage{listings}
\usepackage{soul}
\usepackage{tcolorbox}
\usepackage{diagbox}
\usepackage{makecell}
\usepackage{rotating}
\usepackage{enumitem}
\usepackage{tablefootnote}
\usepackage{footnote}
\usepackage[numbers]{natbib}

\RequirePackage{xspace}
\makeatletter
\DeclareRobustCommand\onedot{\futurelet\@let@token\@onedot}
\def\@onedot{\ifx\@let@token.\else.\null\fi\xspace}

\newtcolorbox{promptblock}{
    colback=gray!5,
    colframe=gray!15,
    boxrule=0.5pt,
    arc=3pt,
    left=12pt,
    right=12pt,
    top=8pt,
    bottom=8pt,
    boxsep=8pt,
    breakable
}

\def\eg{\emph{e.g}\onedot} 
\def\ie{\emph{i.e}\onedot} 
\def\etal{\emph{et al.}\xspace}
\newcommand{\ours}{Lingshu-Med\xspace}
\newcommand{\oureval}{MedEvalKit\xspace}

\newcommand{\teaserfigure}{%
  \setcounter{figure}{0}
  \centering
  \includegraphics[width=\linewidth]{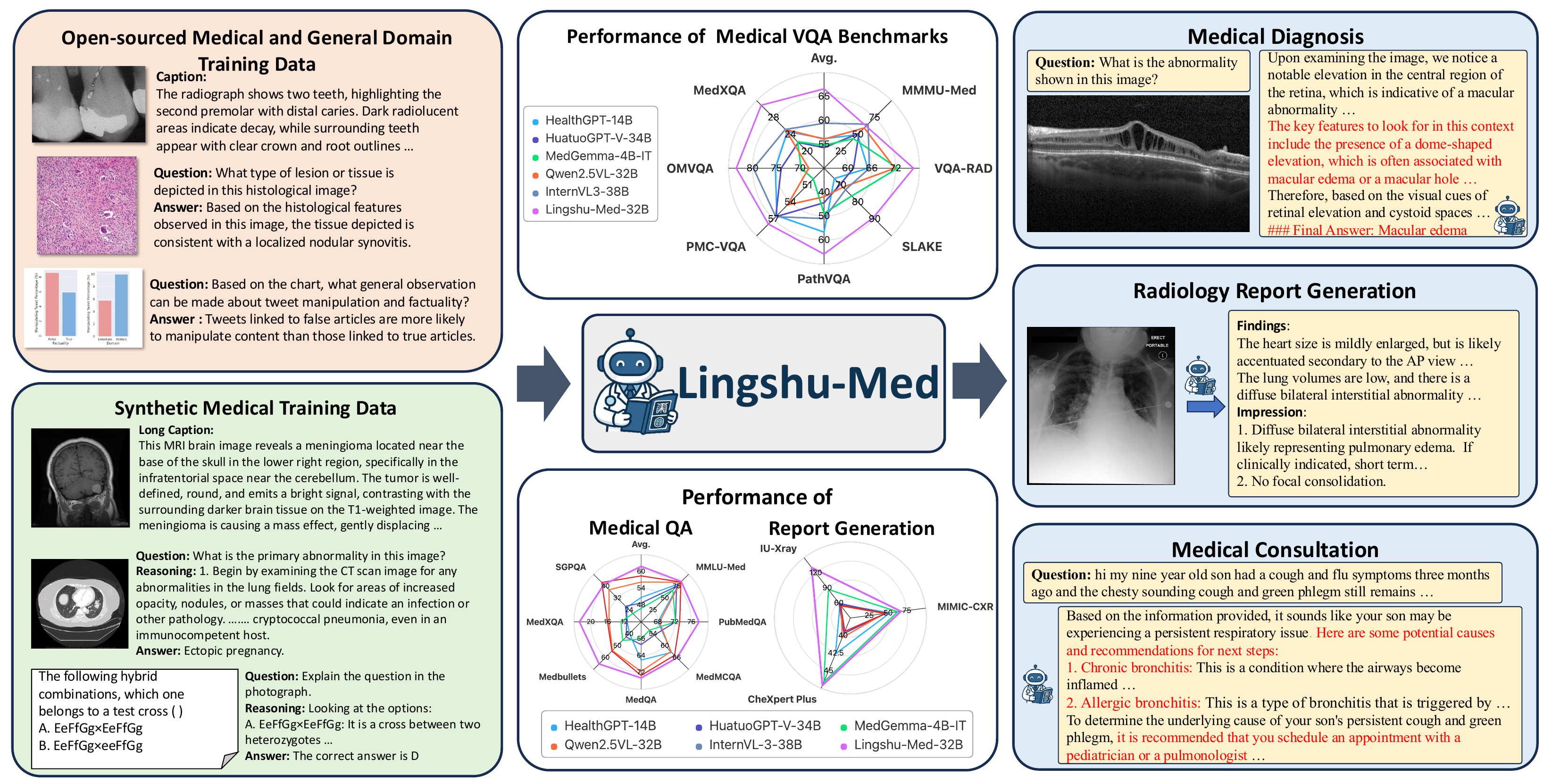}
  \captionof{figure}{\ours is trained on meticulously curated datasets, encompassing rich and diverse medical data across different modalities, general domain, and high-quality synthetic medical data. Our results demonstrate that \ours achieves promising performance across various medical benchmarks and modalities, with significant potential in real-world healthcare settings.}
  \label{fig:lingshu-overview}
}

\begin{document}

\makeatletter
\apptocmd{\@maketitle}{\vspace{0.5\baselineskip}\teaserfigure}{}{}
\makeatother

\title{Lingshu: Generalist Foundation Model for Unified Multimodal Medical Understanding and Reasoning}

\author{Weiwen Xu, Hou Pong Chan, Long Li, Mahani Aljunied, Ruifeng Yuan, Jianyu Wang, Chenghao Xiao, \\Guizhen Chen, Chaoqun Liu, Zhaodonghui Li, Yu Sun, Junao Shen, Chaojun Wang, Can Hu, Siwei Pan,\\
Jie Tan, Tingyang Xu, Deli Zhao, Hao Zhang, and Yu Rong
        % <-this % stops a space
\thanks{W.~Xu, L.~Li, M.~Aljunied, R.~Yuan, J.~Wang, G.~Chen, C.~Liu, Z.~Li, Y.~Sun, J.~Shen, C.~Wang, J.~Tan, D.~Zhao, T.~Xu, H.~Zhang, and Y.~Rong are with DAMO Academy, Alibaba Group.}% <-this % stops a space
\thanks{H.P.~Chan is with the Department of Artificial Intelligence, University of Macau.}
\thanks{C.~Xiao is with Shanghai University of Finance and Economics.}
\thanks{C.~Hu and S.~Pan are with the Department of Gastric Surgery, Zhejiang Cancer Hospital and Hangzhou Institute of Medicine (HIM), Chinese Academy of Sciences.}
\thanks{Equal contributions: W.~Xu, H.P.~Chan and L.~Li.}
\thanks{Corresponding authors: H. Zhang (\texttt{hzhang26@outlook.com}) and Y. Rong (\texttt{yu.rong@hotmail.com}).}
}

\maketitle
\bstctlcite{IEEEtran:BSTcontrol}


% \input{sections/0_pami_rebuttal_checklist}

\input{sections/0_abstract}

\input{sections/1_keywords}

\input{sections/2_introduction}

\input{sections/3_data}

\input{sections/4_training}

\input{sections/5_evaluation}

\input{sections/6_experiments}

\input{sections/7_related_work}

\input{sections/8_conclusion}

\bibliographystyle{IEEEtran}
\bibliography{paper}

\clearpage
\appendix
% ================================
\section{Appendix}
\subsection{More Details about Data Curation}
\label{apdx:sec:data_curation}
% ================================

% ================================
\subsubsection{The collected medical datasets}\label{apdx:ssec:collected_medical_dataset}
% ================================
The complete list of the collected datasets for medical multimodal data is summarized in Table~\ref{tab:data_set_source_multi_modal}. The complete lists of the collected datasets for medical textual data and medical imaging data are provided in Table~\ref{tab:data_set_source_text} and Table~\ref{tab:data_set_source_med_image}, respectively, where the medical imaging data spans a wide range of medical modalities.

\subsubsection{Details of data cleaning for open-sourced medical datasets}
\label{apdx:ssec:data_cleaning_collected_medical_dataset}

\textbf{Data cleaning for medical multimodal data.} 
First, \textbf{image filtering}, where images with dimensions smaller than $64$ pixels are removed to eliminate low-resolution or uninformative content. Second, \textbf{image deduplication}, which applies perceptual hashing with a strict Hamming distance threshold of zero to detect and remove exact duplicate images, retaining only one high-quality instance per set. To accelerate this process, we use a chunk-based deduplication technique.\footnote{\url{https://github.com/xuehuachunsheng/DupImageDetection}} Third, \textbf{text filtering}, where image caption samples with fewer than 10 or more than 1024 tokens are excluded to filter out lengthy and potentially noisy examples. The full pipeline is applied to all medical image captioning datasets. For instruction data, we omit the text filtering step due to the validity of concise responses, \eg, ``yes/no'', word phrases, or short options. Image deduplication in instruction datasets specifically targets PubMedVision~\citep{chen2024huatuogptvision} and LLaVA-Med~\citep{li2023llavamed}, which exhibit a high degree of redundancy. However, cross-dataset deduplication is avoided in this case, as individual images may be paired with diverse instructions. In contrast, image deduplication across all datasets is conducted for medical captioning data.

\textbf{Data cleaning for medical text data.}
First, we apply \textbf{model-based response cleaning} to open-source patient-doctor dialogue datasets, such as HealthCareMagic~\citep{li2023chatdoctor} and iCliniq~\citep{li2023chatdoctor}, which are typically scraped from online healthcare platforms. These responses often contain sensitive identity information or provide explicit diagnoses and prescriptions, raising privacy and legal concerns. Specifically, we employ LLaMA-3.1-70B~\citep{DBLP:journals/corr/llama3} to remove identity-related content and revise responses to avoid direct medical recommendations. Second, to eliminate redundancy, we perform \textbf{text deduplication} across multiple instruction datasets using min-hash locality-sensitive hashing (LSH), retaining only the highest-quality version of each near-duplicate based on the reliability of its source.

% ================================
\subsubsection{Model-based response cleaning for patient-doctor dialogue datasets}\label{apdx:model-based-response-cleaning}
% ================================
We show the prompt for model-based response cleaning for patient-doctor dialogue datasets in Table~\ref{tab:prompt_doctor_patient_dialog_cleaning}.

\input{tables/appdx_medical_multimodal_data_collection}

\input{tables/appdx_medical_textual_data_collection}

\input{tables/appdx_medical_imaging_data_collection}

\input{tables/appdx_patient_doctor_dialogue_clean}

% ================================
\subsubsection{Medical VQA Data Synthesis}\label{apdx:med-vqa-data-synthesis}
\begin{figure*}[]
    \centering
    \includegraphics[width=0.8\textwidth]{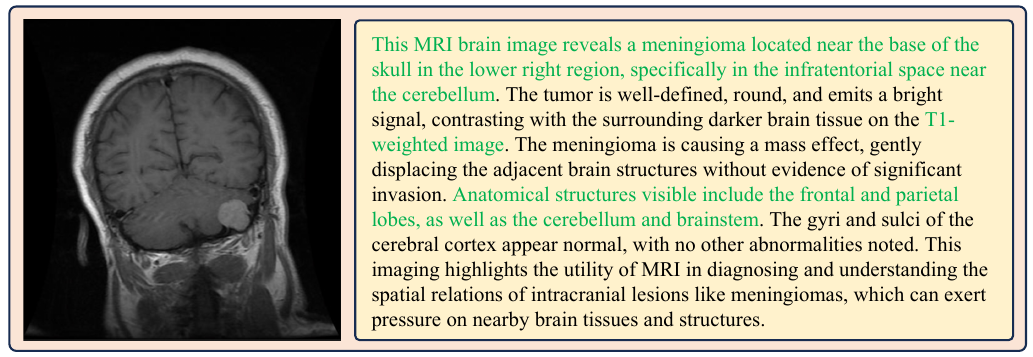}
    \caption{An example of the synthesized medical long-form caption for the MRI image.}
    \label{fig:caption}
\end{figure*}

\begin{figure*}[]
    \centering
    \includegraphics[width=0.9\textwidth]{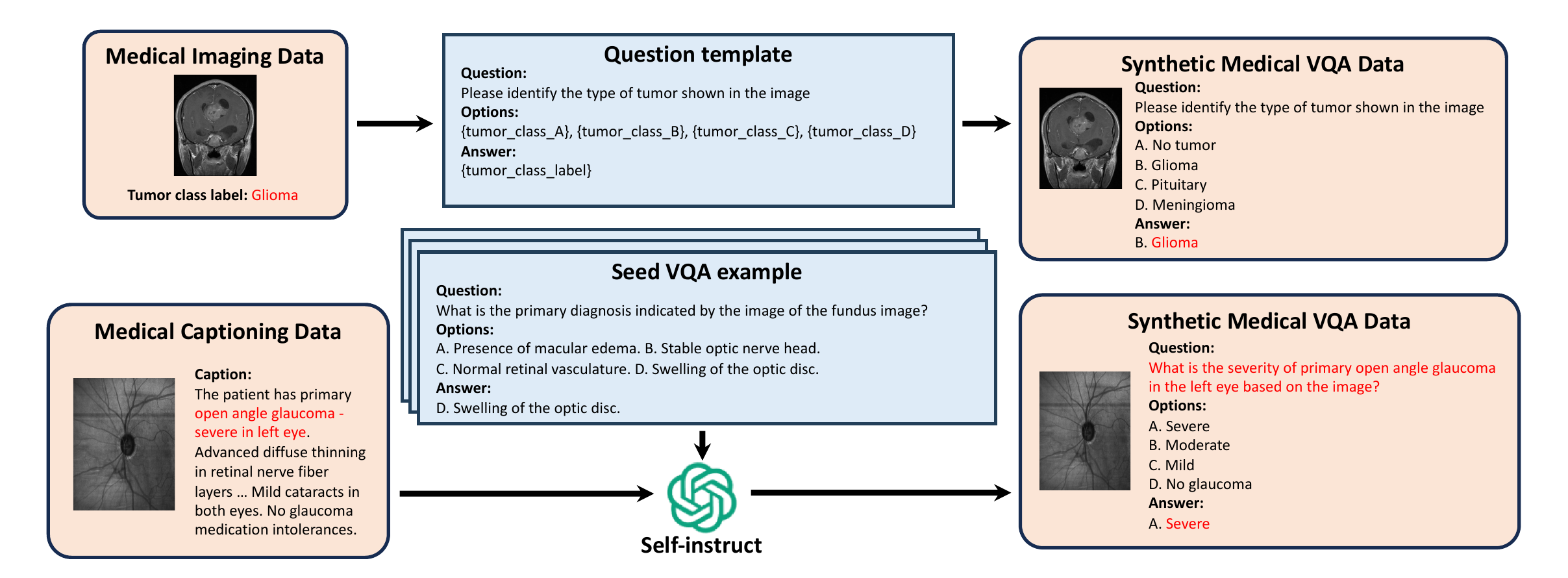}
    \caption{Illustration of the data synthesis process for synthetic medical VQA data.}
    \label{fig:synthetic-sft-illustration}
\end{figure*}

% ================================
The data synthesis process for synthetic medical VQA data is depicted in Fig.~\ref{fig:synthetic-sft-illustration}, where we illustrate both template-based and self-instruct-based methods. Fig.~\ref{fig:caption} presents a representative example of a synthesized long-form caption for an MRI image. Table~\ref{tbl:ocr-example} illustrates an example that a question is rendered as images to form multimodal instruction data, where the question image serves as input and the textual answer as output. We also display the prompt of our self-instruct based method for medical VQA data synthesis in Table~\ref{tab:prompt-self-instruct-vqa-synthesis}.

\paragraph{Further details of template-based method.}
Upon inspecting the medical imaging datasets included in our corpus, we found that their human-annotated meta-information can be largely grouped into four recurring label categories: \textbf{(1) abnormality type}, \textbf{(2) tumor type}, \textbf{(3) anatomical structure}, and \textbf{(4) imaging modality}. We therefore scale question templating by designing templates at the \textbf{label-category level}, rather than crafting dataset-specific templates. Concretely, for each label category we manually write a single seed template, and then use GPT-4o to generate 12 paraphrases that preserve the original semantics while varying wording and syntax. This yields 13 templates per category and 52 templates total (4 categories × 13 templates), which can be instantiated with the dataset-specific label values (\eg, particular tumor types, anatomy terms, or modalities) across the full corpus.

\input{tables/appdx_OCR_based_instruction_example}

\input{tables/appdx_self_instruct_prompt}

% ================================
\subsubsection{Medical reasoning data distillation}\label{apdx:reasoning-distillation}
% ================================
We present the prompts used to distill chain-of-thought (CoT) reasoning from GPT-4o for medical open-ended VQA and medical multiple-choice VQA datasets in Figure~\ref{tab:cot-annotation-qa} and Figure~\ref{tab:cot-annotation-mcq}, respectively. To mitigate the risk that the generated CoT degenerates into post-hoc rationalization rather than faithful clinical reasoning, the distillation prompt explicitly instructs GPT-4o to assume that the ground-truth answer is unknown and to simulate a step-by-step diagnostic reasoning process that naturally leads to the correct answer. To further ensure the faithfulness of the distilled reasoning traces, we introduce a quality-checking stage in which GPT-4o performs self-evaluation to verify consistency between the generated reasoning and the ground-truth answer, filtering out inconsistent traces. The prompt used for this quality-checking stage is shown in Table~\ref{tab:cot-quality-check}. 

\begin{figure*}[t]
    \centering

    \begin{subfigure}[b]{0.8\textwidth}
        \centering
        \includegraphics[width=\textwidth]{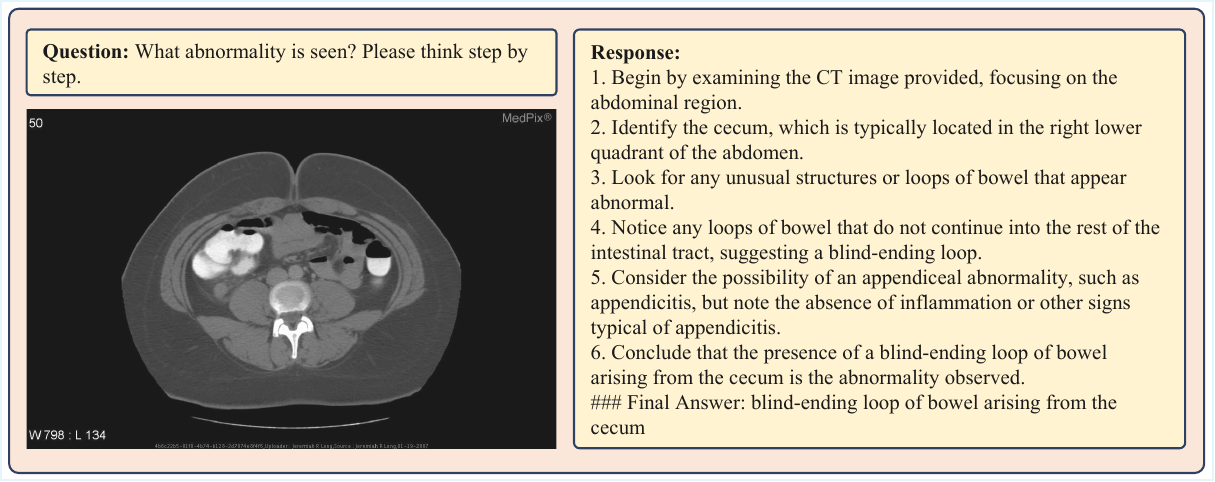}
        \caption{An example of distilled medical reasoning data for a CT scan.}
        \label{fig:cot_annotation_example_ct}
    \end{subfigure}

    \vspace{0.5em}

    \begin{subfigure}[b]{0.8\textwidth}
        \centering
        \includegraphics[width=\textwidth]{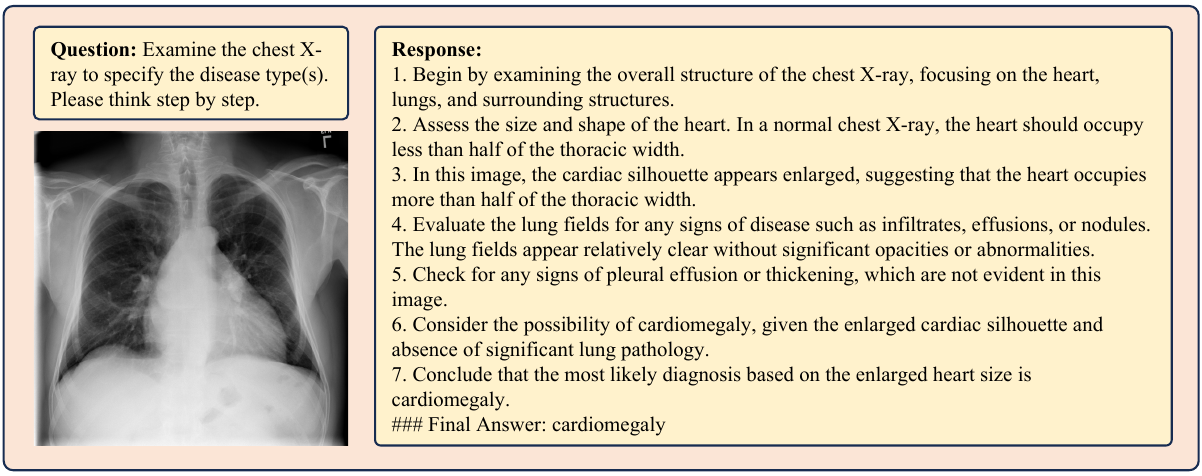}
        \caption{An example of distilled medical reasoning data for an X-ray scan.}
        \label{fig:cot_annotation_example_xray}
    \end{subfigure}

    \caption{Examples of distilled medical reasoning data for different imaging modalities.}
    \label{fig:cot_annotation_examples}
\end{figure*}

We also provide two examples of the distilled medical reasoning data in Fig.~\ref{fig:cot_annotation_example_ct} and Fig.~\ref{fig:cot_annotation_example_xray}, corresponding to CT and X-ray modalities, respectively. These examples serve as qualitative sanity checks. They illustrate that the generated CoT reasoning traces can follow reasonable diagnostic reading patterns and faithfully refer to relevant visual and clinical cues, rather than merely providing post-hoc justifications. Specifically, the CoT in Fig.~\ref{fig:cot_annotation_example_ct} follows a reasonable CT-reading sequence: localizing the image as an abdominal CT scan, identifying key anatomy in the cecum/right lower quadrant, examining nearby bowel structures for abnormal loops, characterizing the morphology as a blind-ending tubular or loop-like structure, considering a relevant differential diagnosis such as an appendiceal abnormality or appendicitis, and finally concluding that the abnormality is a blind-ending loop of bowel arising from the cecum. Similarly, the CoT in Fig.~\ref{fig:cot_annotation_example_xray} follows a reasonable chest X-ray reading sequence: assessing the overall chest structures, evaluating heart size, examining the lung fields, checking for pleural effusion or thickening, and synthesizing the findings to conclude cardiomegaly.

Nevertheless, we acknowledge that these safeguards cannot fully guarantee the faithfulness of the distilled reasoning traces. Because GPT-4o has access to the ground-truth answer during annotation, some generated CoTs may still reflect plausible post-hoc justifications rather than the actual reasoning process that led to the answer. Therefore, we regard the distilled CoT data as weakly supervised reasoning annotations rather than expert-verified clinical rationales.

\input{tables/appdx_CoT_reasoning_open}

\input{tables/appdx_CoT_reasoning_MCQA}

\input{tables/appdx_CoT_reasoning_quality_check}

% ================================
\subsubsection{Details of Medical Long-form Caption Synthesis}\label{apdx:medical-long-caption}
% ================================

\paragraph{Details of prompts.}
We detail the prompts used for medical long-form caption synthesis across various stages. The prompt for Stage 3 (factual knowledge annotation) is presented in Table~\ref{tab:medical-long-caption-prompts-stage-3}. For Stage 4 (annotation with doctor preference), the prompts tailored for different types of medical images are displayed in Tables~\ref{tab:medical-long-caption-prompts-stage-4-1},~\ref{tab:medical-long-caption-prompts-stage-4-2},~\ref{tab:medical-long-caption-prompts-stage-4-3}, and \ref{tab:medical-long-caption-prompts-stage-4-4}. Finally, the summarization prompt for Stage 5 is depicted in Table~\ref{tab:medical-long-caption-prompts-stage-5}. In this Stage 5 prompt, the variable \{doctor\_preferred\_features\} should be replaced by the doctor-preferred features specified in Tables~\ref{tab:medical-long-caption-prompts-stage-5-1} and \ref{tab:medical-long-caption-prompts-stage-5-2}, according to the type of the source image.

\paragraph{Dataset usage. }
We collect BraTS2024~\citep{de20242024}, LLD-MMRI~\citep{lou2025sdr}, MAMA-MIA~\citep{garrucho2025} and Brain-Tumor-MRI for \textbf{MRI}; Mendeley Digital Knee X-ray, CheXpert~\citep{irvin2019chexpert}, and NIH-Chest-X-Ray~\citep{rajpurkar2017chexnet} for \textbf{X-ray}; AbdomenUS~\citep{vitale2020improving} for \textbf{Ultrasound}; PAD-UFES-20~\citep{PACHECO2020106221} for \textbf{Dermoscopy}; KIPA22~\citep{HE2021102055} and DeepLesion~\citep{yan2017deeplesion} for \textbf{CT}; and CPD~\citep{wagner2023semantic}, Breast-Cancer~\citep{ding2023large}, PanNuke~\citep{gamper2020pannuke}, and EBHI-Seg~\citep{shi2023ebhi} for \textbf{Histopathology}. 

\paragraph{More details of stage 1: metadata preparation.}
For example, in CPD~\citep{wagner2023semantic}, we generate short captions using a staining method and nuclei score metadata with the template: ``\textit{A WSI image of carcinogenic DNA damage caused by ultraviolet (UV) radiation. The image is stained for \{staining\}. The relative amount of damaged nuclei (bounded in [0,1]) is \{nuclei score\}.}'' 
To minimize hallucination risk, we exclude metadata not visually inferable from the image, \eg, patient age, based on expert consultation. All personal identifiers are removed to prevent racial or demographic biases. Additionally, we manually retrieve medical knowledge from the web to enhance the annotation process, particularly for models like GPT-4o that may lack domain-specific expertise. Thus, this stage involves metadata preparation from both the dataset and external medical references.

\paragraph{More details of stage 2: RoI identification.}
RoIs are available as either segmentation masks or bounding boxes. Bounding boxes are directly overlaid on the original images, while segmentation masks are converted into bounding boxes by computing the minimal enclosing rectangles. For 3D datasets such as MAMA-MIA~\citep{garrucho2025} and KIPA22~\citep{HE2021102055}, we extract 2D slices and their corresponding masks along the z-axis, and apply the same bounding box rendering procedure to generate 2D images.

\paragraph{More details of stage 3: annotation with factual knowledge.}
This stage synthesizes image descriptions by integrating textual and visual information from the previous two stages. 
For datasets with annotated RoIs, we input bounding-box-rendered images, short captions, and retrieved medical knowledge into GPT-4o to generate descriptive text. For datasets lacking RoI annotations, GPT-4o is prompted with the original image and metadata from Stage 1. Since visual and textual inputs from the last two stages are both manually annotated, the resulting descriptions are considered highly reliable.

\paragraph{More details of stage 4: annotation with doctor preference.}
Although the generated factual descriptions are highly reliable, they are often constrained by explicitly available metadata and may overlook critical visual details beyond RoIs. To mitigate this, we consult medical professionals to identify key factors they consider when interpreting medical images. Their insights are distilled into task-specific instructions. For instance, in MRI analysis, clinicians focus on sequence type, image orientation, anatomical structures, and visible abnormalities. These instructions guide the annotation model, which, given the original image without bounding box overlays, generates alternative descriptions aligned with expert diagnostic perspectives.

\input{tables/appdx_fact_annotation_long_cap}

\input{tables/appdx_doctor_preference_ct_mri_xray_long_cap}

\input{tables/appdx_doctor_preference_hist_lesion_knee_long_cap}

\input{tables/appdx_doctor_preference_gast_long_cap}

\input{tables/appdx_doctor_preference_fundus_ultra_long_cap}

\input{tables/appdx_summarization_doctor_prefer_long_cap}

\input{tables/appdx_summarization_ct_mri_xray_hist_skin_long_cap}

\input{tables/appdx_summarization_gast_fundus_ultra_long_cap}

% ================================
\subsection{Details of \ours's training stages}
\label{apdx:sec:training_stage_data_mixture}
% ================================
The overview of the data mixture for the four training stages of \ours is shown in Table~\ref{tab:data_mix}. Recall that \ours is built upon the Qwen2.5-VL model architecture~\citep{bai2025qwen25vl}, which consists of three key components: a large language model (LLM), a vision encoder, and an MLP-based projector. We employ two parameter scales of Qwen2.5-VL, \ie, 7B-Instruct and 32B-Instruct, as the foundational models. The choice of instruct versions is motivated by two primary considerations. First, these models require significantly less alignment data during training, owing to their inherent image understanding and instruction-following capabilities. Second, they support the direct use of instruction-formatted data during the alignment phase. Consequently, \ours can adopt instruction-style data formatting from the initial training stage, which promotes consistency across the multi-stage training pipeline, reduces the overhead of additional instruction alignment, simplifies the training process, and enhances overall training stability.

\input{tables/appdx_data_mixture_4_training_stages}

We develop a four-stage training framework following a ``shallow-to-deep'' progression: (1) \textbf{Medical Shallow Alignment}, (2) \textbf{Medical Deep Alignment}, (3) \textbf{Medical Instruction Tuning}, and (4) \textbf{Medical-oriented Reinforcement Learning}. 

The initial two stages focus on constructing a robust vision-language foundation tailored to medical scenarios. Medical Shallow Alignment fine-tunes the model via a small set of medical image-text pairs, enabling it to encode medical images and generate corresponding descriptions accurately. This stage facilitates the model's understanding of visual medical content. Medical Deep Alignment builds upon this by using larger, higher-quality, and semantically richer medical image-text pair datasets. This stage deepens the model's domain knowledge and achieves finer-grained vision-language alignment. The subsequent stages aim to enhance the model's utility in practical medical settings. Medical Instruction Tuning improves the model's ability to comprehend and execute task-specific instructions across various medical use cases, thereby enhancing its generalizability to downstream tasks. Finally, Medical-oriented Reinforcement Learning incorporates the Reinforcement Learning with Verifiable Rewards (RLVR; \citep{shao2024deepseekmath,lai2025medr1,pan2025medvlmr1}) paradigm to strengthen the model's medical reasoning, problem-solving capacity, and interpretability.

For \textbf{medical shallow alignment} stage, we primarily utilize the coarse caption data for training. In this case, freezing the LLM is a deliberate choice, as the coarse caption data typically consists of short and information-sparse text. Exposing LLM to such limited textual content could potentially impair its language generation capabilities. In contrast, training vision encoder and projector enables the model to better map medical visual features into LLM's representational space. The use of coarsely annotated data is motivated by its relatively low semantic complexity, which facilitates rapid learning of the general characteristics of medical imaging modalities. This not only aids in effectively adjusting the vision encoder's outputs for better alignment with the LLM but also contributes to faster training and improved convergence stability.

For \textbf{medical deep alignment stage}, we unfreeze all the model parameters and utilize more high-quality image caption datasets for training. Exposure to these structured visual elements enhances the model's ability to perform structured data interpretation and cross-modal reasoning. This is particularly beneficial in medical contexts where diagnostic reports often include laboratory tables, time-series physiological curves, and pathological maps—visual formats that differ significantly from natural images in both distribution and semantics. By learning from these general-domain examples, the model acquires transferable skills in chart interpretation and symbolic reasoning, enabling more precise analysis of abnormal physiological metrics, lesion measurements, and their temporal dynamics in clinical scenarios.

For \textbf{medical instruction tuning stage}, in line with the prevailing practices in MLLMs~\citep{li2023llavamed,chen2024huatuogptvision,zhu2025internvl3,chen2025internvl25}, the medical instruction-tuning stage is designed to refine MLLM's instruction following capability with a broad spectrum of task directives, ensuring that its outputs are accurately aligned with user intentions. To accommodate a wide breadth of clinical use cases, we unlock all parameters and perform large-scale, end-to-end optimization of the entire model. The training corpus transcends conventional instruction formats, such as image descriptions, question-answer pairs, and multiple-choice questions, by integrating an extensive suite of scenario-oriented queries collected and synthesized from multiple sources. These queries span diagnosis, clinical examination, medical knowledge retrieval, clinical report generation, and anatomical structure localization. This breadth markedly augments the model's domain competency. In addition, we enrich the corpus with carefully vetted high-quality general-domain data and medical text data, both of which have proven effective in enhancing performance on downstream medical tasks. Incorporating such data is capable of counterbalancing the inherent image-centric information bias of image-text pair data, expanding the model's conceptual scope beyond visual content, and fostering a more holistic grasp of medical knowledge.

Moreover, our training data comprises advanced and complex data formats, such as multi-image reasoning tasks, multi-turn dialogues, and queries that require detailed reasoning processes, which demand deeper analytical acuity. Collectively, this diversified and sophisticated training data equips the MLLM to navigate complex clinical scenarios with enhanced precision and analytical capabilities. The datasets used in this stage are summarized in Table~\ref{tab:data_mix}, which are categorized based on their domain (\ie, medical or general) and modality (\ie, multimodal or textual):
\begin{itemize}
    \item For \textbf{medical multimodal} subset, we initially employ multiple public medical instruction datasets, but found that they often emphasize specific image regions, which may introduce local view bias. Thus, we supplement the training with high-quality caption datasets—selected from ROCOv2~\citep{ruckert2024rocov2}, PubMedVision~\citep{chen2024huatuogptvision}, and MIMIC-CXR~\citep{johnson2019mimiccxr}—to promote holistic visual understanding. We also incorporate clinical report generation datasets, CheXpert Plus~\citep{chambon2024chexpertplusaugmentinglarge} and IU-Xray~\citep{demner2016preparing-iu-xray}. Lastly, we synthesize additional medical data, including OCR-based, QA, and CoT data.
    \item For \textbf{general multimodal} subset, we select the instruction subsets of LLaVA-1.5~\citep{liu2024llava15} and ALLaVA~\citep{chen2024allava}.
    \item For \textbf{medical text} subset, we curate data from diverse public sources and generate instruction data via an automated synthesis pipeline, followed by strict quality validation. We further incorporate open-source long-form medical reasoning datasets, primarily distilled from OpenAI-o1\footnote{\url{https://openai.com/o1/}} and DeepSeek-R1~\citep{deepseekai2025deepseekr1}, which offer comprehensive analyses of medical problems and effectively enhance the model's medical knowledge.
    \item For \textbf{general text} subset, we leverage the text-only subsets of LLaVA-1.5~\citep{liu2024llava15}, and OpenHermes-2.5\footnote{\url{https://huggingface.co/datasets/teknium/OpenHermes-2.5}}.
\end{itemize}

For \textbf{medical-oriented RL} stage, recent advancements in reasoning models, exemplified by OpenAI's o-series and DeepSeek-R1~\citep{deepseekai2025deepseekr1}, have set new benchmarks in complex tasks through advanced post-training strategies. Central to these improvements is reinforcement learning with verifiable rewards (RLVR), with Group Relative Policy Optimization (GRPO; \citep{shao2024deepseekmath}) emerging as a prominent method due to its efficiency and effectiveness. It has been increasingly adopted for training MLLMs to enhance reasoning capabilities~\citep{meng2025mmeureka,wang2025vlrethinker,tan2025reasonrft}. Building on RLVR's success, some work applies it to medical domains for improving generalizability, interpretability, and reliability~\citep{lai2025medr1,pan2025medvlmr1}. Conventional SFT encounters key limitations: 1) over-reliance on answer supervision, which can cause overfitting and shortcut learning, particularly detrimental in medical scenarios; and 2) limited promotion of deliberate reasoning skills. RLVR addresses these issues by encouraging models to autonomously discover reasoning pathways through reward signals, rather than relying on answer memorization or teacher-guided CoT imitation~\citep{chu2025sft}. Accordingly, we adopt GRPO to train \ours, leveraging a carefully curated medical verifiable dataset.

The medical verifiable dataset is derived from multiple sources (Table~\ref{tab:data_mix}) and undergoes rigorous post-processing. Since most collected samples are in multiple-choice format, we reformulate those with word or phrase answers as open-ended questions to enhance difficulty while maintaining verifiability. This process also leads to a data balance between MCQA and open-ended QA. Additionally, binary-answer questions (\eg, ``Yes''/``No'') are downsampled to approximately 5\% of the dataset to mitigate potential bias from their over-representation. Samples are then selected based on their modality and query format, aiming to ensure the data balance. A total of 100K examples are curated to form the training set for the RL stage. For reward design, we follow common practices~\citep{deepseekai2025deepseekr1,yeo2025demystifying} by using a strict format reward and an accuracy reward, with respective weights of \texttt{0.5} and \texttt{1}.

% ================================
\subsection{Implementation Details}
\label{apdx:sec:implementations}
% ================================
As previously outlined, \ours builds upon two variants of the Qwen2.5-VL-Instruct model, with parameter sizes of 7B and 32B, and is further optimized via continual training. The standard training process is organized into three primary stages, which leads to our \ours. Across all stages, we adopt the AdamW optimizer with cosine learning rate scheduler and a warm-up step of \texttt{100}. The maximum sequence length is set to \texttt{8,192} tokens, per-device training batch size is configured as \texttt{1}, and gradient accumulation step is set to \texttt{8}.

In Medical Shallow Alignment stage, LLM remains frozen, only the vision encoder and the projection layer are fine-tuned for \textit{one} epoch, with learning rates set to \texttt{2e-6} and \texttt{1e-5}, respectively. During the Medical Deep Alignment and Medical Instruction Tuning stages, we unfreeze the LLM and fine-tune it with a learning rate of \texttt{1e-5}, while maintaining the same learning rates for the vision encoder and projector. These stages are trained for \textit{one} and \textit{two} epochs, respectively.

To enhance training efficiency, we employ data packing during the Medical Instruction Tuning stage. Preliminary experiments suggest that data packing at this stage does not adversely affect model performance. However, similar attempts in Medical Shallow and Deep Alignment stages have led to a marked decline in model performance. This decline is likely attributable to the high prevalence of short textual samples within the medical image-caption dataset.\footnote{Although we synthesize a certain amount of long-form captions, it constitutes only a small proportion of the overall dataset.} When packed into longer sequences, these short entries contribute sparse gradients during normalization, undermining the training dynamics. Additionally, data packing substantially reduces the number of training steps, which may hinder sufficient convergence. Extending the number of epochs to compensate could result in overfitting. Consequently, we opt not to use data packing during the Medical Shallow and Deep Alignment stages.

For the Medical-oriented RL stage, we initialize training from \ours checkpoint and apply GRPO for one epoch using the AdamW optimizer. The maximum sequence length is set to \texttt{4096}, with a rollout batch size of \texttt{512} and a global batch size of \texttt{128}. The learning rate is configured at \texttt{1e-6}, the sampling temperature at \texttt{1.0}, the KL divergence loss coefficient at \texttt{1e-3}, and \texttt{16} responses are sampled for each prompt. For reward design, we follow common practices~\citep{deepseekai2025deepseekr1,yeo2025demystifying} by using a strict format reward and an accuracy reward, with respective weights of \texttt{0.5} and \texttt{1}. The resultant model is named as \ours-RL.

% ================================
\subsection{More Details about \oureval}
\label{apdx:sec:medevalkit}
% ================================
The rapid progress of medical MLLMs has yielded impressive performance across various tasks. However, the absence of a unified evaluation framework has led to inconsistencies in performance comparison across models. Additionally, reproducing or deploying existing models in a standardized environment often requires substantial time and computational resources. A systematic and rigorous evaluation across diverse medical tasks is essential for accurately assessing model capabilities, identifying strengths and limitations, and providing actionable feedback for further development. Reproducible, quantitative evaluation is thus vital to advancing the optimization of medical MLLMs. An effective evaluation of medical MLLMs requires a multiple-dimensional assessment, including medical knowledge, visual understanding, and reasoning ability. While existing frameworks, such as LMMs-Eval~\citep{lmms_eval2024}, Eval-Harness,\footnote{\url{https://zenodo.org/records/12608602}} and VLMEvalKit~\citep{duan2024vlmevalkit}, can be applied to evaluate medical MLLMs, they are primarily designed for general multimodal tasks. As a result, they exhibit limitations in terms of medical data coverage and model adaptation, making them difficult to apply directly. This not only reduces evaluation efficiency but may also introduce potential biases.

To comprehensively and efficiently evaluate the performance of medical MLLMs, we develop a systematic evaluation framework, \oureval, which integrates mainstream medical benchmarks and task types, supporting a variety of question formats, including multiple-choice, closed-ended, and open-ended questions as well as medical report generation. It covers representative tasks in medical multimodal and textual understanding, offering broad applicability. \oureval accommodates both multimodal and text-only inputs, enabling unified evaluation of both medical MLLMs and LLMs. To enhance standardization and reproducibility, we standardized data preprocessing formats and post-processing protocols. A consistent model deployment and inference interface is provided to support rapid integration and one-click evaluation. For result assessment, \oureval incorporates a dual verification mechanism that combines rule-based evaluation with LLM-as-a-Judge strategy, integrating both objective and subjective assessment to improve evaluation stability and reliability. Additionally, \oureval supports inference acceleration via vLLM~\citep{kwon2023efficient}, enabling high-throughput and parallel evaluation, with strong scalability and engineering usability.

\begin{figure*}[t]
    \centering
    \includegraphics[width=\textwidth]{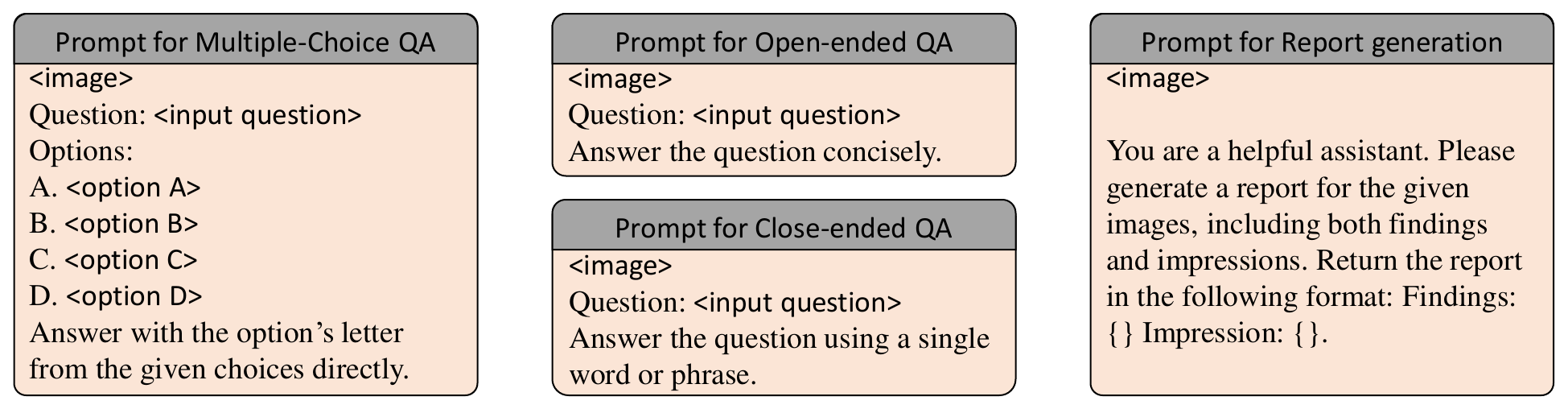}
    \caption{The prompt formats of different medical question types.}
    \label{fig:evaluation_prompts}
\end{figure*}

The input prompt format of each question type is shown in Fig.~\ref{fig:evaluation_prompts}. The judgment prompt used for GPT-4.1 in the LLM-as-a-Judge module is presented in Table~\ref{tab:prompt_for_llm_judge}. The judge is provided only with the question, the standard answer, and the user's answer. No information about the originating model is exposed, ensuring that the verdict cannot be influenced by model identity. We also normalize candidate answers (\eg, stripping formatting markers and standardizing option/letter expressions) prior to judging to reduce superficial mismatches.

\input{tables/appdx_LLM_as_a_Judge}

For the report generation task, we develop a multi-metric evaluation approach that integrates semantic and model-based metrics. Our \oureval supports the following metrics, and in this paper, we selectively present a subset of these.

The semantic evaluation metrics:
\begin{itemize}
    \item \textbf{Rouge}~\citep{lin2004rouge}: ROUGE (Recall-Oriented Understudy for Gisting Evaluation) comprises a set of metrics designed to automatically assess summary quality by comparing generated summaries against reference human-authored summaries. This metric evaluates the overlap between the generated text and reference summaries, focusing particularly on recall-oriented measurements.
    \item \textbf{CIDEr}~\citep{vedantam2015cider}: CIDEr (Consensus-based Image Description Evaluation) is a metric that quantifies the similarity between generated sentences and a collection of human-authored reference sentences. 
    This metric specifically measures how well the generated content aligns with human consensus, effectively capturing the semantic similarity between machine-generated and reference descriptions.
\end{itemize}

The model-based evaluation metrics:
\begin{itemize}
    \item \textbf{SembScore}~\citep{smit2020chexbert}: SembScore (CheXbert Labeler Vector Similarity) is a domain-specific metric developed for radiology report evaluation. This metric calculates the cosine similarity between vectors of 14 pathological indicators, which are automatically extracted by the CheXbert labeler from both model-generated and ground-truth radiology reports. This specialized metric focuses specifically on assessing the clinical content alignment between generated and reference reports.
    \item \textbf{RadCliQ}$_{v1}$~\citep{yu2023evaluating}: \textbf{RadCliQ}$_{v1}$\footnote{We implement it using the official codebase at \url{https://github.com/rajpurkarlab/CXR-Report-Metric}} is a comprehensive metric designed for evaluating radiology report generation, integrating multiple complementary measures including BLEU, BERTScore, SembScore, and RadGraph-F1 to provide a holistic assessment of report quality. 
    \item \textbf{RaTEScore}~\citep{zhao2024ratescore}: RaTEScore\footnote{We implement it using the official codebase at \url{https://github.com/MAGIC-AI4Med/RaTEScore}} is an entity-aware metric specifically designed for radiology report evaluation. This metric emphasizes critical medical entities, including diagnostic findings and anatomical details, while demonstrating robustness to complex medical synonyms and sensitivity to negation expressions.
\end{itemize}

Table~\ref{tab:benchmarks_overview} summarizes the statistics of the evaluation benchmarks utilized in the \oureval framework.

\input{tables/appdx_benchmarks_overview}

% ================================
\subsection{Effectiveness of \ours Pipeline}
\label{apdx:sec:lingshu_internvl}

\begin{table*}[t]
\centering
\small
\caption{Performance of applying \ours Pipeline on both Qwen2.5-VL-7B and InternVL3-8B.}
\adjustbox{max width=1.0\textwidth}{
\begin{tabular}{lcccccccc}
\toprule
\textbf{Models} & \textbf{MMMU-Med} & \textbf{VQA-RAD} & \textbf{SLAKE} & \textbf{PathVQA} & \textbf{PMC-VQA} & \textbf{OmniMedVQA} & \textbf{MedXpertQA} & \textbf{Avg.}\\

\midrule
 Qwen2.5-VL-7B & 50.6 & 64.5 & 67.2 & 44.1 & 51.9 & 63.6 & 22.3 & 52.0\\
 + \ours Pipeline& \textbf{54.0} & \textbf{67.9} & \textbf{83.1} & \textbf{61.9} & \textbf{56.3} & \textbf{82.9} & \textbf{26.7} & \textbf{61.8}\\
\midrule
InternVL3-8B & \textbf{59.2} & 65.4 & {72.8} & {48.6} & {53.8} & 79.1 & {22.4} & {57.3} \\
+ \ours Pipeline& 49.1 & \textbf{73.0} & \textbf{91.6} & \textbf{74.9} & \textbf{55.8} & \textbf{79.7} & \textbf{27.5} & \textbf{64.5}\\
\bottomrule
\end{tabular}
}
\label{tab:mm_lingshu_internvl}
\end{table*}
As demonstrated in our main paper, the \ours pipeline yields substantial improvements when applied to Qwen2.5-VL-7B. To verify that these gains stem from our proposed methodology rather than the specific choice of base model, we conducted additional experiments applying the identical data curation and training pipeline to InternVL3-8B, a different vision-language foundation model with distinct architecture and pretraining strategy. As shown in Table~\ref{tab:mm_lingshu_internvl}, the \ours pipeline improves the average performance and most individual medical benchmarks for InternVL3-8B, while potentially trading off performance on MMMU-Med. The average score increased from 57.3 to  64.5 (+7.2 points), including VQA-RAD (65.4→73.0), SLAKE (72.8→91.6), and PathVQA (48.6→74.9). The only exception is MMMU-Med, which shows approximately a 10-point decrease (59.2→49.1). A possible explanation is that the InternVL series (InternVL2.5 and InternVL3) may be particularly well-optimized for popular general benchmarks such as MMMU, as suggested by its notably high baseline on this benchmark in Table~1 of the main paper. Taken together, these results suggest that the effectiveness of the \ours pipeline is largely transferable across different base-model architectures, rather than being tied to a particular backbone.

\subsection{Physician Evaluation Setup}

\textbf{Raters.} The evaluation was conducted by two board-certified radiologists (an associate chief radiologist and a senior attending radiologist) with 12 and 7 years of post-residency clinical experience, respectively. All cases in both the medical diagnosis and report generation tasks were graded by these radiologists, ensuring that the radiology reports under evaluation were assessed by domain-matched experts. Neither rater was involved in model development, and both were blinded to model identities throughout the study.

\textbf{Rating Procedure.} We randomly sampled 50 cases from each of the medical diagnosis and medical report generation test sets. For each case, responses were generated by three models (\ours, GPT-4.1, and Qwen2.5-VL) and strictly anonymised before annotation: model identities were concealed and responses were randomly assigned as Model-1, Model-2, and Model-3 to eliminate any potential source bias. Physicians rated each response independently on the following six dimensions using a 1–5 integer scale, where higher scores indicate better performance:
\begin{itemize}[noitemsep,topsep=0pt]
\item \textbf{Clinical Relevance}: Whether the response captures the key clinical features of the case without omitting critical information.
\item \textbf{Clinical Actionability}: Whether the proposed management plan is appropriate and provides clinically valuable guidance.
\item \textbf{Safety}: Whether the response contains potentially misleading or dangerous information, hallucinated content, or recommendations that could delay or misdirect clinical care.
\item \textbf{Completeness}: Whether the response is comprehensive and free of significant omissions.
\item \textbf{Clarity}: Whether the response is clearly expressed, logically coherent, and easy to follow.
\item \textbf{Professionalism}: Whether medical terminology is used accurately and the response adheres to accepted clinical conventions.
\end{itemize}

\textbf{Score aggregation and inter-rater agreement.}
The final score of each response on each dimension is the mean of the two radiologists' independent ratings. Because each response was rated on an ordinal 1--5 scale, we quantified rating consistency with the quadratic-weighted Cohen's $\kappa$, which penalises larger disagreements more heavily. Agreement was \emph{substantial} across all six dimensions (mean $\kappa = 0.64$, range $0.59$--$0.68$)~\citep{landis1977}. In addition, the two radiologists assigned identical scores in 36\% of responses and agreed within $\pm 1$ point in 81\%.

\subsection{Effect of General-domain Multimodal Data}
\begin{figure*}[t]
    \centering
    \includegraphics[width=0.95\textwidth]{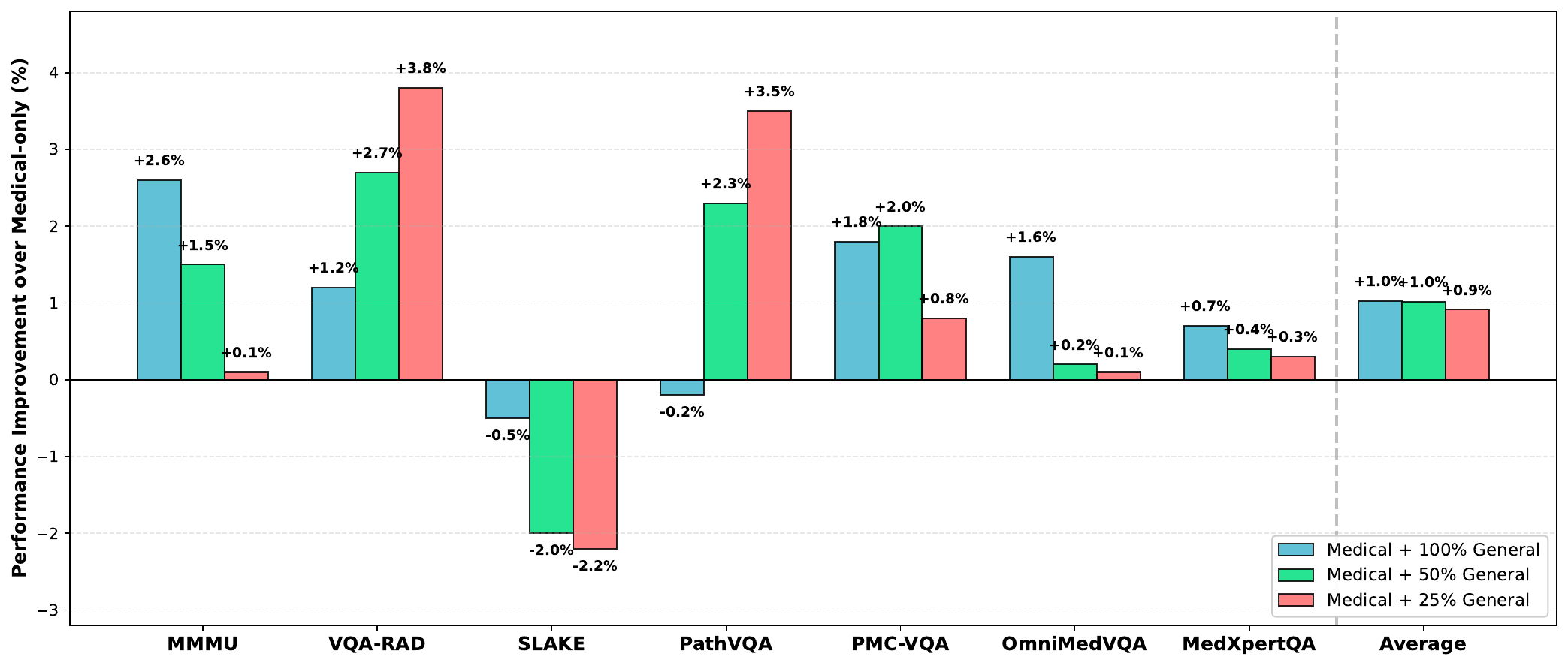}
    \caption{Performance improvement of adding general multimodal data to \ours training.}
    \label{fig:general_ratio}
\end{figure*}

Figure~\ref{fig:general_ratio} illustrates the impact of incorporating general multimodal data into \ours training. We observe a consistent improvement in performance as the proportion of general data increases from 25\% to 100\%. This indicates that general-domain data contributes meaningfully to medical performance, likely because such data contains relevant domain-specific medical knowledge. However, given the relatively small performance variance across these ratios (ranging from 0.9\% to 1.0\% on average), we forgo exhaustive exploration of data mixing strategies and instead utilize all available general-domain data for training.

\subsection{Prompt for Evaluating the Data Quality}
\label{apdx:sec:prompt_data_quality}
We sample 100 examples from each of the following data sources: our synthetic multimodal caption/instruction data, collected caption data for stage 1 and stage 2 training, medical text, medical chain-of-thought (CoT) text, collected medical multimodal data, and general multimodal data. We leverage GPT5.2 to annotate the data quality for each category. The detailed prompts can be found in Table~\ref{tab:eval_data_quality}. Regarding response length, we compute the tiktoken length and apply the following grading criterion:
\begin{itemize}[noitemsep,topsep=0pt]
    \item Score 0: length $< 10$
    \item Score 1: $10 \leq \text{length} < 100$
    \item Score 2: $100 \leq \text{length} < 200$
    \item Score 3: $200 \leq \text{length} < 500$
    \item Score 4: $500 \leq \text{length} < 1000$
    \item Score 5: length $\geq 1000$
\end{itemize}

\input{tables/appdx_eval_data_quality}

% ================================
\subsection{More Case Studies}
\label{apdx:sec:case_studies}
% ================================
Here we present illustrative case studies that showcase \ours in various clinical applications. All examples are produced with the \ours-32B model and, compared with equivalently sized Qwen2.5-VL. The resulting comparisons elucidate the tangible advantages \ours can offer to both healthcare professionals and patients.

% ================================
\subsubsection{VQA across Different Medical Imaging Modalities}
\label{apdx:ssec:vqa_modalities}
\begin{figure*}[t]
    \centering
    \includegraphics[width=\textwidth]{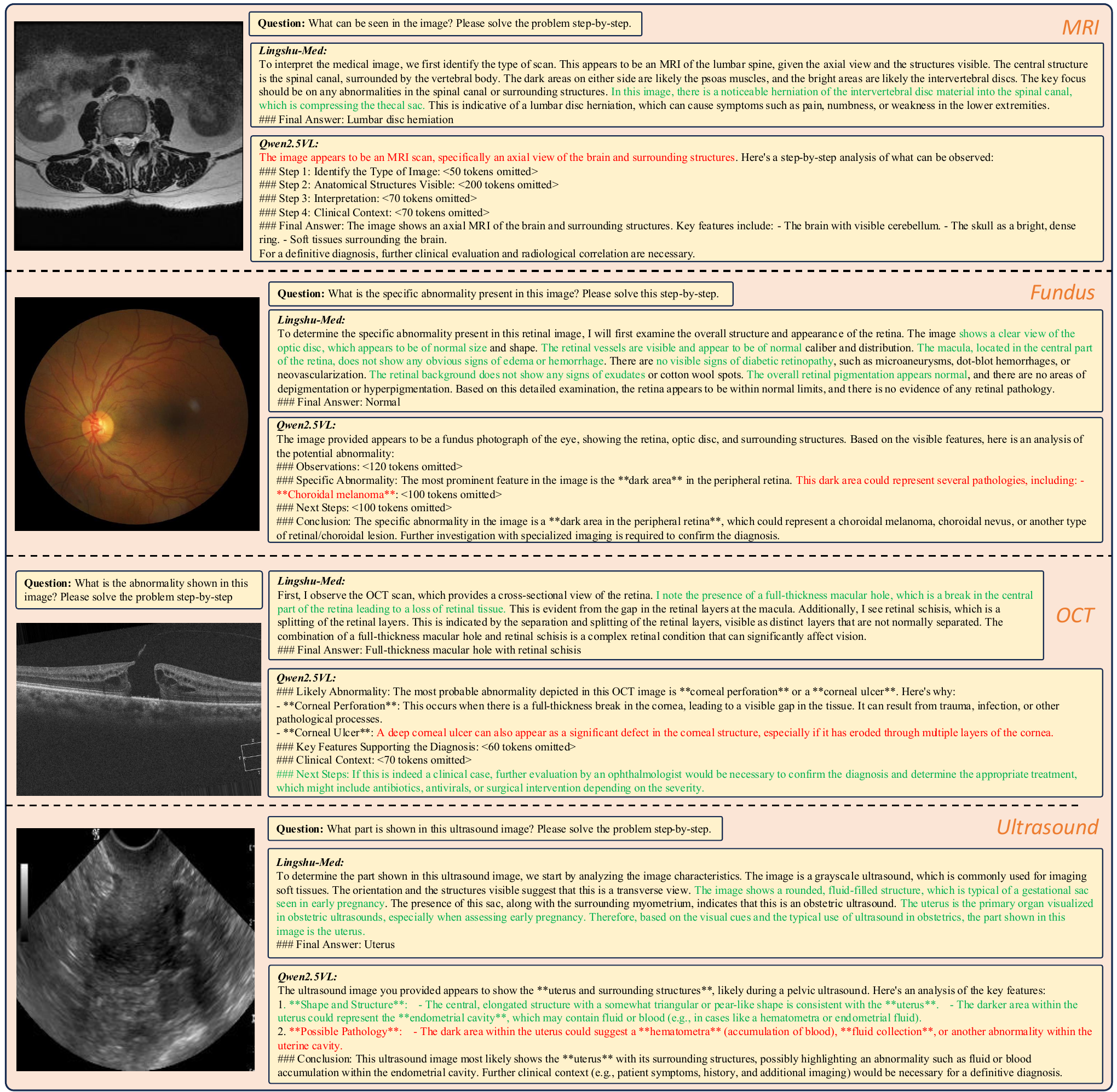}
    \caption{Case studies on \ours in visual question answering across various medical imaging modalities (part 1). The \textcolor{green}{ green parts} are some important reasoning during the generation. The \textcolor{red}{red part} indicates the thoughtless reasoning.}
    \label{fig:modality_vqa1}
\end{figure*}

\begin{figure*}[t]
    \centering
    \includegraphics[width=\textwidth]{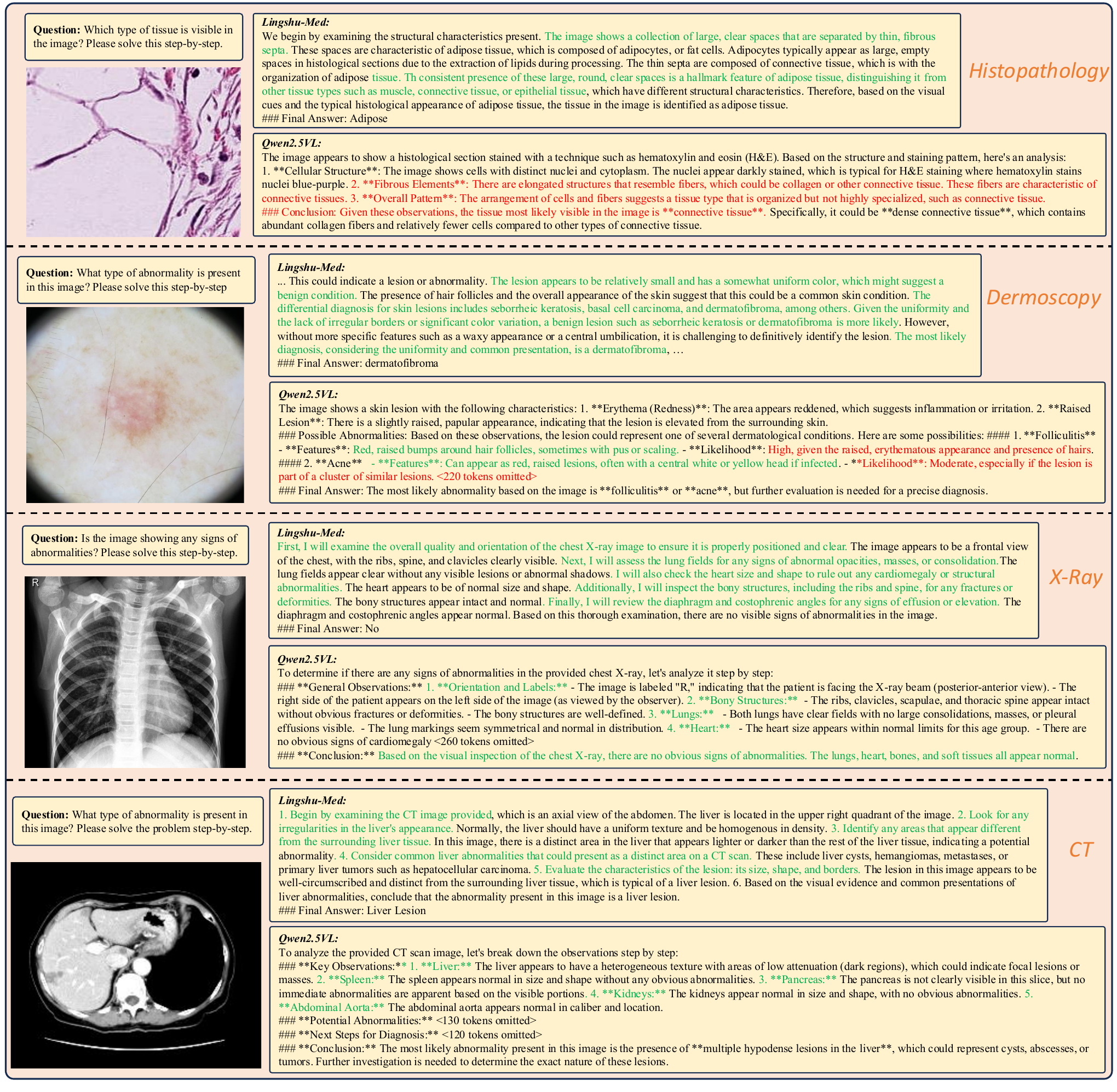}
    \caption{Case studies on \ours in visual question answering across various medical imaging modalities (part 2).}
    \label{fig:modality_vqa2}
\end{figure*}

% ================================
Fig.~\ref{fig:modality_vqa1} and Fig.~\ref{fig:modality_vqa2} present medical VQA cases of \ours across eight distinct medical modalities, where \ours is capable of transparently showing its reasoning process to derive a reliable decision.

\textbf{\ours.} \ours traces a diagnostic path that mirrors standard clinical workflows. In the chest-radiograph example, for instance, it inspects anatomical regions in the canonical order: first the ``\emph{lungs},'' then the ``\emph{heart},'' followed by the ``\emph{bony structures},'' and finally the ``\emph{diaphragm}'' and ``\emph{costophrenic angles}.'' This systematic progression ensures comprehensive coverage of relevant findings. Rather than issuing a final diagnosis immediately, the model incrementally constrains the differential. In the dermoscopy case, the initial hypothesis set included ``\emph{seborrheic keratosis},'' ``\emph{basal-cell carcinoma},'' and ``\emph{dermatofibroma}.'' Continued analysis reduced the options to the first and third, before converging on ``\emph{dermatofibroma}'' after incorporating additional visual cues. This step-wise narrowing underscores the reliability and clinical plausibility of \ours's decision-making process.

\textbf{Qwen2.5-VL.} The model demonstrates strong logical reasoning and presentation quality, with even superior formatting compared to \ours in certain aspects. However, it suffers from a critical limitation: severely compromised factual correctness across multiple medical imaging modalities. In the MRI case, Qwen2.5-VL misidentifies a lumbar spine MRI as a brain MRI, a fundamental anatomical error. Similarly, in histopathology analysis, it incorrectly classifies adipose tissue as connective tissue. More concerning are the hallucinations observed in fundus and ultrasound images, where the model fabricates non-existent diseases. In OCT and dermoscopy cases, while the visual descriptions are accurate, the final diagnoses are incorrect, undermining clinical utility. Nevertheless, Qwen2.5-VL performs good only on chest X-rays and CT scans, systematically describing anatomical structures and reaching correct conclusions. This is presumably due to the abundance of such data in training corpora (as shown in Fig. 3 in main paper), leading to better model exposure and performance on these common modalities. These findings underscore a critical gap between superficial reasoning coherence and medical factual accuracy.

% ================================
\subsubsection{Report Generation}
\label{apdx:ssec:case_report}
\begin{figure*}[t]
    \centering
    \includegraphics[width=\textwidth]{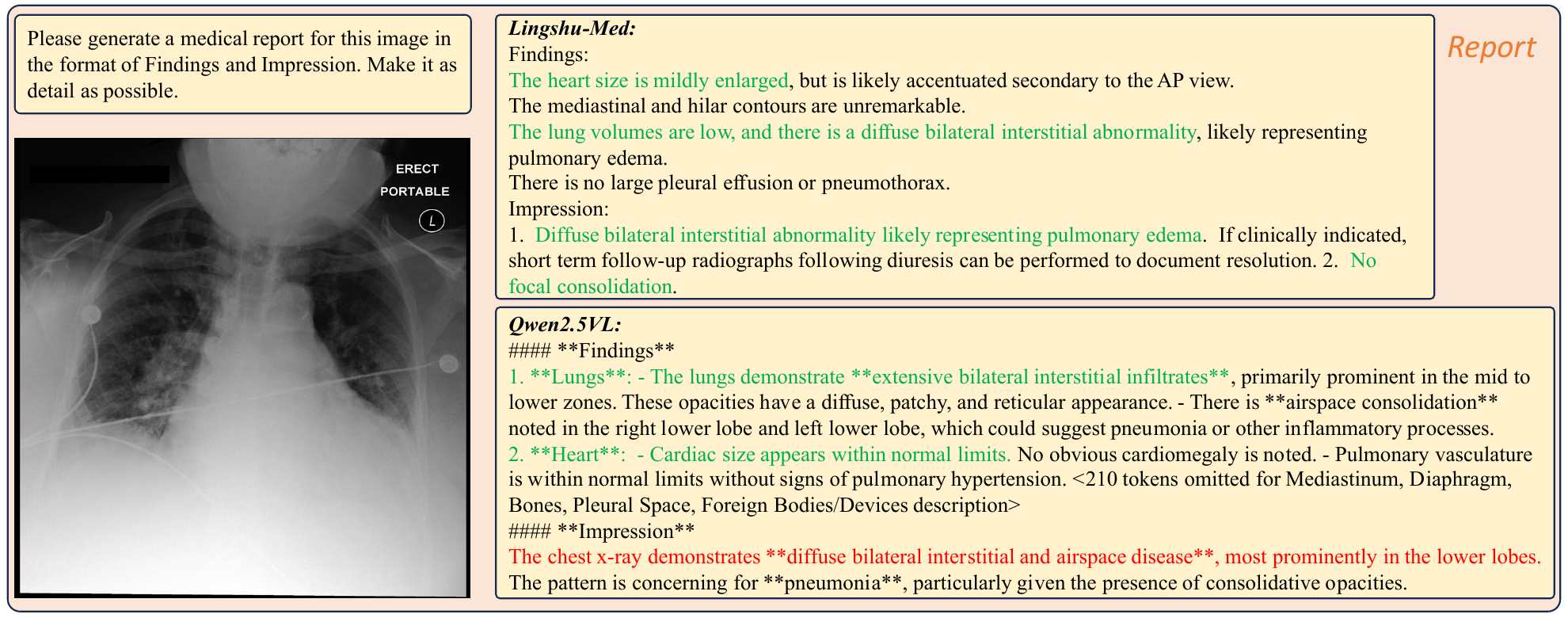}
    \caption{Case study of \ours and Qwen2.5-VL in generating medical report.}
    \label{fig:report}
\end{figure*}

% ================================
Fig.~\ref{fig:report} illustrates an X-ray report produced by \ours and Qwen2.5-VL. \ours rigorously follows the conventional radiological structure, \ie, distinct ``Findings'' followed by ``Impression.'' Within the Findings, it correctly delineates suspected abnormalities involving both the cardiac silhouette and pulmonary parenchyma. \ours also exhibits clinical insight by appending actionable recommendations, such as obtaining a follow-up radiograph after diuretic therapy. On the contrary, despite accurate feature description in Findings, Qwen2.5-VL reaches an erroneous Impression of pneumonia, highlighting a critical gap between observation and clinical reasoning compared to \ours's correct diagnosis of pulmonary edema. These results indicate that \ours not only detects radiographic anomalies with precision but also furnishes contextually appropriate management advice, reflecting the holistic practices of experienced radiologists.

% ================================
\subsubsection{Medical Diagnosis}
\label{apdx:ssec:case_diagnosis}
\begin{figure*}[t]
    \centering
    \includegraphics[width=\textwidth]{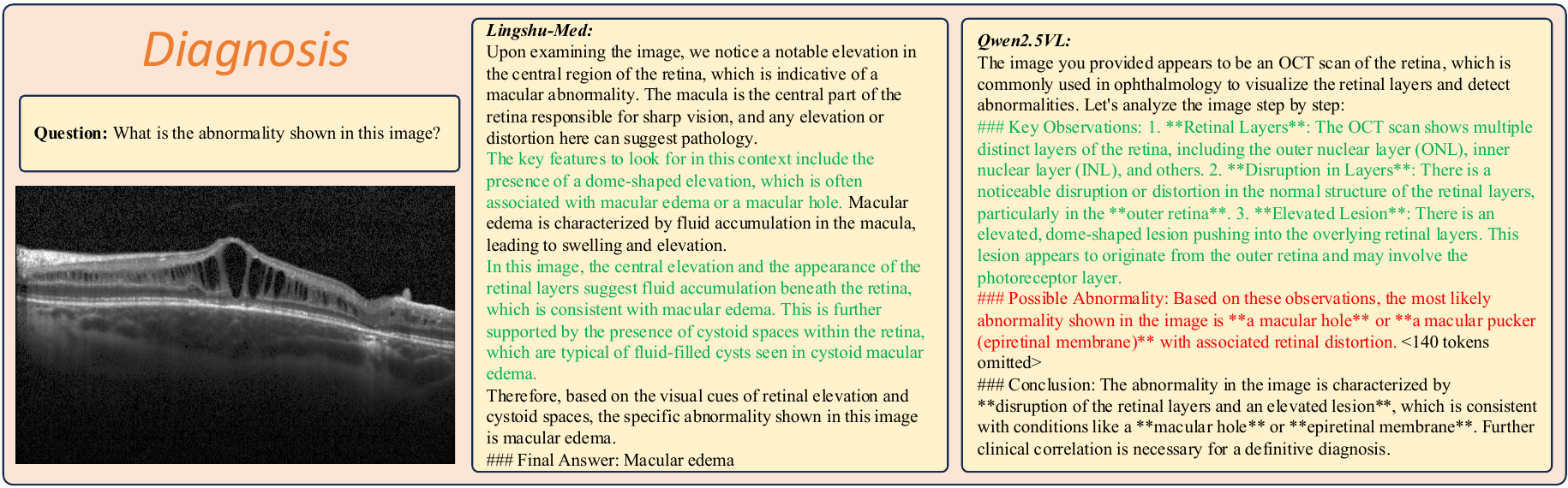}
    \caption{The comparison between \ours and Qwen2.5-VL on medical diagnosis.}
    \label{fig:diagnosis}
\end{figure*}

% ================================
Fig.~\ref{fig:diagnosis} compares the diagnostic reasoning of \ours with Qwen2.5-VL on a medical diagnosis case. Guided by the dome-shaped retinal elevation, \ours initially considers two differential diagnoses, ``\emph{macular edema}'' and ``\emph{macular hole},'' and subsequently converges on ``\emph{macular edema}'' after incorporating additional clinical cues such as subretinal fluid accumulation and cystoid spaces.  Qwen2.5-VL once again exhibits correct understanding of the medical image features but arrives at an incorrect diagnostic conclusion. From these cases, \ours demonstrates markedly reduced hallucination rates and improved clinical accuracy. However, \ours's outputs exhibit less structural formatting and coverage compared to Qwen2.5-VL, indicating a clear avenue for methodological enhancement in our future work.

% ================================
\subsubsection{Medical Knowledge}
\label{apdx:ssec:case_medical}
\begin{figure*}[t]
    \centering
    \includegraphics[width=\textwidth]{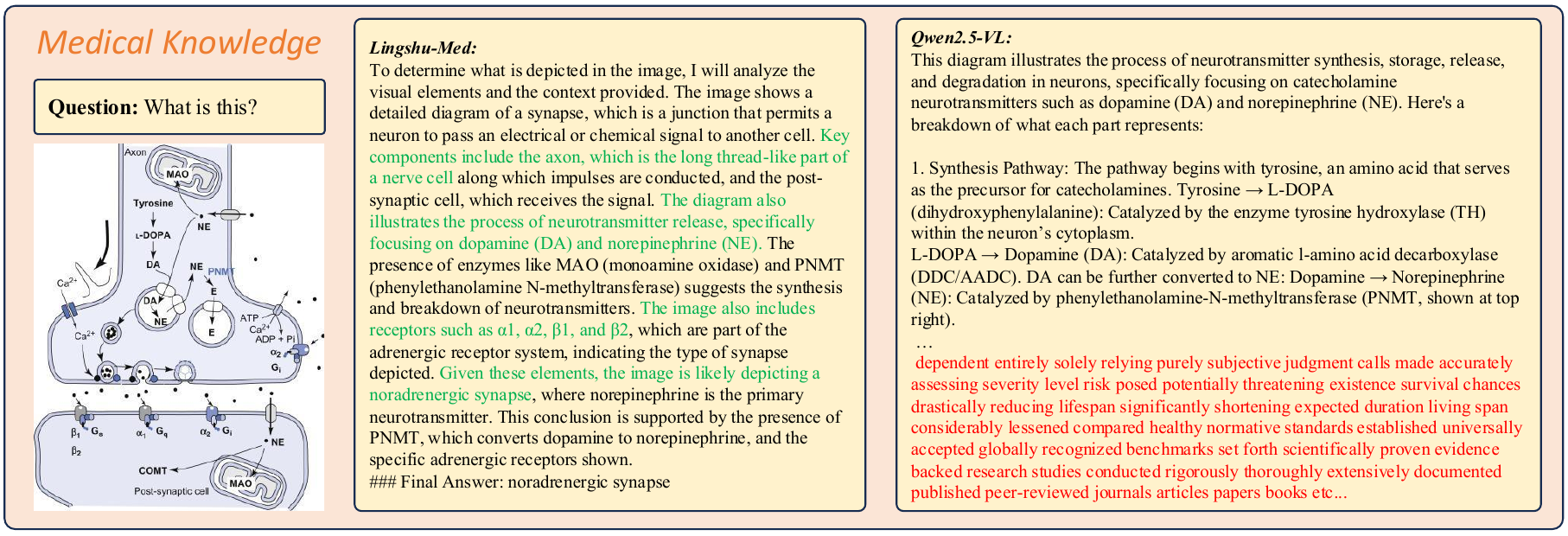}
    \caption{Case study  in explaining medical knowledge.}
    \label{fig:knowledge}
\end{figure*}

% ================================
Beyond its proficiency in clinical image interpretation, \ours exhibits sustained mastery of specialized biomedical knowledge. Depicted in Fig.~\ref{fig:knowledge}, \ours clearly interprets the intricate processes occurring within synapses. Notably, it precisely identifies the synapse type as noradrenergic, based on a thorough understanding of neurotransmitter and receptor interactions. This is in stark contrast to the original Qwen2.5-VL model, which, while adept at describing synaptic processes, fails to accurately classify them, a crucial piece of information. Moreover, the Qwen2.5-VL produces a great amount of incoherent or irrelevant content, a sign of unreliability in its understanding and application of medical knowledge.

% ================================
\subsubsection{Public Health}
\label{apdx:ssecc:pub_heealth}
\begin{figure*}[t]
    \centering
    \includegraphics[width=\textwidth]{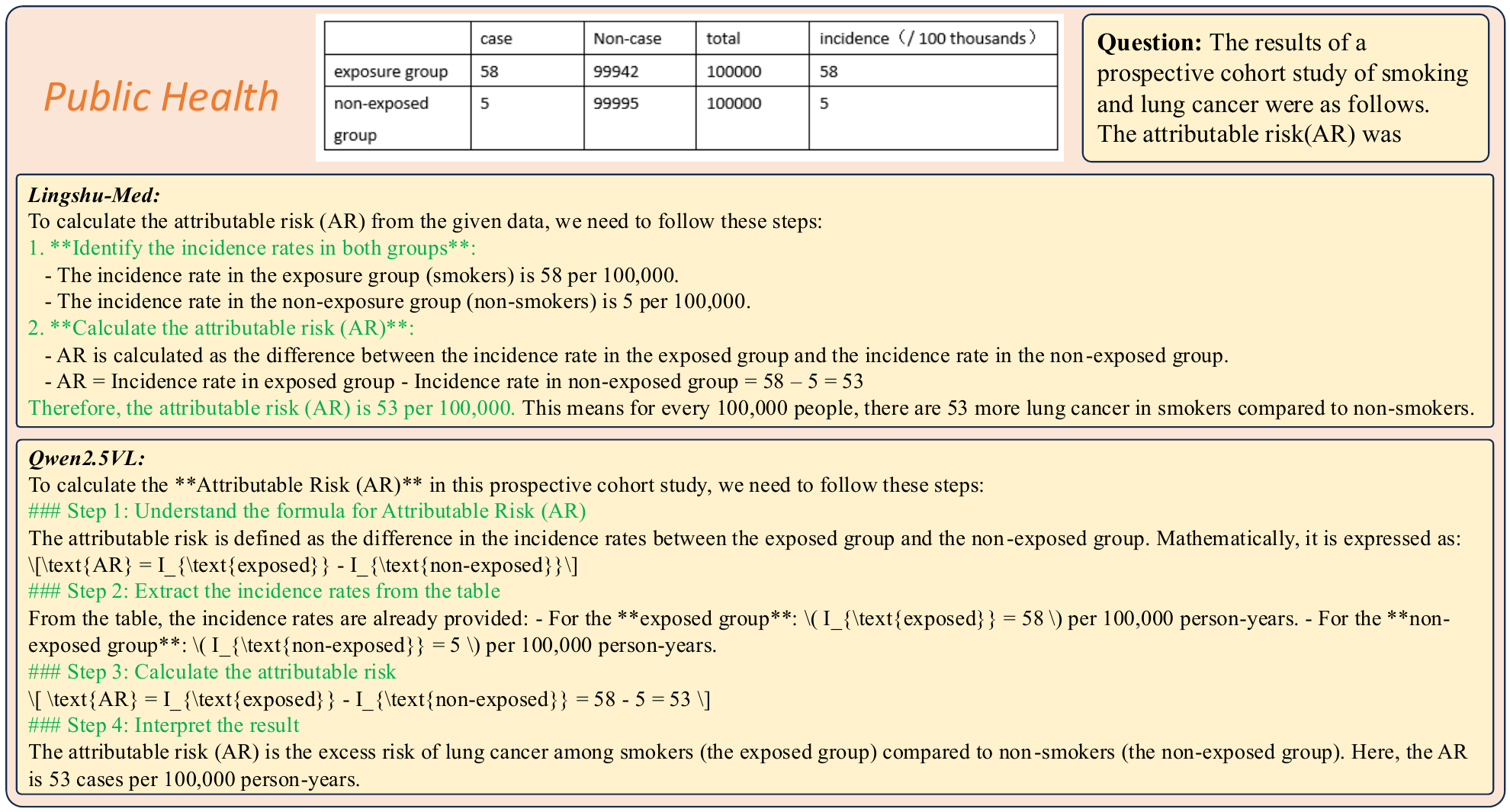}
    \caption{Case study in dealing with a public health problem.}
    \label{fig:public}
\end{figure*}

% ================================
As shown in Fig.~\ref{fig:public}, both \ours and Qwen2.5-VL precisely extract the relevant elements from the cohort‐study table, identifying incidence rates of $58$ per $100,000$ in the smoking (exposed) cohort and $5$ per $100,000$ among non-smokers. They apply the standard formula, subtracting the latter from the former to obtain an attributable risk of $53$ per $100,000$. This example highlights that \ours maintains the strong table comprehension abilities of its foundation model, accurately processing structured epidemiological data and deriving clinically meaningful risk estimates.

% ================================
\subsubsection{Patient-Doctor Dialogue}
\label{apdx:ssec:case_dialogue}
\begin{figure*}[t]
    \centering
    \includegraphics[width=\textwidth]{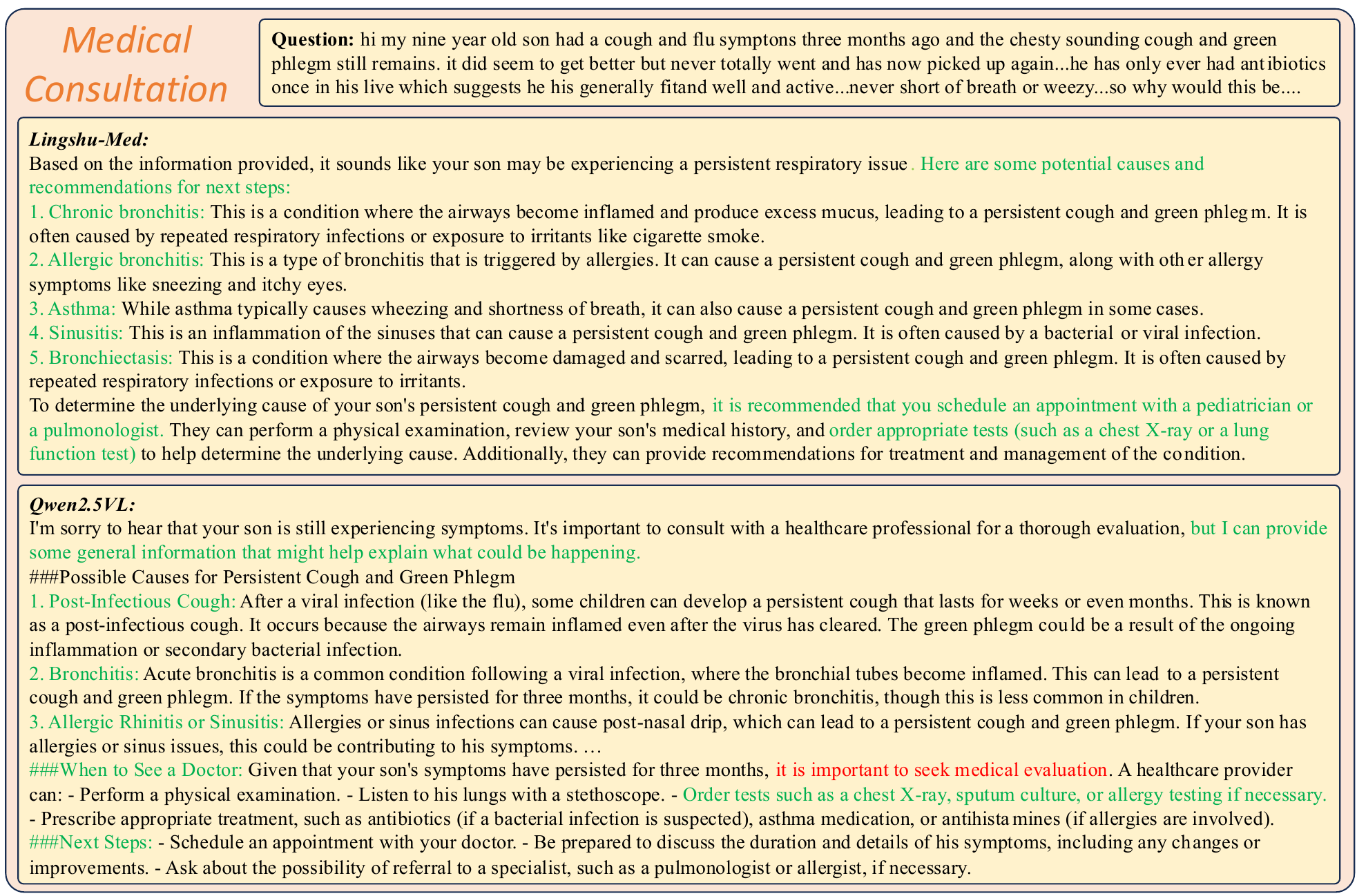}
    \caption{Case study in a real-life medical support.}
    \label{fig:patient-doctor-dialog}
\end{figure*}

% ================================
Fig.~\ref{fig:patient-doctor-dialog} showcases a simple real-world consultation scenario with \ours and Qwen2.5-VL respectively. Both models interpret the patient's concerns, propose plausible differential diagnoses, direct the individual to the relevant clinical specialty, and specify suitable diagnostic tests. However, \ours demonstrates more clinically actionable guidance by explicitly recommending consultation with specific specialists (pediatrician or pulmonologist), whereas Qwen2.5-VL, despite superior formatting and comprehensive explanations, provides more general advice without specific specialist referrals.

\end{document}

%% file: sections/0_abstract.tex
\begin{abstract}
Multimodal Large Language Models (MLLMs) excel at understanding generic visual content, such as landscapes, objects, and events, thanks to extensive datasets and advanced training regimes. However, their effectiveness in medical applications remains limited due to the inherent discrepancies between data and tasks in medical scenarios and those in the general domain. Existing medical MLLMs face the following critical deficiencies: 1) inadequate coverage of medical knowledge beyond imaging; 2) elevated propensity for hallucinations due to suboptimal data curation; and 3) limited reasoning capacity tailored to complex medical tasks. To address these challenges, we first propose a comprehensive data-curation procedure that 1) efficiently acquires rich medical knowledge data not only from medical imaging but also from extensive medical texts and general domain data; and 2) synthesizes high-quality medical captions, visual question answering, and reasoning samples. Leveraging the curated data, we build a multimodal dataset imbued with extensive medical knowledge and develop our medical-specialized MLLM, \ours, which undergoes multi-stage training to embed the medical expertise and enhance task-solving capabilities progressively. We also investigate reinforcement learning with verifiable rewards to further refine \ours's medical reasoning abilities. For rigorous assessment, we introduce \oureval, a unified evaluation framework that consolidates the leading multimodal and textual medical benchmarks for standardized, fair, and efficient model assessment. On three core medical tasks—multimodal QA, textual QA, and radiology report generation, \ours consistently outperforms existing multimodal baselines in most tasks. Moreover, we conduct five case studies drawn from real-world clinical scenarios that illustrate its practical utility in medical contexts.\footnote{\ours models and training data are available at \url{https://huggingface.co/lingshu-medical-mllm} }
\end{abstract}

%% file: sections/1_keywords.tex
\begin{IEEEkeywords}
Multimodal Large Language Models, Medical Multimodal Foundation Model, Medical Understanding, Medical Reasoning, Medical Image Analysis, Reinforcement Learning, Data Synthesis, Medical Evaluation Framework, Medical Question Answering, Medical Report Generation, Medical Diagnosis.
\end{IEEEkeywords}

%% file: sections/2_introduction.tex
% ================================
\section{Introduction}
\label{sec:intro}
% ================================

\IEEEPARstart{R}{}ecent advancements in Multimodal Large Language Models (MLLMs) have significantly propelled progress in multimodal domains~\citep{openai2024gpt4o,gemini2025gemini,zhu2025internvl3}, demonstrating capabilities that approach expert-level proficiency in areas such as image captioning, visual question answering, and complex reasoning. Despite these successes, the performance of MLLMs in medical applications remains notably constrained~\citep{yan2023multimodal,lee2023benefits}. Biomedical image-text pairs differ fundamentally from typical web-based content, presenting unique challenges that general visual assistants are not well-equipped to handle~\citep{li2025gmaivl}. Lacking domain-specific medical visual knowledge, these models often respond to biomedical inquiries with uncertainty or, worse, with incorrect information and hallucinations~\citep{li2023llavamed,chen2024huatuogptvision}.

To address these gaps, recent studies have adapted models for medical contexts by integrating general-purpose MLLMs with medical multimodal data to build specialized models~\citep{wu2023generalist,zhang2024biomedgpt,zhang2024pmcvqa,seyfioglu2025quiltllava}. These models demonstrate encouraging results in medical visual question answering (VQA), report generation, and diagnostic support, underscoring the potential of MLLMs in healthcare scenarios. Pioneering models, such as LLaVA-Med~\citep{li2023llavamed}, directly employ PubMed-derived datasets for vision-language alignment and fine-tuning. However, their performance remains limited due to low data quality and high levels of noise within PubMed. Subsequent research~\citep{chen2024huatuogptvision,li2025gmaivl,lin2025healthgpt} addresses the challenges with refined training recipes and larger, higher-quality medical datasets. Concurrently, advances in reasoning models, exemplified by OpenAI o-series\footnote{\url{https://openai.com/index/introducing-o3-and-o4-mini/}} and DeepSeek-R1~\citep{deepseekai2025deepseekr1}, set new benchmarks through advanced post-training techniques. Inspired by reinforcement learning with verifiable rewards (RLVR~\citep{shao2024deepseekmath}), some studies~\citep{lai2025medr1,pan2025medvlmr1} integrate it into medical models to improve reasoning and enhance performance in clinical applications.

Despite significant progress, existing medical MLLMs face four critical limitations that fundamentally constrain their clinical utility: \textbf{First, inadequate medical knowledge coverage beyond imaging}. Existing approaches typically focus on distilling image-text pairs from advanced models~\citep{openai2024gpt4o,gemini2025gemini}. This approach is scalable, but image-based prompts fail to elicit crucial non-visual medical knowledge in pharmacology, public health, and clinical contexts from advanced models. This narrow focus on visual modalities creates knowledge gaps that limit performance on tasks requiring comprehensive medical understanding. \textbf{Second, increasing hallucinations without proper data quality control.} Many distillation methods rely solely on model outputs without additional supervision, which increases the risk of generating false information. \textbf{Third, lack of systematic training strategies and data composition strategies for medical MLLMs.} Most existing works adopt ad-hoc approaches: directly fine-tuning general-purpose MLLMs with all available medical multimodal data~\cite{chen2024huatuogptvision,lin2025healthgpt}. However, critical questions remain unexplored: \textit{when to introduce different data types}, \textit{what each component contributes}, and \textit{whether non-medical data benefits medical training}. This lack of principled investigation results in suboptimal protocols that fail to effectively leverage available data.

In this work, we develop our medical MLLMs \ours\footnote{\ours originates from a chapter title of \emph{Huangdi Neijing} (\url{https://en.wikipedia.org/wiki/Huangdi_Neijing}), one of the earliest classical texts of traditional Chinese medicine.}. \ours addresses the above limitations by: 1) a comprehensive data curation framework that efficiently acquires medical knowledge from diverse sources: not only medical multimodal data but also extensive medical texts and general-domain data; 2) A robust data synthesizing  pipeline for generating captions, QA pairs, and chain-of-thought (CoT) reasoning samples with rigorous quality control; and 3)  systematic investigation of training strategies and data composition through comprehensive ablation studies, based on which we design a principled three-stage training paradigm that progressively infuses medical knowledge while preserving general capabilities.
We also investigate RLVR for enhancing multimodal medical reasoning. While our initial exploration does not yield significant improvements, we conduct a thorough analysis to identify key challenges and provide valuable insights for future work. In addition, we find that current models are often evaluated in isolated, non-standardized environments. The lack of accessible and unified evaluation frameworks further hinders the development of the medical field. Therefore, we introduce \oureval, a unified framework that consolidates the major multimodal and textual medical benchmarks for an efficient, standardized model assessment.

Extensive experiments demonstrate that \ours consistently achieves SOTA performance across most multimodal and textual medical VQA tasks, as well as in report generation, in both 7B and 32B configurations. Notably, in medical VQA, \ours-32B outperforms the second-best model by an average of $7.2$ accuracy points across seven tasks, even surpassing proprietary models like GPT-4.1 and Claude Sonnet 4. Through comprehensive analyses, we reveal critical insights for medical MLLM development: (1) medical textual data and general-domain data contain essential medical knowledge absent in multimodal medical datasets, significantly improving knowledge coverage and downstream performance; (2) data with high logical complexity substantially enhances medical reasoning capabilities; (3) early-stage training on medical captions, despite temporarily sacrificing instruction-following ability, establishes a stronger foundation that enables higher performance ceilings in later stages. Finally, the case study underscores the great potential of \ours across numerous medical applications, including medical report generation, medical consultation, and clinical surgery assistance.

In summary, our contributions are fourfold: 1) \textbf{Data}: We introduce a novel data curation pipeline that efficiently gathers medical knowledge from various resources and synthesizes high-quality medical captions, QA pairs, and CoT reasoning samples; 2) \textbf{Training}: We conduct systematic investigations of training strategies and data composition for medical MLLMs through comprehensive ablation studies, revealing insights into optimal inclusion and integration of training data. Based on these findings, we propose \ours, medical-specific MLLMs in 7B and 32B sizes, trained with a principled three-stage paradigm that progressively infuses medical knowledge; 3) \textbf{Evaluation}: We propose \oureval, a unified evaluation framework that consolidates major benchmarks for multimodal and textual medical tasks, streamlining model assessment and advancing standardized performance evaluation in the medical field; 4) \textbf{Performance}: Through rigorous experimental validations, \ours demonstrates state-of-the-art performance across multiple multimodal and textual medical QA tasks, as well as in radiology report generation.

%% file: sections/3_data.tex
% ================================
\section{Data Curation}
\label{sec:data_curation}
\begin{figure*}[t]
    \centering
    \includegraphics[width=\textwidth]{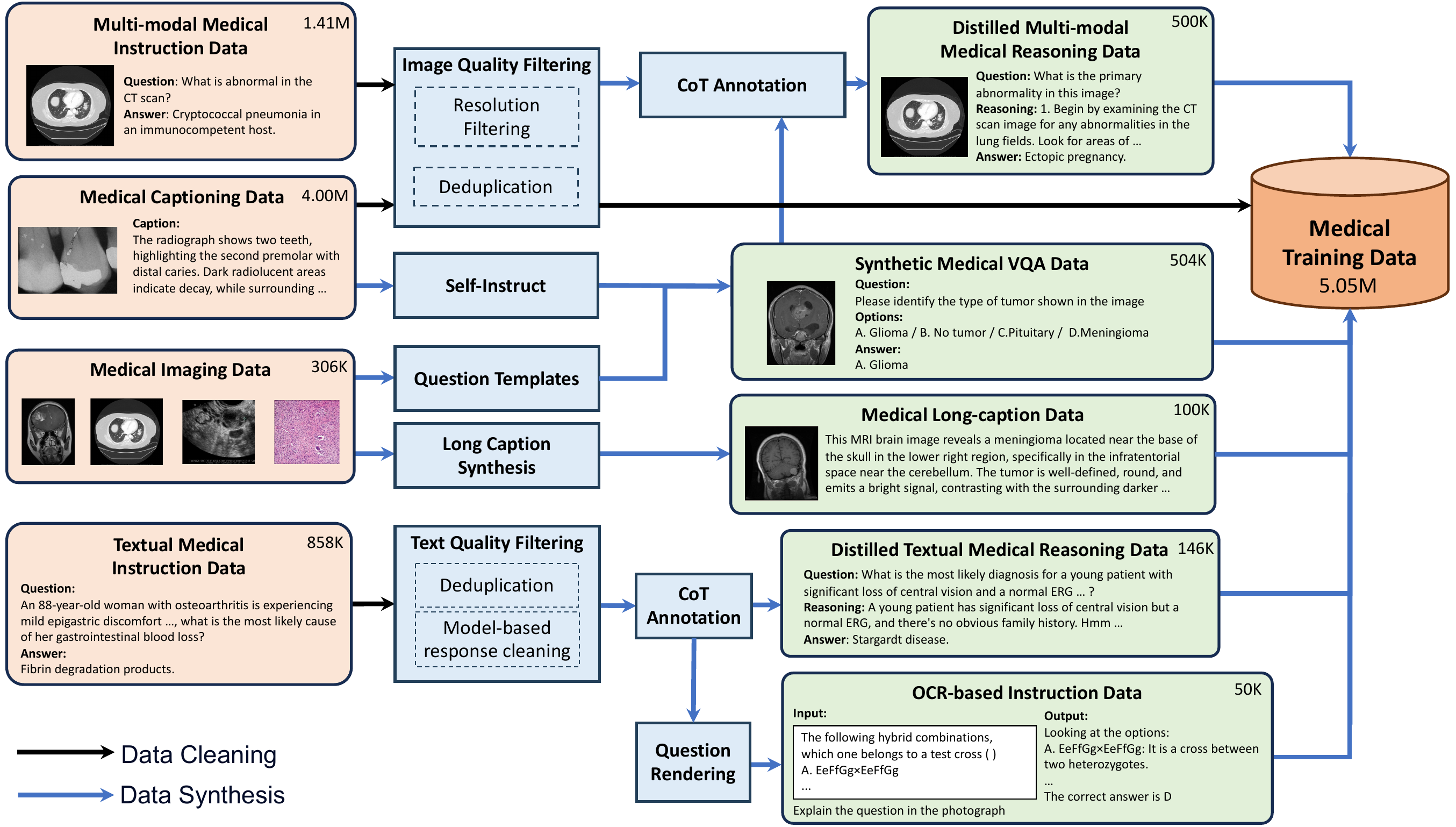}
    \caption{The overall data curation pipeline of medical multimodal and textual data.}
    \label{fig:overall-data-pipeline}
    \vspace{-0.15in}
\end{figure*}

% ================================
We address the intricate challenges of medical data curation via two primary strategies: 1) a comprehensive collection of diverse medical knowledge sources, including medical multimodal datasets, extensive medical texts, and general-domain multimodal and textual data; and 2) the synthesis of high-quality medical data through a robust pipeline capable of generating detailed captions, OCR-based samples, QA pairs, and chain-of-thought reasoning examples. The overall medical data curation pipeline is illustrated in Fig.~\ref{fig:overall-data-pipeline}.

For \textit{data collection}, we curate a wide range of open-source datasets from the web, categorized as follows: 1) \textbf{multimodal medical data} for enhancing \ours's comprehension of medical images and downstream task performance; 2) \textbf{unimodal medical data}, including standalone medical images and texts, to enrich domain knowledge and support synthetic data generation; 3) \textbf{general-domain data} to improve general visual and language understanding. All data undergo strict quality filtering before use. Each dataset is individually preprocessed and cleaned to ensure the quality and relevance.

For \textit{medical data synthesis}, we generate targeted synthetic data to strengthen specific model capabilities: 1) \textbf{long-form captions} to improve diagnostic detail extraction from images; 2) \textbf{OCR-based instruction data} to enhance text recognition in medical imagery; 3) \textbf{VQA samples} for tasks such as diagnosis, anatomical recognition, and modality classification; and 4) \textbf{distilled reasoning trajectories} to promote advanced medical reasoning. All synthesized data are subject to a rigorous quality control process to ensure accuracy and reliability.

% ================================
\subsection{Data Collection}
\label{ssec:data_collection}
% ================================

% ================================
\subsubsection{Medical Multimodal Data Collection}
\label{sssec:multimodal_data_collection}
% ================================
We incorporate two primary categories of existing multimodal medical data into our training set: 1) \textbf{medical caption data}, which pairs one or more medical images with descriptive text to facilitate semantic alignment between visual and textual modalities; and 2) \textbf{medical multimodal instruction data}, which includes image-instruction pairs. The latter comprises two main types: \textit{medical VQA data}, featuring open-ended or multiple-choice questions about medical images, and \textit{medical report data}, consisting of radiology reports that detail clinical findings and impressions. These datasets support MLLMs in learning a broad range of medical multimodal tasks.

% ================================
\subsubsection{Medical Unimodal Data Collection}
\label{sssec:unimodal_data_collection}
% ================================
\textbf{Medical text instruction data.}
We incorporate medical text instruction data into our training corpus to enhance the domain knowledge of \ours. The data spans four categories: 1) \textit{medical factoid QA}, including both free-form and multiple-choice questions on medical knowledge; 2) \textit{distilled reasoning data}, which supplements answers with step-by-step reasoning path annotated by large reasoning models; 3) \textit{patient-doctor dialogues}, comprising single-turn exchanges with patient queries and corresponding doctor responses; and 4) \textit{general medical instruction data}, covering diverse tasks such as medication summarization and medical text translation. 

\textbf{Medical imaging data.}
It comprises raw images paired with human-annotated metadata, including depicted organs, diagnostic labels, and imaging modalities. This metadata enables the synthesis of additional captioning and VQA samples, thereby improving medical image understanding and performance of downstream tasks. We primarily use these datasets for data synthesis (as detailed in \S\ref{ssec:med_data_syn}).

The complete lists of the collected datasets for medical multimodal and unimodal data are provided in Appendix \S\text{A1}.

% ================================
\subsubsection{General Domain Data Collection}
% ================================
To enhance the general multimodal understanding, generation, and instruction-following abilities of \ours, we incorporate diverse general-domain datasets, including image captioning, textual instruction-following, and multimodal instruction-following data, which collectively enhance the model's generalization to a wide range of vision-language tasks. Specifically, we utilize LLaVA-1.5 captions~\citep{liu2024llava15} and PixMo~\citep{deitke2024pixmo} for image captioning; LLaVA-1.5 (textual instruction subset) and OpenHermes-2.5\footnote{\url{https://huggingface.co/datasets/teknium/OpenHermes-2.5}} for textual instruction-following; and LLaVA-1.5 (multimodal instruction subset), along with ALLaVA~\citep{chen2024allava} for multimodal instruction-following.

% ================================
\subsubsection{Data Cleaning Process}
% ================================
\textbf{Data cleaning for medical multimodal data.}
Many open-source multimodal medical datasets are automatically extracted from scientific papers and thus often contain noise and redundancy. To ensure data quality, we implement a three-stage cleaning pipeline: 1) \textbf{image filtering}, where images with dimensions smaller than $64$ pixels are removed to eliminate low-resolution or uninformative content; 2) \textbf{image deduplication}, which applies perceptual hashing to remove exact duplicate images; 3) \textbf{text filtering}, which filters out lengthy and potentially noisy examples. Further details are provided in Appendix \S\text{A2}.

\textbf{Data cleaning for medical text data.}
To ensure the quality and compliance of the collected medical text instruction data, we implement two key cleaning procedures. First, we apply \textbf{model-based response cleaning} to remove sensitive identifying information and explicit prescriptions from the responses in patient-doctor dialogue datasets, thereby mitigating privacy and legal risks. Second, to eliminate redundancy, we perform \textbf{text deduplication} across multiple instruction datasets using MinHash locality-sensitive hashing (LSH). Implementation details are provided in Appendix~\S\text{A2}.

% ================================
\subsection{Medical Data Synthesis}
\label{ssec:med_data_syn}
% ================================
This section details our methodologies for synthesizing four types of medical multimodal data—long-form captions, OCR-based instruction samples, VQA instances, and distilled reasoning examples—alongside the quality control strategies employed to ensure their accuracy and reliability.

% ================================
\subsubsection{Medical Long-form Caption Synthesis}\label{sec:medical-long-caption-synthesis}
% ================================
Given the limited descriptive depth of existing medical captions, we construct enriched captions that capture key visual features of medical images. To this end, we utilize data from medical image segmentation and classification tasks, including human-annotated disease regions, diagnostic labels, and other visible metadata (\eg, gender and camera view). This factual knowledge enables the synthesis of authentic and detailed medical captions. To ensure the model's visual perception across imaging modalities, we source data from multiple modalities and balance their volume. Construction details are provided in \S\text{A6}. In the following, we show the key methods.

Specifically, for each image, we employ GPT-4o~\citep{openai2024gpt4o} to generate a long-form caption that captures fine-grained visual features. We devise the following five-stage pipeline to synthesize approximately 100K high-quality medical image captions.

\textbf{1) Metadata preparation:}  We leverage metadata to provide essential contextual information for a given image, including the imaging modality, type of disease, and other observable characteristics of the patient. Following Xie \etal~\citep{xie2025medtrinity}, we extract metadata from each dataset and apply dataset-specific rules to convert it into short captions. \textbf{2) RoI identification:} In addition to the textual metadata, we incorporate complementary visual information. For image segmentation datasets, they provide the location of the abnormality through RoIs, which are critical to reduce hallucinations and must be clearly marked in the images. We apply different ways to overlay RoIs on the source images. \textbf{3) Annotation with factual knowledge:} We prompt GPT-4o to synthesize image descriptions by integrating textual and visual information from the previous two stages. \textbf{4) Annotation with doctor preference:} Although the generated factual descriptions are highly reliable, they are often constrained by explicitly available metadata and may overlook critical visual details beyond RoIs. To mitigate this, we consult medical professionals to identify key factors they use to interpret medical images. Their insights are distilled into task-specific instructions for prompt GPT-4o. \textbf{5) Summarization:} We use GPT-4o to combine the outputs of stages 3 and 4 to consolidate the final caption for each medical image. To avoid conflict, the description from stage 3 is prioritized due to its higher reliability.

\textbf{Quality control:} To ensure caption quality, we implement an iterative validation process with medical professionals. We progressively annotate small batches of samples of all modalities and have doctors review them for accuracy and clinical relevance. Based on their feedback, we refine our annotation pipeline and prompts. Only after obtaining approval from medical experts on the quality of these pilot annotations do we proceed with large-scale captioning. This iterative approach ensures that our final captions meet clinical standards before scaling up the annotation process. Appendix~\S\text{A4} presents an example of a synthesized long-form caption for an MRI image.

% ================================
\subsubsection{Medical OCR-based Instruction Data Synthesis}
% ================================
OCR-based instruction datasets enhance the model's ability to interpret textual information embedded within medical images, a common scenario involving elements such as patient identifiers, age, and biological metrics. We collect biology and chemistry exam questions with ground-truth answers from online sources, as these subjects are closely related to the medical domain and offer rich resources. Each question is annotated with detailed reasoning steps using Gemini-2.0-Flash-Thinking~\citep{gemini2025gemini}, and only samples with answers that exactly match the ground truth are retained. The validated questions are then rendered as images to form multimodal instruction data, where the question image serves as input and the textual answer as output (see Table V in Appendix \S\text{A4}). This process yields a total of 50K high-quality OCR-based samples.

% ================================
\subsubsection{Medical VQA Data Synthesis}
\label{sec:MCQA-data-synthesis}
% ================================
Medical diagnosis, anatomy identification, and modality classification are core competencies required for MLLM to perform various multimodal medical tasks. To strengthen \ours's capabilities in these areas, we synthesize additional medical VQA data using two complementary strategies: a template-based method and a self-instruct-based method.

\textbf{Template-based method.}
This method constructs VQA samples using human-annotated meta-information, \eg, abnormality labels, from public medical imaging datasets. Specifically, we extract datasets containing labels related to anatomical structures, abnormalities, or imaging modalities. For each label type, we manually design question templates. Ground-truth answers are derived from the original annotations, while distractor options are sampled from the label space of the respective dataset. This approach enables the structured generation of high-quality VQA data.

\textbf{Self-instruct-based method.}
To enhance question diversity and linguistic variation, we leverage GPT-4o to generate VQA samples from \texttt{(image, caption)} pairs sourced from medical captioning datasets. GPT-4o is prompted using a few-shot format, where seed examples—drawn from open-source medical VQA datasets—demonstrate questions focused on diagnosis and anatomy. It then produces a question, options, and the correct answer based on the provided caption to enable more varied and natural question formulations.

The overall pipeline of both template-based and self-instruct-based methods and the prompt design of the self-instruct-based method are detailed in Appendix \S\text{A4}. 

\begin{figure*}[t]
    \centering
    \includegraphics[width=\textwidth]{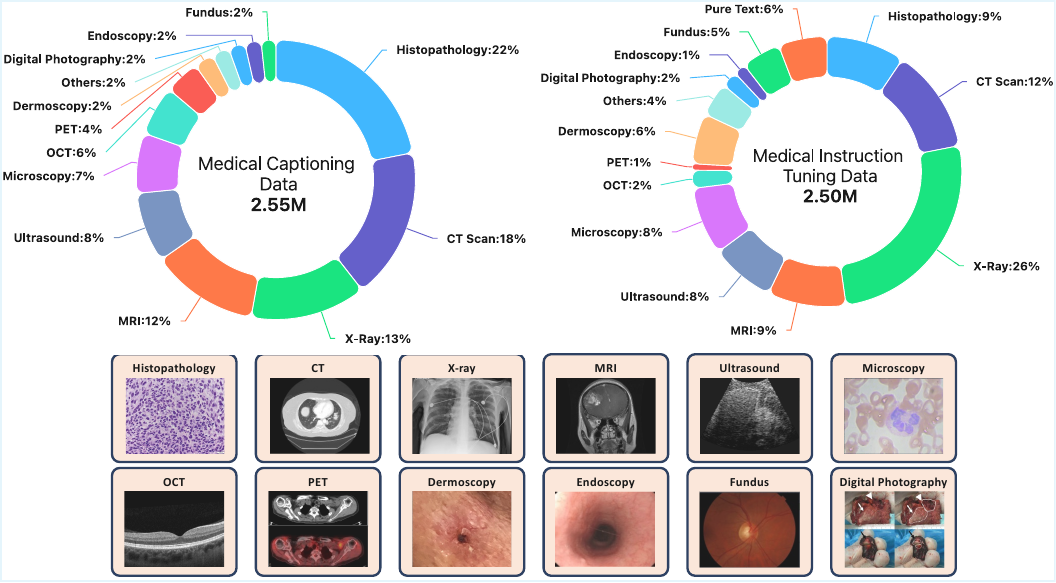}
    \caption{The distribution of medical image modality in our training data (medical subset).
    }
    \label{fig:modality-distribution}
    \vspace{-0.15in}
\end{figure*}

\subsubsection{Medical Reasoning Data Distillation}
Most existing open-source and synthesized medical data, primarily composed of short-answer and multiple-choice questions, lack explicit annotations of the underlying reasoning process. To strengthen the model's medical reasoning capabilities, we employ GPT-4o to generate chain-of-thought (CoT) reasoning trajectories for a subset of both multimodal and textual instruction data. Specifically, for each sample, we provide GPT-4o with the question, ground-truth answer, and, if applicable, the answer options, and the corresponding medical image. GPT-4o is instructed to produce a step-by-step reasoning path without relying on or explicitly referencing the ground-truth answer. To ensure quality, we implement an LLM-based validation process wherein GPT-4o evaluates the consistency between the reasoning trace and the ground-truth answer. Samples deemed inconsistent are excluded from the final dataset. The prompts are detailed in Appendix~\S\text{A5}.

\subsection{Dataset Summary}
The data collection and synthesis pipelines yield \texttt{3.75M} open-source and \texttt{1.30M} synthetic medical samples, respectively. We also conduct rigorous quality checks on all samples to ensure overall data quality. \textbf{To prevent data contamination, we perform strict image and text deduplication by excluding images and samples that  overlap with the evaluation benchmarks in \oureval}. Moreover, we examine the distribution of medical imaging modalities within the training set. As many multimodal datasets lack explicit modality annotations, we use a BiomedCLIP-based classifier~\citep{zhang2025biomedclip} to infer the modalities of unlabeled samples. Fig.~\ref{fig:modality-distribution} presents the resulting modality distribution. Our dataset covers over \texttt{12} medical imaging modalities, substantially enhancing the model's versatility and applicability on medical downstream tasks across a wide spectrum of modalities.

%% file: sections/4_training.tex
% ================================
\section{Model Training}
\label{sec:training}
% ================================
\ours is built upon the Qwen2.5-VL~\citep{bai2025qwen25vl}, consisting of three key components: an LLM, a vision encoder, and an MLP-based projector. We employ two parameter scales of Qwen2.5-VL, \ie, 7B and 32B, as the foundational models. 

\begin{figure*}[t]
    \centering
    \includegraphics[width=0.8\linewidth]{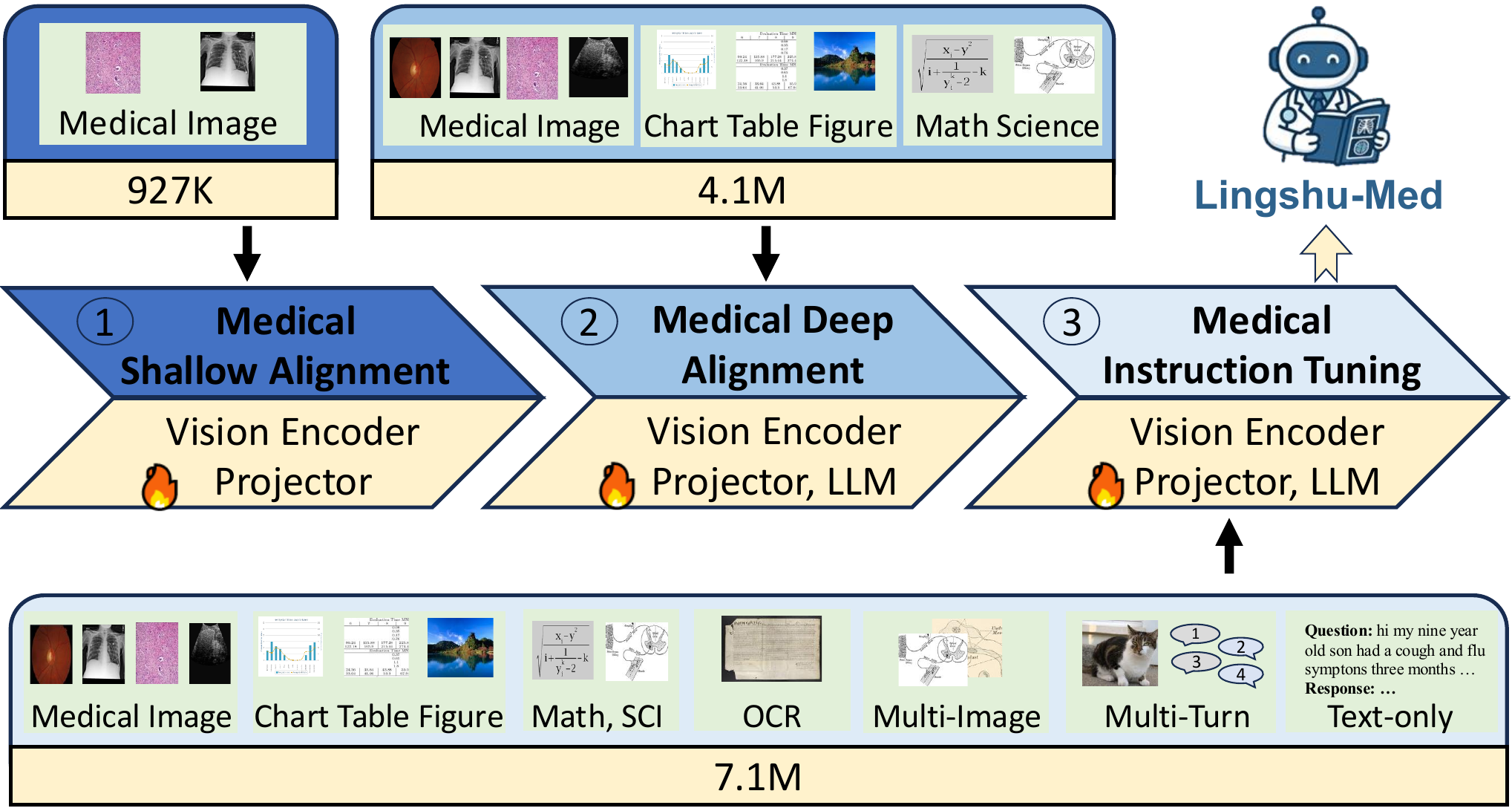}
    \caption{The training pipeline of \ours: medical shallow alignment, medical deep alignment, and medical instruction tuning.}
    \label{fig:training_recipe}
    \vspace{-0.1in}
\end{figure*}

Building on prior MLLM advances~\citep{bai2025qwen25vl,zhu2025internvl3}, we develop a three-stage training framework following a ``shallow-to-deep'' progression: 1) \textbf{Medical Shallow Alignment}, 2) \textbf{Medical Deep Alignment}, and 3) \textbf{Medical Instruction Tuning}, as shown in Fig.~\ref{fig:training_recipe}. The first two stages build a solid medical vision-language foundation. Medical Shallow Alignment utilizes a small set of medical captions for basic visual understanding. Medical Deep Alignment uses larger and semantically richer datasets to deepen domain knowledge and achieve finer vision-language alignment. The later stages enhance practical utility. Medical Instruction Tuning improves task-specific instruction following across medical use cases.

% ================================
\subsection{Medical Shallow Alignment}
\label{ssec:alignment_training}
% ================================
This stage aims to establish an effective alignment between diverse medical imaging modalities and their corresponding descriptions. This alignment enhances the model's ability to comprehend and interpret medical images, thereby laying a solid foundation for subsequent integration of medical knowledge. During this phase, the LLM is kept frozen, while only the vision encoder and the projector are trained using coarsely annotated medical image-caption data. We deliberately freeze the LLM because the limited semantic complexity of coarse caption data could potentially compromise the language generation capabilities of the LLM. Specifically, the coarse caption data comprise two datasets with relatively short captions, \ie, PMC-OA~\citep{liu2023pmcclip} and ROCO~\citep{pelka2018roco}.

% ================================
\subsection{Medical Deep Alignment}
\label{ssec:pretraining}
% ================================
The goal of medical deep alignment is to comprehensively integrate medical knowledge into the MLLM, enhancing its ability to understand diverse medical concepts and adapt to various clinical contexts. To achieve this, all model parameters are unfrozen to allow end-to-end fine-tuning. Compared to the preceding shallow alignment stage, the dataset in this stage is significantly expanded in terms of modality diversity, linguistic complexity, and structural completeness. It includes images from less common medical modalities, longer and more structured captions, as well as image-caption pairs synthesized from medical image classification and segmentation tasks. Moreover, high-quality general-domain multimodal image-text data are incorporated into the training process alongside medical data. This joint training not only helps preserve the MLLM's general multimodal capabilities but also introduces it to a wider variety of visual formats, such as charts, tables, graphs, mathematics, and scientific illustrations.

% ================================
\subsection{Medical Instruction Tuning}
\label{ssec:sft}
% ================================
Following standard practices~\citep{li2023llavamed,chen2024huatuogptvision,chen2025internvl25}, we fine-tune the MLLM for instruction following across various clinical tasks. To accommodate a wide breadth of clinical scenarios, we unlock all parameters for training and leverage a comprehensive training corpus that goes beyond basic formats like image captions and QA pairs. Our data include comprehensive medical scenarios spanning diagnosis, clinical examination, knowledge retrieval, report generation, and anatomical localization. We also add carefully vetted general-domain and medical text data to reduce image-centric bias and to improve overall medical understanding. The corpus also includes complex formats such as multi-image reasoning, multi-turn dialogues, and detailed reasoning tasks, enabling the MLLM to handle clinical scenarios requiring greater analytical capabilities.

%% file: sections/5_evaluation.tex
% ================================
\section{Unified Medical Evaluation Framework}
\label{sec:eval_framework}
% ================================

\begin{figure*}[t]
    \centering
    \begin{subfigure}[t]{0.34\textwidth}
        \centering
        \includegraphics[width=0.95\textwidth]{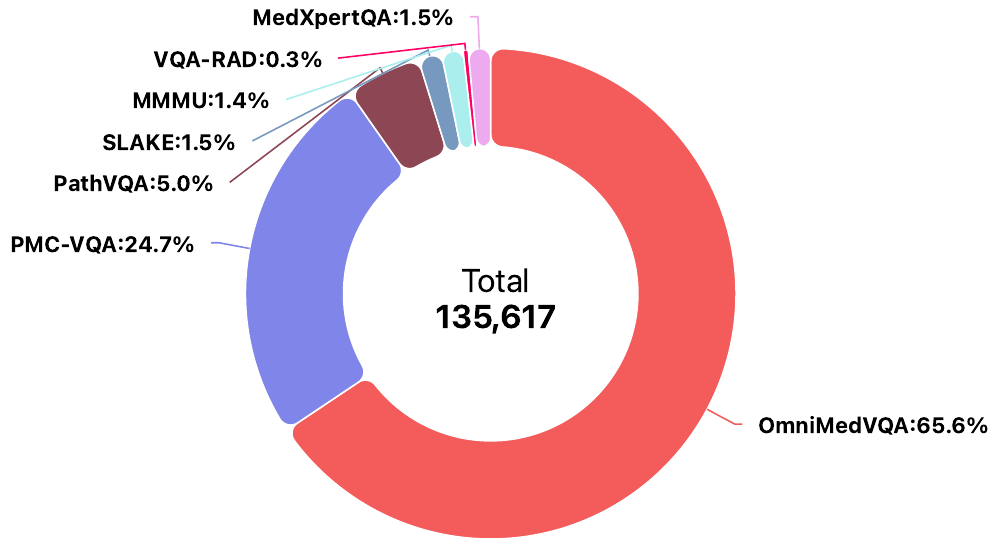}
        \caption{Medical Multimodal Benchmark}
        \label{fig:benchmark_multimodal}
    \end{subfigure}
    \hfill
    \begin{subfigure}[t]{0.32\textwidth}
        \centering
        \includegraphics[width=0.95\textwidth]{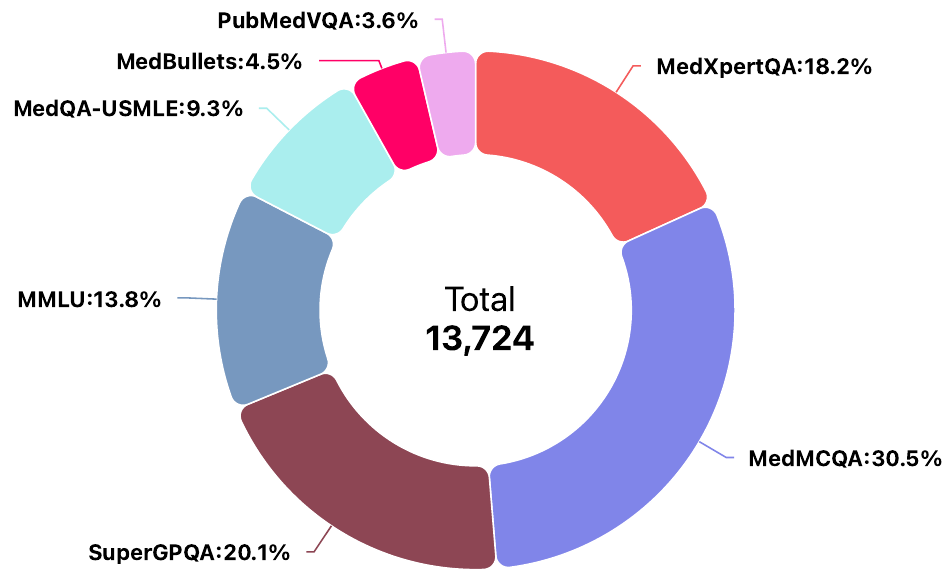}
        \caption{Medical Text-only Benchmark}
        \label{fig:benchmark_text}
    \end{subfigure}
    \hfill
    \begin{subfigure}[t]{0.28\textwidth}
        \centering
        \includegraphics[width=0.95\textwidth]{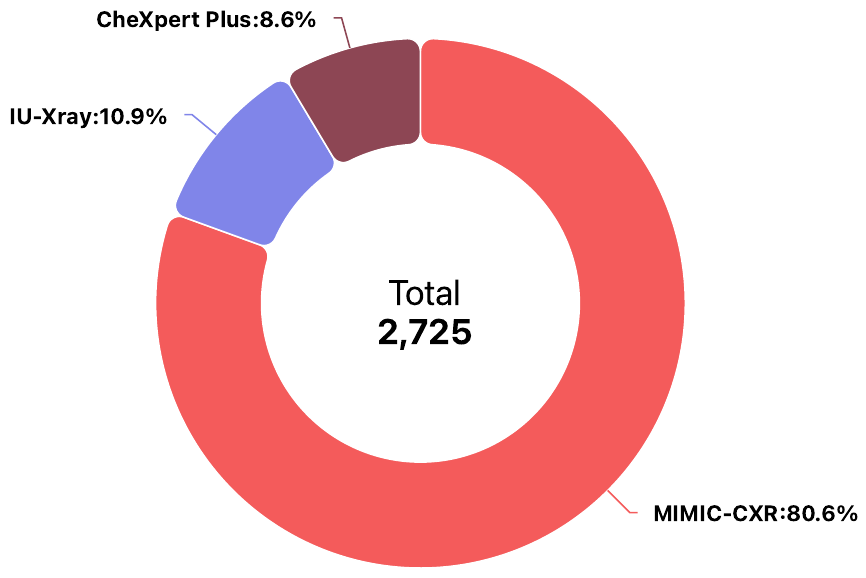}
        \caption{Report Gen. Benchmark}
        \label{fig:benchmark_report}
    \end{subfigure}
    \caption{An overview of the data distribution across benchmarks for different types of medical tasks.}
    \label{fig:benchmarks_overview}
    \vspace{-0.1in}
\end{figure*}

Rapid progress of medical MLLMs has yielded impressive performance across various tasks. However, the absence of a unified evaluation framework has led to inconsistencies in performance comparison across models. While existing frameworks like LMMs-Eval~\citep{lmms_eval2024}, Eval-Harness,\footnote{\url{https://zenodo.org/records/12608602}} and VLMEvalKit~\citep{duan2024vlmevalkit} can be applied to evaluate medical MLLMs, they are primarily designed for general tasks. Thus, they exhibit limitations in medical data coverage and model adaptation, making them difficult to apply directly.

To comprehensively evaluate the performance of medical MLLMs, we develop a systematic evaluation framework, dubbed \oureval, which integrates mainstream medical benchmarks and task types and supports a variety of question formats, including multiple-choice, closed-ended, open-ended questions, as well as medical report generation. It covers representative tasks in medical multimodal and textual understanding, thereby offering broad applicability. \oureval accommodates both multimodal and text-only inputs, enabling a unified evaluation of both medical MLLMs and LLMs. To enhance reproducibility, we standardize data preprocessing formats and post-processing protocols. \oureval also incorporates a dual verification mechanism that combines rule-based evaluation with the LLM-as-a-Judge strategy, integrating both objective and subjective assessments to improve evaluation stability and reliability of the evaluation. Additionally, \oureval supports inference acceleration via vLLM~\citep{kwon2023efficient}, enabling high-throughput and parallel evaluation, and offering strong scalability and engineering usability.

% ================================
\subsection{A Large-scale Medical Evaluation Benchmark}
% ================================
We curate a collection of mainstream medical benchmarks covering multimodal QA, text-only QA, and report generation, comprising a total of \texttt{152,066} samples with \texttt{121,622} distinct medical images spanning \texttt{16} benchmark datasets. An overview of the medical benchmark is depicted in Fig.~\ref{fig:benchmarks_overview}. Moreover, we standardize the input format for all questions based on their types, adhering to the official chat template of  medical MLLM candidates to be evaluated. More details of the benchmark and prompt format are in Appendix \S\text{D}.

\textbf{Multimodal QA} contains the test sets from VQA-RAD~\citep{lau2018vqarad}, SLAKE~\citep{liu2021slake}, PathVQA~\citep{he2020pathvqa}, PMC-VQA (v2; \citep{zhang2024pmcvqa}), OmniMedVQA~\citep{hu2024omnimedvqa}, MMMU~\citep{yue2024mmmu}, and MedXpertQA~\citep{zuo2025medxpertqa}. Specifically, we select the Health\&Medical subset of MMMU, multimodal set of MedXpertQA, and open-access part of OmniMedVQA. By incorporating these datasets, \oureval encompasses various medical imaging modalities, including X-ray, CT, MRI, PET, ultrasound, microscopy, pathology, OCT, dermoscopy, gastrointestinal, endoscopy, fundus, and medical-related charts, tables, and figures.

\textbf{Text-only QA} covers multiple medical text benchmarks, including the test sets of MedQA-USMLE (\citep{jin2020medqausmle}), MMLU-Med~\citep{hendrycks2021mmlu}, PubMedQA~\citep{jin2019pubmedqa}, MedMCQA~\citep{pal2022medmcqa}, MedBullets~\citep{chen2025medbullets}, MedXpertQA~\citep{zuo2025medxpertqa}, and SuperGPQA~\citep{pteam2025supergpqa}. We select the text portion of MedXpertQA, the medicine-related questions of MMLU by following the split in Wang \etal~\citep{wang2024apollo}, and the test set of SuperGPQA by the official split criteria.

\textbf{Report Generation} contains the official test splits of MIMIC-CXR,\footnote{\url{https://physionet.org/content/mimic-cxr/2.1.0/}} IU-Xray,\footnote{\url{https://openi.nlm.nih.gov/faq?download=true}} and the validation set of CheXpert Plus,\footnote{\url{https://aimi.stanford.edu/datasets/chexpert-plus}} which consist of chest X-rays paired with corresponding radiology reports. We follow Zhang~\etal~\citep{zhang2024rexrank} to filter out samples where both findings and impressions are empty.

% ================================
\subsection{Evaluation Metrics}
% ================================
We use \textbf{Accuracy} as the primary metric for the QA task. For \textit{multiple-choice questions}, a two-stage strategy is adopted: rule-based matching is first applied to assess alignment with answer options; if this is insufficient, we further employ the official MMMU codebase\footnote{\url{https://github.com/MMMU-Benchmark/MMMU}} to compute similarity scores and select the option with the highest score as the final answer. For \textit{open-ended questions}, we directly leverage GPT-4.1\footnote{\url{https://openai.com/index/gpt-4-1/}} to assess consistency between model outputs and reference answers, with the prompting strategy detailed in Appendix \S\text{D}. As GPT-4.1 also appears as a compared model in our benchmark tables, the LLM judge receives only the question, the reference answer, and the candidate output, with all model identities withheld. Candidate outputs are normalized to a consistent textual format (\eg, the choice index ``C'' for multi-choice QA) before judging. It is therefore blind to which model produced each answer.  To verify the reliability of the automatic judge, we conducted a human spot-check on 100 randomly sampled responses, on which the GPT-4.1 verdicts agreed with human
annotators in 98\% of cases.

For the report generation task, we develop a multi-metric evaluation approach that integrates semantic and model-based metrics, specifically tailored to address the challenges posed by the lengthy and highly open-ended nature of the outputs. This design enables a more comprehensive and granular assessment of the performance of model . For the semantic evaluation metrics, we choose \textbf{Rouge-L}~\citep{lin2004rouge} and \textbf{CIDEr}~\citep{vedantam2015cider} to measure the quality of model-generated reports with respect to reference answers. For the model-based evaluation metrics, we follow ReXrank~\citep{zhang2024rexrank} and utilize  \textbf{SembScore}~\citep{smit2020chexbert} and \textbf{RaTEScore}~\citep{zhao2024ratescore} to measure the candidate reports. We also include a composite metric \textbf{RadCliQ}$_{v1}$~\citep{yu2023evaluating}. As lower \textbf{RadCliQ}$_{v1}$ scores indicate better performance, we follow Zhang \etal~\citep{zhang2024rexrank} to take its reciprocal, \ie, \textbf{RadCliQ}$_{v1}^{-1}$. Detailed descriptions of these metrics are presented in Appendix \S\text{D}. We note that automatic report-generation metrics primarily capture surface-level overlap and entity matching, and may not fully reflect factual correctness. We therefore complement them with the physician study in \S V-F, where the Safety, Completeness, and Clinical Relevance dimensions directly assess hallucinated content, clinically significant omissions, and factual correctness of the generated reports.

%% file: sections/6_experiments.tex
% ================================
\section{Experiments}
\label{sec:experiments}
% ================================

% ================================
\subsection{Experimental Settings}
\label{ssec:exp_setup}
% ================================
In the main paper, we primarily implement \ours pipeline using Qwen2.5-VL. To demonstrate its generalizability, we also evaluate the pipeline on the recent InternVL3 model, with detailed results provided in Appendix \S E. To comprehensively evaluate \ours on various medical benchmarks, we compare it against a diverse set of baselines, encompassing proprietary and open-source MLLMs from both the general and medical domains. The open-source MLLMs are further categorized based on their parameter sizes. Specifically, our evaluation includes the following models:

\noindent\textbf{Proprietary Models}: GPT-4.1\footnote{\url{https://openai.com/index/gpt-4-1/}} (\texttt{2025-04-14}), Claude Sonnet 4\footnote{\url{https://www.anthropic.com/news/claude-4}}, and Gemini-2.5-Flash (\texttt{Preview 05-20}; \citep{gemini2025gemini}), which are the most representative proprietary models available.

\noindent\textbf{Medical-Specific MLLMs}: BiomedGPT~\citep{zhang2024biomedgpt}, Med-R1~\citep{lai2025medr1}, MedVLM-R1~\citep{pan2025medvlmr1}, HuatuoGPT-V~\citep{chen2024huatuogptvision}, MedGemma~\citep{sellergren2025medgemma}, HealthGPT~\citep{lin2025healthgpt}, LLaVA-Med~\citep{li2023llavamed}, BioMediX2~\citep{mullappilly2024bimedix2}, and MedDr~\citep{he2024gsco}.

\noindent\textbf{General-Purpose MLLMs}: Qwen2.5-VL-Instruct~\citep{bai2025qwen25vl}, InternVL2.5~\citep{chen2025internvl25}, VideoLLaMA3~\citep{zhang2025videollama}, and InternVL3~\citep{zhu2025internvl3}.\footnote{Although categorized as general-purpose MLLMs, InternVL series incorporates massive medical multimodal data into its training corpus~\citep{li2025gmaivl,chen2025internvl25,zhu2025internvl3}.}

% ================================
\subsection{Performance on Medical Multimodal Benchmarks}
\label{ssec:med_multimodal_results}
% ================================

\begin{table*}[t]
\centering
\small
\caption{Performance comparison of  MLLMs on medical multimodal benchmarks, where \textbf{bold} and \underline{underline} scores indicate the best and the second-best method, respectively. $^\diamondsuit$Med-R1 is trained on the OmniMedVQA test set, making its results on OmniMedVQA meaningless. $^\heartsuit$BiomedGPT and MedDr lack multi-image support needed for MedXpertQA. Models colored in \colorbox[HTML]{F5F5F5}{light gray}, \colorbox[HTML]{EBEBEB}{gray} and \colorbox[HTML]{E8F5E9}{green} denote medical MLLMs, general-purpose MLLMs and \ours, respectively. }
\adjustbox{max width=\textwidth}{
\begin{tabular}{lcccccccc}
\toprule
\textbf{Models} & \textbf{MMMU-Med} & \textbf{VQA-RAD} & \textbf{SLAKE} & \textbf{PathVQA} & \textbf{PMC-VQA} & \textbf{OmniMedVQA} & \textbf{MedXpertQA} & \textbf{Avg.}\\
\midrule
\multicolumn{9}{c}{\texttt{Proprietary Models}}\\
\midrule
GPT-4.1 & 75.2  & 65.0 & 72.2 & 55.5 & 55.2 & 75.5 & 45.2 & 63.4\\
Claude Sonnet 4  & 74.6 & 67.6 & 70.6 & 54.2 & 54.4 & 65.5 & 43.3 & 61.5\\
Gemini-2.5-Flash & 76.9 & 68.5 & 75.8 & 55.4 & 55.4 & 71.0 & 52.8 & 65.1\\
\midrule
\multicolumn{9}{c}{\texttt{Open-source Models (<10B)}}\\
\midrule
\rowcolor[HTML]{F5F5F5} BiomedGPT$^\heartsuit$ & 24.9 & 16.6 & 13.6 & 11.3 & 27.6 & 27.9 & - & - \\
\rowcolor[HTML]{F5F5F5} Med-R1-2B$^\diamondsuit$ & 34.8 & 39.0 & 54.5& 15.3 & 47.4 & - & 21.1 & -\\
\rowcolor[HTML]{F5F5F5} MedVLM-R1-2B & 35.2 & 48.6 & 56.0 & 32.5 & 47.6 & 77.7 & 20.4 & 45.4\\
\rowcolor[HTML]{F5F5F5} MedGemma-4B-IT & 43.7 & \textbf{72.5} & \underline{76.4} & \underline{48.8} & 49.9 & 69.8 & 22.3 & 54.8 \\
\rowcolor[HTML]{F5F5F5} LLaVA-Med-7B & 29.3 & 53.7 & 48.0 & 38.8 & 30.5 & 44.3 & 20.3 & 37.8 \\
\rowcolor[HTML]{F5F5F5} HuatuoGPT-V-7B & 47.3 & {67.0} & 67.8 & 48.0 & 53.3 & 74.2 & 21.6 & 54.2\\
\rowcolor[HTML]{F5F5F5} BioMediX2-8B & 39.8 & 49.2 & 57.7 & 37.0 & 43.5 & 63.3 & 21.8 & 44.6\\
\rowcolor[HTML]{EBEBEB} Qwen2.5-VL-7B & 50.6 & 64.5 & 67.2 & 44.1 & 51.9 & 63.6 & 22.3 & 52.0\\
\rowcolor[HTML]{EBEBEB} VideoLLaMA3-7B & 46. & 1.4 & 64.1 & 56.0 & \textbf{56.8} & 75.5 & 21.9 & 54.5 \\
\rowcolor[HTML]{EBEBEB} InternVL2.5-8B & 53.5 & 59.4 & 69.0 & 42.1 & 51.3 & \underline{81.3} & 21.7 & 54.0\\
\rowcolor[HTML]{EBEBEB} InternVL3-8B & \textbf{59.2} & 65.4 & {72.8} & {48.6} & {53.8} & 79.1 & \underline{22.4} & \underline{57.3} \\
\hline
\rowcolor[HTML]{E8F5E9} \ours-7B & \underline{54.0} & \underline{67.9} & \textbf{83.1} & \textbf{61.9} & \underline{56.3} & \textbf{82.9} & \textbf{26.7} & \textbf{61.8}\\
\midrule
\multicolumn{9}{c}{\texttt{Open-source Models (>10B)}}\\
\midrule
\rowcolor[HTML]{F5F5F5} HealthGPT-14B & 49.6 & 65.0 & 66.1 & \underline{56.7} & 56.4 & 75.2 & 24.7 & 56.2\\
\rowcolor[HTML]{F5F5F5} HuatuoGPT-V-34B & 51.8 & 61.4 & 69.5 & 44.4 & 56.6 & 74.0 & 22.1 & 54.3\\
\rowcolor[HTML]{F5F5F5} MedDr-40B$^\heartsuit$       & 49.3 & 65.2 & 66.4 & 53.5 & 13.9 & 64.3 & - & - \\
\rowcolor[HTML]{EBEBEB} InternVL3-14B & \underline{63.1} & 66.3 & \underline{72.8} & 48.0 & 54.1 & 78.9 & 23.1 & 58.0\\
\rowcolor[HTML]{EBEBEB} Qwen2.5V-32B & 59.6 & \underline{71.8} & 71.2 & 41.9 & 54.5 & 68.2 & 25.2 & 56.1\\
\rowcolor[HTML]{EBEBEB} InternVL2.5-38B & 61.6 & 61.4 & 70.3 & 46.9 & \underline{57.2} & \underline{79.9} & 24.4 & 57.4\\
\rowcolor[HTML]{EBEBEB} InternVL3-38B & \textbf{65.2} & 65.4 & {72.7} & 51.0 & 56.6 & 79.8 & \underline{25.2} & \underline{59.4}\\

\hline
\rowcolor[HTML]{E8F5E9} \ours-32B & 62.3 & \textbf{76.5} & \textbf{89.2} & \textbf{65.9} & \textbf{57.9} & \textbf{83.4} & \textbf{30.9} & \textbf{66.6}\\
\bottomrule
\end{tabular}
}
\label{tab:mm_result}
\end{table*}

Table~\ref{tab:mm_result} presents a detailed comparison between \ours and diverse MLLMs across seven medical multimodal benchmarks. Proprietary models such as GPT-4.1, Claude Sonnet 4, and Gemini-2.5-Flash demonstrate consistently strong performance, with Gemini-2.5-Flash attaining the highest average score ($65.1$) among them. Among open-source models with $<$10B parameters, \ours-7B achieves the highest average score ($61.8$), exceeding the best-performing baseline in this category by a substantial margin ($+4.5$). It ranks first on five out of seven benchmarks—SLAKE ($83.1$), PathVQA ($61.9$), PMC-VQA ($56.3$), OmniMedVQA ($82.9$), and MedXpertQA ($26.7$)—and secures second place on the remaining two, \ie, MMMU-Med and VQA-RAD. On PathVQA, \ours-7B outperforms the next-best open-source model, MedGemma-4B-IT, by a wide margin ($61.9$ versus $48.8$). Furthermore, \ours-7B achieves an average performance comparable to GPT-4.1 and Claude Sonnet 4, highlighting its competitiveness despite a significantly smaller parameter size. With increased model capacity, \ours-32B attains a state-of-the-art average score of $66.6$ across all benchmarks, outperforming both proprietary and open-source counterparts. These results validate the effectiveness of our approach and demonstrate its strong potential for broad application across diverse medical multimodal tasks.

% ================================
\subsection{Performance on Medical Textual Benchmarks}
\label{ssec:med_textual_results}
% ================================

\begin{table*}[t]
\centering
\small
\caption{Performance of \ours and other MLLMs on medical text benchmarks. Models colored in \colorbox[HTML]{F5F5F5}{light gray}, \colorbox[HTML]{EBEBEB}{gray} and \colorbox[HTML]{E8F5E9}{green} denote medical MLLMs, general-purpose MLLMs and \ours, respectively.}
\adjustbox{max width=\textwidth}{
\begin{tabular}{lcccccccccc}
\toprule
\textbf{Models} & \textbf{MMLU-Med} & \textbf{PubMedQA} & \textbf{MedMCQA} & \textbf{MedQA-USMLE} & \textbf{Medbullets} & \textbf{MedXpertQA} & \textbf{SuperGPQA} & \textbf{Avg.}\\
\midrule
\midrule
GPT-4.1          & 89.6 & 75.6 & 77.7 & 89.1 & 77.0 & 30.9 & 49.9 & 70.0 \\
Claude Sonnet 4  & 91.3 & 78.6 & 79.3 & 92.1 & 80.2 & 33.6 & 56.3 & 73.1 \\
Gemini-2.5-Flash & 84.2 & 73.8 & 73.6 & 91.2 & 77.6 & 35.6 & 53.3 & 69.9 \\
\midrule
\multicolumn{9}{c}{\texttt{Open-source Models (<10B)}}\\
\midrule
\rowcolor[HTML]{F5F5F5} Med-R1-2B & 51.5 & 66.2 & 39.1 & 39.9 & 33.6 & 11.2 & 17.9 & 37.0\\
\rowcolor[HTML]{F5F5F5} MedVLM-R1-2B & 51.8 & 66.4 & 39.7 & 42.3 & 33.8 & 11.8 & 19.1 & 37.8 \\
\rowcolor[HTML]{F5F5F5} MedGemma-4B-IT & 66.7 & 72.2 & 52.2 & 56.2 & 45.6 & 12.8 & 21.6 & 46.8\\
\rowcolor[HTML]{F5F5F5} LLaVA-Med-7B & 50.6 & 26.4 & 39.4 & 42.0 & 34.4 & 9.9 & 16.1 & 31.3\\
\rowcolor[HTML]{F5F5F5} HuatuoGPT-V-7B & 69.3 & 72.8 & 51.2 & 52.9 & 40.9 & 10.1 & 21.9 & 45.6 \\
\rowcolor[HTML]{F5F5F5} BioMediX2-8B & 68.6 & 75.2 & 52.9 & 58.9 & 45.9 & 13.4 & 25.2 & 48.6 \\
\rowcolor[HTML]{EBEBEB} Qwen2.5-VL-7B & 73.4 & {76.4} & 52.6 & 57.3 & 42.1 & 12.8 & 26.3 & 48.7 \\
\rowcolor[HTML]{EBEBEB} VideoLLaMA3-7B & 70.1 & \textbf{76.8} & 49.7 & 56.5 & 42.0 & 12.7 & \underline{29.5} & 48.2 \\
\rowcolor[HTML]{EBEBEB} InternVL2.5-8B & 74.2 & 76.4 & 52.4 & 53.7 & 42.4 & 11.6 & 26.1 & 48.1 \\
\rowcolor[HTML]{EBEBEB} InternVL3-8B & \textbf{77.5} & 75.4 & \textbf{57.7} & \underline{62.1} & \underline{48.5} & \underline{13.1} & \textbf{31.2} & \underline{52.2} \\
\hline
\rowcolor[HTML]{E8F5E9} \ours-7B & \underline{74.5} & \underline{76.6} & \underline{55.9} & \textbf{63.3} & \textbf{56.2} & \textbf{16.5} & {26.3} & \textbf{52.8} \\
\midrule
\multicolumn{9}{c}{\texttt{Open-source Models (>10B)}}\\
\midrule
\rowcolor[HTML]{F5F5F5} HealthGPT-14B & 80.2 & 68.0 & 63.4 & 66.2 & 39.8 & 11.3 & 25.7 & 50.7 \\
\rowcolor[HTML]{F5F5F5} HuatuoGPT-V-34B & 74.7 & 72.2 & 54.7 & 58.8 & 42.7 & 11.4 & 26.5 & 48.7 \\
\rowcolor[HTML]{F5F5F5} MedDr-40B       & 65.2 & \underline{77.4} & 38.4 & 59.2 & 44.3 & 12.0 & 24.0 & 45.8\\
\rowcolor[HTML]{EBEBEB} InternVL3-14B & 81.7 & 77.2 & 62.0 & 70.1 & 49.5 & 14.1 & 37.9 & 56.1 \\
\rowcolor[HTML]{EBEBEB} Qwen2.5-VL-32B & 83.2 & 68.4 & 63.0 & 71.6 & 54.2 & 15.6 & 37.6 & 56.2 \\
\rowcolor[HTML]{EBEBEB} InternVL2.5-38B & \underline{84.6} & 74.2 & \underline{65.9} & \underline{74.4} & \underline{55.0} & 14.7 & 39.9 & 58.4 \\
\rowcolor[HTML]{EBEBEB} InternVL3-38B & 83.8 & 73.2 & 64.9 & 73.5 & 54.6 & \underline{16.0} & \textbf{42.5} & \underline{58.4} \\

\hline
\rowcolor[HTML]{E8F5E9} \ours-32B & \textbf{84.7} & \textbf{77.8} & \textbf{66.1} & \textbf{74.7} & \textbf{65.4} & \textbf{22.7} & \underline{41.1} & \textbf{61.8}\\
\bottomrule
\end{tabular}
}
\label{tab:text_result}
\end{table*}

Although primarily designed as multimodal models, \ours exhibit remarkable proficiency in handling medical text-based tasks, underscoring their generalization capabilities across data modalities. Reported in Table~\ref{tab:text_result}, \ours-7B achieves the highest average accuracy among all open source models with parameters $<$ 10B, leading in four key benchmarks: PubMedQA ($76.6$), MedQA-USMLE ($63.3$), Medbullets ($56.2$) and MedXpertQA ($16.5$) and ranks a close second on the remaining three benchmarks. For larger-scale models, \ours-32B delivers more pronounced improvements. It outperforms the second-best open-source competitor, InternVL3-38B, by an average of $3.4$ points in accuracy and secures top performance on six of the seven benchmarks, further highlighting the scalability of \ours. Nevertheless, a performance gap persists between \ours and leading proprietary systems, particularly in tasks requiring deep clinical reasoning or access to extensive medical knowledge. Closing this gap is a primary objective of our ongoing research, which aims to improve domain adaptation, instruction following, and medical reasoning proficiency  in future iterations of \ours.

% ================================
\subsection{Performance on Report Generation Benchmarks}
% ================================
We evaluate the model on a clinically relevant task: automatic medical report generation. Table~\ref{tab:report_result} shows the results of three widely adopted benchmarks: MIMIC-CXR, CheXpert Plus, and IU-Xray. For smaller models ($<$10B), \ours-7B consistently surpasses all competitors on all three datasets, attaining top or second-best scores on every metric. For larger models ($>$10B), performance gaps among open-source models become more pronounced. \ours-32B delivers the strongest overall results, leading in almost all metrics. In particular, its IU-Xray score nearly doubles baseline performance. Besides, there are two notable qualitative observations: 1) On MIMIC-CXR and CheXpert Plus, larger models sometimes underperform smaller ones. This likely occurs because larger models generate more elaborate, stylistically varied descriptions that current automatic metrics do not adequately capture. This highlights the need for improved evaluation methods for medical report generation. While InternVL models excel on multimodal and textual QA benchmarks (see Tables~\ref{tab:mm_result} and~\ref{tab:text_result}), they perform poorly in report generation. This gap reveals a disconnect between benchmark evaluations and real-world medical applications, emphasizing the need for more realistic medical AI assessments.

\begin{table*}[t]
\centering
\caption{Performance of \ours and other MLLMs on medical report generation, all scores are scaled by 100 for clarity. Models colored in \colorbox[HTML]{F5F5F5}{light gray}, \colorbox[HTML]{EBEBEB}{gray} and \colorbox[HTML]{E8F5E9}{green} denote medical MLLMs, general-purpose MLLMs and \ours, respectively.}
\adjustbox{max width=\textwidth}{
\setlength{\tabcolsep}{0.8mm}{
\begin{tabular}{lccccccccccccccc}
\toprule
\multirow{3}{*}{\textbf{Models}}& \multicolumn{5}{c}{\textbf{MIMIC-CXR}} & \multicolumn{5}{c}{\textbf{CheXpert Plus}} & \multicolumn{5}{c}{\textbf{IU-Xray}} \\
\cmidrule(lr){2-6} \cmidrule(lr){7-11} \cmidrule(lr){12-16}
 & \textbf{ROUGE-L} & \textbf{CIDEr} & \textbf{RaTE}  & \textbf{Semb}  & \textbf{RadCliQ}$_{v1}^{-1}$ & \textbf{ROUGE-L} & \textbf{CIDEr} & \textbf{RaTE}  & \textbf{Semb}  & \textbf{RadCliQ}$_{v1}^{-1}$& \textbf{ROUGE-L} & \textbf{CIDEr} & \textbf{RaTE}  & \textbf{Semb}  & \textbf{RadCliQ}$_{v1}^{-1}$\\
\midrule
\multicolumn{16}{c}{\texttt{Proprietary Models}}\\
\midrule
GPT-4.1 & 9.0/24.4 & 82.8/78.9 & 51.3/50.8 & 23.9/23.1 & 57.1/56.3 & 24.5 & 78.8 & 45.5 & 23.2 & 45.5 & 30.2  & 124.6 & 51.3 & 47.5 & 80.3 \\
Claude Sonnet 4 & 20.0  &56.6  &45.6  &19.7  &53.4  &22.0  &59.5  & 43.5 & 18.9 & 43.3 &25.4 & 88.3 & 55.4 & 41.0 & 72.1  \\
Gemini-2.5-Flash & 25.4 & 80.7 & 50.3 &29.7 & 59.4 & 23.6 & 72.2 & 44.3 & 27.4 & 44.0 & 33.5 & 129.3 & 55.6 & 50.9 & 91.6  \\
\midrule
\multicolumn{16}{c}{\texttt{Open-source Models (<10B)}}\\
\midrule
\rowcolor[HTML]{F5F5F5} Med-R1-2B & 19.3 & 35.4 & 40.6 & 14.8 & 42.4 & 18.6 & 37.1 & 38.5 & 17.8 & 37.6 & 16.1 & 38.3 & 41.4 & 12.5 & 43.6 \\
\rowcolor[HTML]{F5F5F5} MedVLM-R1-2B & 20.3 & 40.1& 41.6& 14.2 & 48.3 & 20.9 &43.5 &38.9 & 15.5 & 40.9 & 22.7 & 61.1 & 46.1 & 22.7 & 54.3\\
\rowcolor[HTML]{F5F5F5} MedGemma-4B-IT & \underline{25.6} & 	\underline{81.0} & 	\textbf{52.4} & 	\underline{29.2} & \underline{62.9} &   \textbf{27.1} & 	\underline{79.0} & 	\textbf{47.2} & \textbf{29.3} & 	\underline{46.6} &  \underline{30.8} & 	{103.6} & 	\underline{57.0} & 	\underline{46.8} & \underline{86.7}\\
\rowcolor[HTML]{F5F5F5} LLaVA-Med-7B & 15.0 & 43.4 & 12.8 & 18.3 & 52.9 & 18.4 & 45.5 & 38.8 & 23.5 & 44.0 & 18.8 & 68.2 & 40.9 & 16.0 & 58.1\\
\rowcolor[HTML]{F5F5F5} HuatuoGPT-V-7B & 23.4 & 69.5 & 48.9 & 20.0 & 48.2& 21.3 & 64.7 & 44.2 &19.3 &39.4 & 29.6 &\underline{104.3} &52.9 & 40.7 & 63.6 \\
\rowcolor[HTML]{F5F5F5} BioMediX2-8B & 20.0  & 52.8 & 44.4 & 17.7 & 53.0 & 18.1  & 47.9 & 40.8 & 21.6 & 43.3 & 19.6  & 58.8 & 40.1 & 11.6 & 53.8\\
\rowcolor[HTML]{EBEBEB} Qwen2.5-VL-7B & 24.1 & 63.7 & 47.0  & 18.4 & 55.1 & 22.2 & 62.0 & 41.0 & 17.2 & 43.1 & 26.5 & 78.1 & 48.4  & 36.3 & 66.1 \\
\rowcolor[HTML]{EBEBEB} VideoLLaMA3-7B & 20.8 & 48.7 & 41.9 & 16.4 & 52.9 & 20.6 & 47.6 & 39.4 & 17.3 & 42.0 & 22.0 & 51.9 & 43.8  & 22.5 & 57.4 \\
\rowcolor[HTML]{EBEBEB} InternVL2.5-8B   & 23.2 & 61.8 & 47.0 &21.0 & 56.2 & 20.6 & 58.5  & 43.1 & 19.7 & 42.7 & 24.8 & 75.4 & 51.1 & 36.7 & 67.0 \\
\rowcolor[HTML]{EBEBEB} InternVL3-8B & 22.9 & 66.2 & 48.2 &21.5 & 55.1& 20.9 & 65.4 & 44.3 & 25.2 & 43.7 & 22.9  & 76.2 & 51.2 & 31.3 & 59.9 \\
\hline
\rowcolor[HTML]{E8F5E9} \ours-7B &\textbf{30.8} & \textbf{109.4} & \underline{52.1}  & \textbf{30.0} & \textbf{69.2} & \underline{26.5} & \textbf{79.0} & \underline{45.4} & \underline{26.8} & \textbf{47.3} & \textbf{41.2}  & \textbf{180.7} & \textbf{57.6} & \textbf{48.4} & \textbf{108.1}  \\
\midrule
\multicolumn{16}{c}{\texttt{Open-source Models (>10B)}}\\
\midrule
\rowcolor[HTML]{F5F5F5} HealthGPT-14B &  21.4 &	64.7 & 48.4 &	16.5 &	{52.7} & 20.6 & \underline{66.2} & \underline{44.4} &	22.7 &	42.6 & 22.9 &	81.9 &	50.8 &	16.6 &	56.9 \\
\rowcolor[HTML]{F5F5F5} HuatuoGPT-V-34B & \underline{23.5} & \underline{68.5} & {48.5} & \underline{23.0} & 47.1 & 22.5 & 62.8 & 42.9 & 22.1 & 39.7 & 28.2 & \underline{108.3} & {54.4} & \underline{42.2} & 59.3 \\
\rowcolor[HTML]{F5F5F5} MedDr-40B  & 15.7 & 62.3 & 45.2 & 12.2 & 47.0 & \underline{24.1} & 66.1 & \textbf{44.7} & \underline{24.2} & {44.7} & 19.4 & 62.9 & 40.3 &{7.3} & 48.9\\
\rowcolor[HTML]{EBEBEB} InternVL3-14B & 22.0 & 63.7 & \underline{48.6} & 17.4 & 46.5 & 20.4 & 60.2 &44.1 &20.7 & 39.4& 24.8 &93.7 & \underline{55.0} &38.7 &55.0 \\
\rowcolor[HTML]{EBEBEB} Qwen2.5-VL-32B  & 15.7 & 50.2 & 47.5 & 17.1 & 45.2 & 15.2 & 54.8 & 43.4 & 18.5 & 40.3 & 18.9 & 73.3 & 51.3 & 38.1 & 54.0 \\
\rowcolor[HTML]{EBEBEB} InternVL2.5-38B & 22.7 & 61.4 & 47.5 &18.2 & \underline{54.9}& 21.6 & 60.6 & 42.6 & 20.3 & \underline{45.4} & \underline{28.9} & 96.5  & 53.5 & 38.5 & \underline{69.7} \\
\rowcolor[HTML]{EBEBEB} InternVL3-38B & 22.8 & 64.6 & 47.9 &18.1 & 47.2& 20.5 & 62.7 & 43.8&20.2 & 39.4& 25.5 & 90.7 & 53.5 &33.1 & 55.2 \\

\hline
\rowcolor[HTML]{E8F5E9} \ours-32B & \textbf{28.8} & \textbf{96.4} & \textbf{50.8} & \textbf{30.1} & \textbf{67.1} & \textbf{25.3} & \textbf{75.9} & 43.4 & \textbf{24.2} & \textbf{47.1} & \textbf{42.8} & \textbf{189.2} & \textbf{63.5} & \textbf{54.6} & \textbf{130.4} \\
\bottomrule
\end{tabular}}}
\label{tab:report_result}
\end{table*}

% ================================
\subsection{Performance of Medical-oriented RL}
\label{ssec:rl_performancce}
% ================================
We conduct an initial exploration of reinforcement learning with verifiable rewards (RLVR) to enhance medical multimodal reasoning capabilities of \ours.

% ================================
\subsubsection{Medical-oriented Reinforcement Learning}
\label{ssec:rl}
% ================================
Recent reasoning models such as OpenAI's o-series and DeepSeek-R1~\citep{deepseekai2025deepseekr1} achieve strong performance through RLVR, particularly through Group Relative Policy Optimization (GRPO; \citep{shao2024deepseekmath}). RLVR encourages autonomous reasoning discovery through reward signals rather than answer memorization, addressing the limitations of SFT, such as over-reliance on supervision and limited reasoning development~\citep{chu2025sft}. Motivated by this success, we extend RLVR to medical multimodal tasks. The medical dataset combines multiple sources with rigorous post-processing. First, we reformulate MCQA with word or phrase answers into open-ended QA to enhance difficulty while maintaining verifiability, leading to a data balance between MCQA and open-ended QA. Binary-answer questions (\eg, ``Yes/No'') are downsampled to comprise  5\% of the dataset to reduce bias from over-representation. Samples are then selected based on their modality and query format, aiming to ensure the data balance. Finally, a total of 100K examples are curated to form the training set. 

Details about the dataset usage and mixture for each stage are summarized in the Appendix~\S\text{B}. The detailed implementations for both \ours (featured model) and \ours-RL (exploratory variant) are described in Appendix \S\text{C}.

\begin{table*}[t]
\centering
\caption{Performance of \ours and its RL variant (\ours-RL) on medical multimodal benchmarks.}
\adjustbox{max width=\textwidth}{
\begin{tabular}{lcccccccc}
\toprule
\textbf{Models} & \textbf{MMMU-Med} & \textbf{VQA-RAD} & \textbf{SLAKE} & \textbf{PathVQA} & \textbf{PMC-VQA} & \textbf{OmniMedVQA} & \textbf{MedXpertQA} & \textbf{Avg.}\\
\midrule
\ours-7B & 54.0 & 67.9 & \textbf{83.1} & \textbf{61.9} & 56.3 & \textbf{82.9} & \textbf{26.7} & \textbf{61.8}\\
\ours-RL-7B & \textbf{54.7} & \textbf{68.3} & 82.5 & 61.1 & \textbf{57.0} & 81.3 & \textbf{26.7} & 61.5\\
\bottomrule
\end{tabular}
}
\label{tab:rl_result}
\end{table*}

% ========================
\subsubsection{Result Analysis}
% ========================
In Table~\ref{tab:rl_result}, while RL-trained variant (\ours-RL-7B) achieves marginal improvements on MMMU-Med ($+0.7\%$), PMC-VQA ($+0.7\%$), and VQA-RAD ($+0.4\%$), it exhibits a performance degradation on SLAKE ($-0.6\%$), PathVQA ($-0.8\%$), and OmniMedVQA ($-1.6\%$). The average performance remains stable, indicating that the current RLVR training has not achieved the same improvements as those observed in textual math and code tasks. We analyze the potential causes from three aspects, and emphasize that
this analysis is a hypothesis rather than a fully established mechanism:
\begin{enumerate}
    \item \textbf{Vision-Reasoning Bottleneck:} We hypothesize that a plausible reason for the limited gains of RLVR in the medical domain is a disconnect between textual reasoning optimization and visual perception. Unlike mathematical or coding task, where reasoning is purely symbolic and resides within the textual domain, medical VQA is a heavily visually grounded challenge. Accurate visual perception is a key prerequisite for subsequent logical deduction. Our RL implementation mainly enhances textual reasoning trajectories, but does not fundamentally improve the fine-grained visual feature extraction of the vision encoder, which may create a mismatch. For tasks requiring precise anatomical localization (SLAKE) or microscopic texture recognition (PathVQA), textual reasoning cannot compensate for visual deficits. RL tends to prioritize logically coherent, well-formatted text while overlooking ambiguous visual evidence, leading to the observed drops (-0.6, -0.8). The same effect may explain the severe degradation (-1.6) on OmniMedVQA, where distractor options are often clinically plausible, so a misread visual cue leads textual reasoning to a wrong option. In contrast, RL yields marginal gains (+0.4, +0.7) on VQA-RAD and MMMU-Med, where questions often rely on general knowledge recall once the modality or broad anatomy is identified, masking the visual bottleneck.
    \item \textbf{Reward Design:} Our implementation adopts the prevailing accuracy-oriented, rule-based rewards, which may provide unreliable signals for open-ended medical questions with substantial linguistic variability. However, we believe this is not the main cause. On the one hand, the vast majority of our training data has definitive answers (\eg, multiple-choice options or medical terms). On the other hand, the model still suffers a severe degradation on the multiple-choice benchmark OmniMedVQA.
    \item \textbf{Data Quality:} A significant proportion of medical multimodal tasks, such as lesion classification, demand visual understanding rather than complex reasoning; this distribution further exacerbates the vision-reasoning bottleneck. Moreover, the benchmarks used here are not tailored for text-centric medical reasoning, which prevents our model's text-based reasoning abilities from being fully demonstrated.
\end{enumerate}
Overall, this analysis is exploratory: our observations are consistent with a vision-reasoning bottleneck but do not conclusively isolate it from factors such as reward design or data distribution. We thus present it as a hypothesis and leave rigorous verification to future work.

% ========================
\subsection{Human Expert Study}
% ========================

\begin{figure}[htbp]
    \centering
        \includegraphics[width=\linewidth]{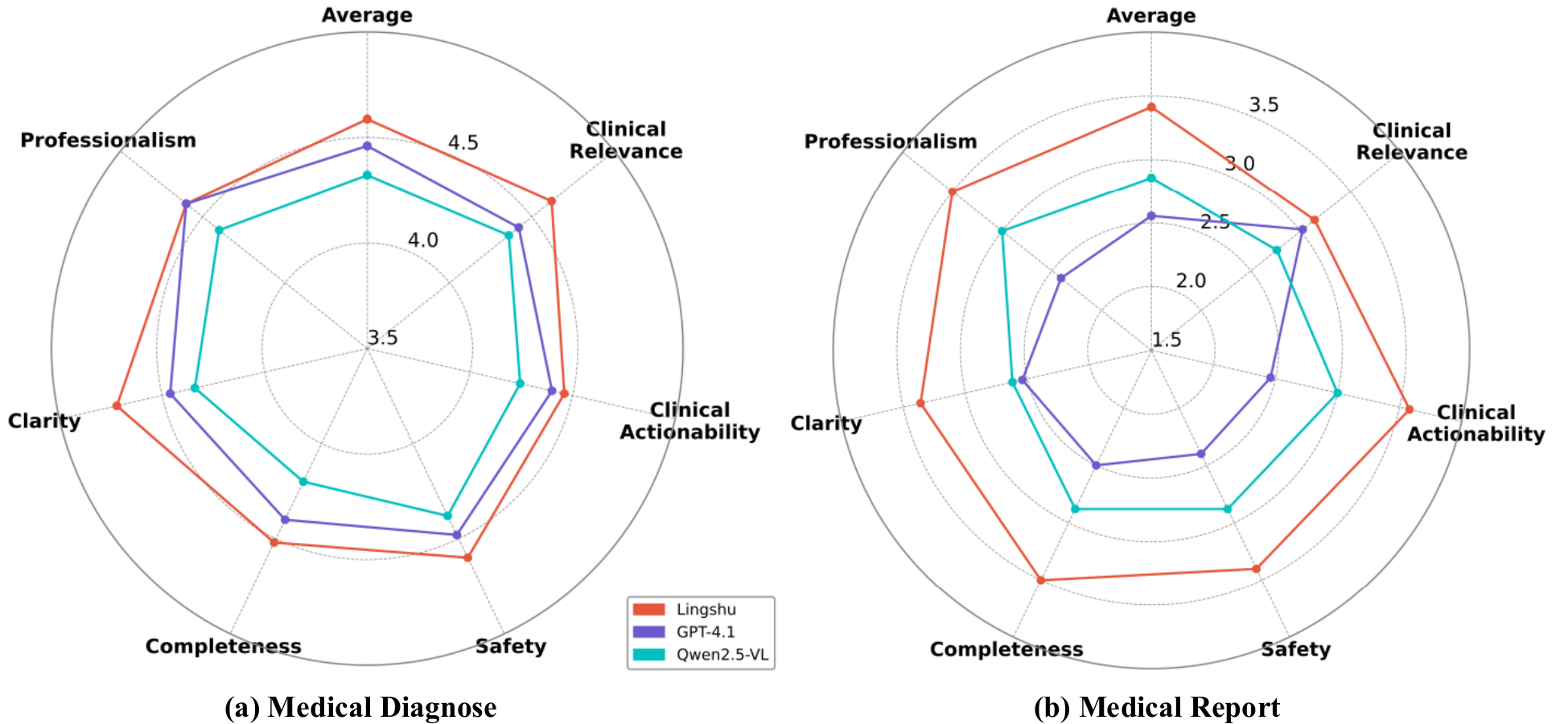}
        \caption{Physician ratings of \ours, GPT-4.1, and Qwen2.5-VL across six clinical dimensions on Medical Diagnosis and Medical Report generation tasks.}
    \label{fig:human_study}
\end{figure}

To assess the real-world clinical utility of MLLMs, we conducted a blinded evaluation in which two board-certified radiologists rated outputs from \ours, GPT-4.1, and Qwen2.5-VL across both medical diagnosis and report generation tasks. The radiologists were asked to evaluate each response on six dimensions: Clinical Relevance, Actionability, Safety, Completeness, Clarity, and Professionalism (1--5 scale). The full protocol, including the rater qualifications, case-to-rater assignment, scoring aggregation, and inter-rater agreement, is detailed in Appendix~\S F. As shown in Fig.~\ref{fig:human_study}, \ours consistently outperforms both GPT-4.1 and Qwen2.5-VL across all dimensions on both tasks, with the performance gap being particularly pronounced in report generation, where the demands of synthesizing complex clinical information into structured narratives better distinguish the models. Notably, \ours's advantages in Safety and Clinical Actionability suggest that its outputs are not only less prone to hallucinated or misleading content, but also more immediately useful to clinicians in practice, demonstrating that \ours delivers superior clinical value as judged by domain experts under realistic conditions.

% ================================
\subsection{Ablation Study on Data Composition}
% ================================

To gain insights into the impact of different data types during the medical instruction tuning stage, we conducted an ablation study by selectively removing specific categories of training data, as shown in Table~\ref{tab:ablation}. In general, all four categories of training data outlined in \S\ref{ssec:sft}, namely, \texttt{\{Medical, General\} $\times$ \{Multimodal, Text\}}, collectively contribute to \ours's task-solving ability, even when some data sources are not explicitly tailored for the medical domain. Notably, medical textual data emerge as the most critical component: the removal of merely 173K samples leads to substantial performance degradation on five of the seven tasks.

Different types of data impact performance uniquely. Removing medical multimodal data (\textit{- Medical Multimodal}) reduces SLAKE, PathVQA, and OmniMedVQA scores since these benchmarks focus on histopathology and X-ray images, which are prevalent in medical multimodal data. Removing general multimodal data (\textit{- General Multimodal}) severely impacts MMMU-Med and PMC-VQA (see Appendix \S G for fine-grained scale analysis). These tasks require a nuanced understanding of medical content in complex formats, such as medical charts, which are lacking in medical multimodal datasets. We conjecture that the multimodal and textual data from the general domain can effectively supplement this specialized knowledge by compensating for domain-specific limitations. This supports our hypothesis: \textit{ the acquisition of comprehensive medical knowledge for multimodal problems should not be based exclusively on multimodal medical data}. Rather, \textbf{ it requires the integration of medical text and general data for a more robust and generalizable model.}

Beyond the overall data composition, we examine the contributions of several high-value data components:
\begin{itemize}
    \item \textbf{Medical Caption (Stage 1\&2)}: Skipping Stage 1\&2 alignment causes moderate performance drops on PathVQA, PMC-VQA, and OmniMedVQA, suggesting that early-stage alignment remains a critical step in the training pipeline.
    \item \textbf{Medical Caption (Stage 3)}: As mentioned in \S\ref{ssec:sft}, we retain some high-quality medical captions during Stage 3 to mitigate local view bias. Removing this data severely degrades performance on MMMU-Med, VQA-RAD, PMC-VQA, and OmniMedVQA, underscoring the importance of caption data throughout training.
    \item \textbf{Synthetic Multimodal Medical data}: Removing all synthetic multimodal data described  in \S\ref{ssec:med_data_syn} causes notable declines in MMMU-Med, SLAKE, PathVQA, and especially OmniMedVQA, which covers a broader range of medical modalities. This shows synthetic data is crucial for filling data gaps and enhancing medical knowledge across modalities.
    \item \textbf{Distilled Textual Medical Reasoning Data}: Given the importance of textual data, we further isolate the impact of medical reasoning texts. Removing this subset causes performance drops on five of the seven tasks, nearly equivalent to removing all medical text. We attribute this effect to the dense medical knowledge encoded within CoT reasoning traces. This finding suggests a promising direction for future research: developing diverse, high-quality corpora of medical reasoning texts to systematically enhance the acquisition of medical expertise.
\end{itemize}

\begin{table*}[t]
\centering
\caption{Effect of data composition at the medical instruction tuning stage. $\Delta\vert$\textbf{Data}$\vert$ denotes the reduced amount of training data, where $\color{red}\downarrow$ indicates significantly lower performance compared to \ours, while  $\color{red}\downarrow\downarrow$ shows an even more substantial decrease.}
\adjustbox{max width=\textwidth}{
\begin{tabular}{lclllllll}
\toprule
\textbf{Models} & $\Delta\vert$\textbf{Data}$\vert$ &\textbf{MMMU-Med} & \textbf{VQA-RAD} & \textbf{SLAKE} & \textbf{PathVQA} & \textbf{PMC-VQA} & \textbf{OmniMedVQA} & \textbf{MedXpertQA} \\
\midrule
Qwen2.5-VL-7B & - & 50.6 & 64.5 & 67.2 & 44.1 & 51.9 & 63.6 & 22.3 \\
\ours-7B & -& {54.0} & {67.9} & {83.1} & {61.9} & {56.3} & {82.9} & {26.7}\\
\midrule
\multicolumn{9}{c}{\textit{Overall Data Composition}}\\
\midrule
- Medical Multimodal &2.7M   & 54.1 & 66.5$\color{red}\downarrow$ & 69.0$\color{red}\downarrow\downarrow$ & 45.2$\color{red}\downarrow\downarrow$ & 56.2 & 78.0$\color{red}\downarrow\downarrow$ & 25.6$\color{red}\downarrow$  \\
- General Multimodal   &1.2M  & 51.4$\color{red}\downarrow\downarrow$ & 66.7$\color{red}\downarrow$ & 83.6 & 62.1 & 54.5$\color{red}\downarrow$ & 81.3$\color{red}\downarrow$ & 26.0  \\
- General Text     &1M     & 53.3$\color{red}\downarrow$ & 65.2$\color{red}\downarrow$ & 83.6 & 61.4 & 55.5$\color{red}\downarrow$ & 81.3$\color{red}\downarrow$ & 25.9$\color{red}\downarrow$  \\
- Medical Text     &173K     &  50.9$\color{red}\downarrow\downarrow$ & 66.7$\color{red}\downarrow$ & 82.4 & 60.2$\color{red}\downarrow$ & 54.8$\color{red}\downarrow$ & 82.0 & 24.0$\color{red}\downarrow\downarrow$  \\
\midrule
\multicolumn{9}{c}{\textit{Effect of Selected High-value Data}}\\
\midrule
- Medical Caption (Stage 1\&2) & 4.8M & 53.8 & 67.4 & 83.3 & 60.6$\color{red}\downarrow$ & 54.9$\color{red}\downarrow$ & 81.5$\color{red}\downarrow$ & 26.7 \\
- Medical Caption (Stage 3) & 2M & 52.9$\color{red}\downarrow$ & 66.3$\color{red}\downarrow$ & 82.7 & 61.7 & 54.1$\color{red}\downarrow$ & 81.4$\color{red}\downarrow$ & 26.8  \\
- Synthetic Multimodal  & 1.1M & 52.1$\color{red}\downarrow$ & 67.4 & 81.6$\color{red}\downarrow$ & 60.1$\color{red}\downarrow$ & 56.5 & 71.6$\color{red}\downarrow\downarrow$ & 26.1 \\
- Medical CoT-Text   & 162K & 52.6$\color{red}\downarrow$ & 66.3$\color{red}\downarrow$ & 83.6 & 61.1 & 53.6$\color{red}\downarrow\downarrow$ & 81.6$\color{red}\downarrow$ &25.3$\color{red}\downarrow$  \\
\bottomrule
\end{tabular}}
\label{tab:ablation}
\vspace{-0.1in}
\end{table*}

% ================================
\subsection{Analysis of Stage-wise Training}
\label{sec:training_dynamics}
% ================================

\begin{figure}[htbp]
    \centering
        \includegraphics[width=\linewidth]{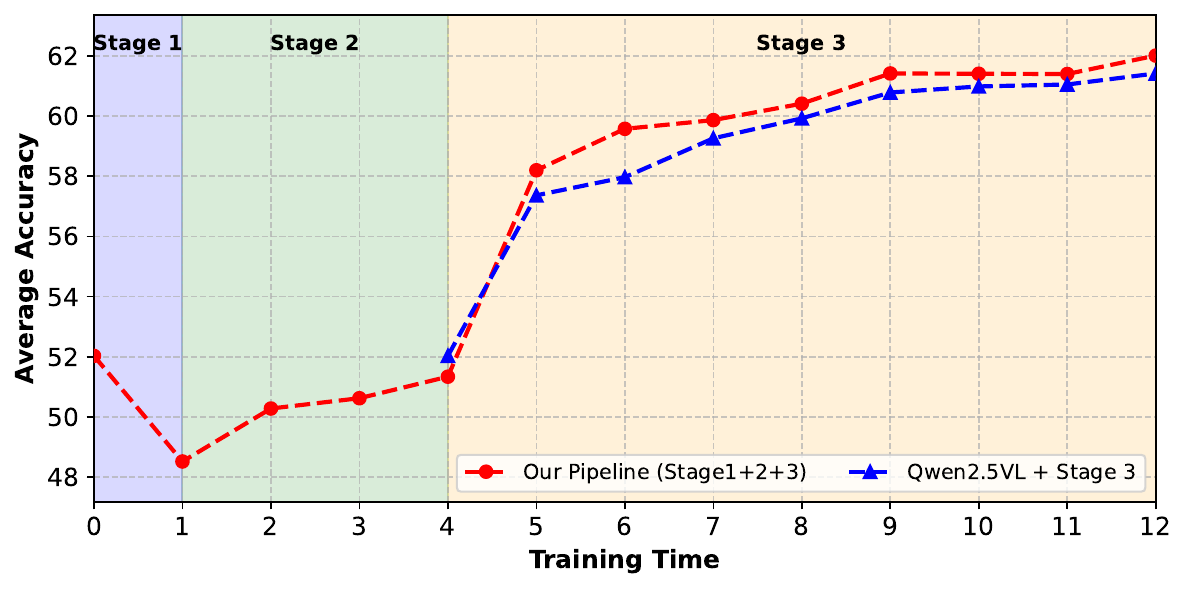}
        \caption{Downstream task performance evaluated at intermediate checkpoints throughout the three-stage training process.}
    \label{fig:per_step}
\end{figure}

To validate our multi-stage paradigm, we analyze the model's evolution across training. Figure~\ref{fig:per_step} illustrates the average accuracy on medical benchmarks during training. In Stage 1, we observe a marginal performance decrease. This can be attributed to the objective shift from general-purpose instruction following to medical-specific image-text alignment. However, Stage 2 shows a steady recovery. Since Stage 2 only uses high-quality medical captions, this gain confirms that the model successfully internalizes complex medical concepts from dense descriptive data. Stage 3 then achieves a performance breakthrough with clear scaling law behavior. To further isolate the benefits of our alignment stages, we introduce a baseline where the Qwen2.5-VL model is directly fine-tuned on the Stage 3 dataset. As shown in Figure~\ref{fig:per_step}, although both models improve rapidly during Stage 3, our full pipeline consistently outperforms the direct fine-tuning approach throughout the entire stage, achieving a higher performance ceiling. This gap demonstrates that the alignment stages do not just "warm up" the model, but effectively build a domain-specific knowledge base that direct fine-tuning cannot easily replicate.

% ================================
\subsection{Quality Analysis of \ours Training Data}
% ================================
\begin{table}[htbp]
\centering
\caption{Data quality across four dimensions for Caption Data and Instruction Data.}
\label{tab:training_data_quality}
\resizebox{\columnwidth}{!}{%
\setlength{\tabcolsep}{0.8mm}{
\begin{tabular}{lcccc}
\toprule
\textbf{Dataset} & \textbf{Sent. Length} & \textbf{Logic. Complexity} & \textbf{Vocab. Density} & \textbf{Term. Diversity} \\
\midrule
\multicolumn{5}{c}{\textit{Caption Data}}\\
\midrule
Synthetic Multimodal   & 2.21 & 3.12 & 4.04 & 3.66 \\
Caption (Stage 1)      & 0.97 & 1.17 & 3.67 & 2.60 \\
Caption (Stage 2)      & 1.80 & 1.80 & 2.49 & 1.93 \\
\midrule
\multicolumn{5}{c}{\textit{Instruction Data}}\\
\midrule
Synthetic CoT          & 2.21 & 3.25 & 3.46 & 2.90 \\
Medical Reasoning Text & 2.99 & 2.96 & 3.58 & 3.52 \\
Medical Text           & 1.56 & 1.53 & 2.34 & 2.06 \\
Medical Multimodal     & 1.04 & 2.33 & 2.68 & 2.17 \\
General Multimodal     & 1.93 & 2.61 & 2.45 & 1.92 \\
\bottomrule
\end{tabular}%
}}
\end{table}

We further conduct a quality analysis to investigate why our data improves medical capabilities from four dimensions: \textit{medical terminology diversity}, \textit{medical vocabulary density}, \textit{logical complexity}, and \textit{response length}, as shown in Table~\ref{tab:training_data_quality}. The detailed annotation process and definitions of these four dimensions are provided in Appendix \S H. For caption data, our synthetic long-caption data is considerably longer than existing caption data, consequently encompassing more extensive and diverse medical knowledge as well as more sophisticated reasoning logic. The caption data in stage 2 exhibits higher logical complexity,  which may benefit the medical reasoning. This observation may explain the performance gains achieved through stage 2 training in Fig.~\ref{fig:per_step}. Regarding instruction data, both our synthetic multimodal data and medical CoT text data demonstrate high quality across all four dimensions, which aligns with their great effectiveness in Table~\ref{tab:ablation}. Interestingly, general multimodal data, despite not being designed for medical applications, contains a comparable amount of medical knowledge to existing medical multimodal datasets. Moreover, general multimodal data exhibits strong logical complexity, enabling the model to learn reasoning capabilities that can transfer effectively to the medical domain.

% ================================
\subsection{Results per Modality on OmniMedVQA}
\begin{figure}[t]
    \centering
    \includegraphics[width=\linewidth]{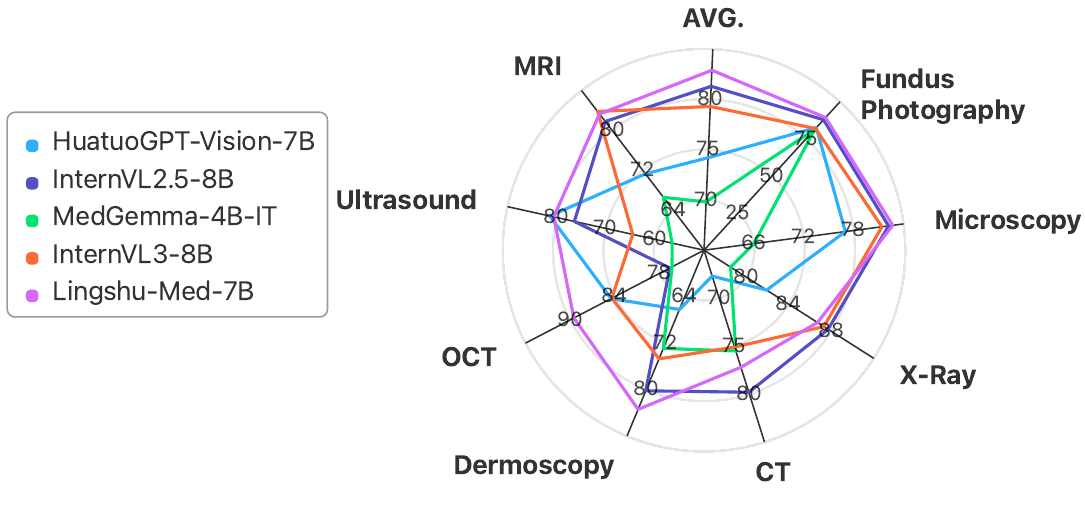}
    \caption{Performance of different MLLMs across medical modalities on the OmniMedVQA.}
    \label{fig:modality}
    \vspace{-0.1in}
\end{figure}

% ================================
The analysis in \S\ref{ssec:med_multimodal_results} shows the effectiveness of \ours in diverse medical tasks. Here, we investigate how these MLLMs perform across different medical imaging modalities. Fig.~\ref{fig:modality} benchmarks \ours-7B against the four strongest baselines, based on their average performance in medical multimodal tasks, \ie, InternVL-3-8B, InternVL-2.5-8B, HuatuoGPT-V-7B, and MedGemma-4B-IT.

Across eight medical imaging modalities, \ours either matches or exceeds its competitors on most fronts. It achieves leading performance in microscopy, MRI, dermoscopy, and OCT modalities that demand precise recognition of micro-textural and macro-structural cues. This suggests that \ours is particularly effective in capturing both high-frequency detail and anatomical regularities. \ours also delivers promising performance on ultrasound and fundus photography, underscoring its robustness in domains characterized by speckle noise (ultrasound) and fine vascular patterns (fundus). In CT and X-ray, \ours trails the best baseline by a modest margin. Overall, \ours establishes itself as a modality-agnostic yet high-resolution specialist, outperforming larger competitors despite its relatively compact parameter budget.

\begin{figure*}[t]
    \centering
    \includegraphics[width=\textwidth]{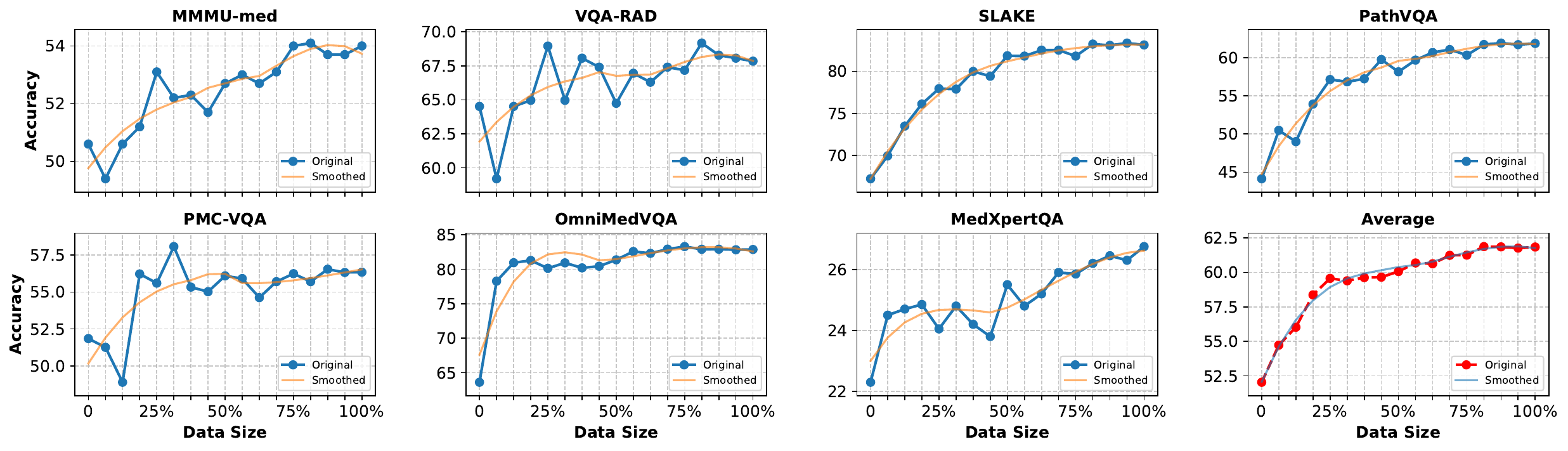}
    \caption{The impact on downstream task performance when increasing the scale of training data.}
    \label{fig:scaling}
    \vspace{-0.1in}
\end{figure*}

% ================================
\subsection{Analysis of Data Volume Scaling}
% ================================
Fig.~\ref{fig:scaling} shows how accuracy changes as the training size increases from 0\% to 100\% of the full dataset. A common pattern emerges across all seven benchmarks: accuracy rises sharply with the first quarter of data, then continues to climb with diminishing gains, until it converges near the $75\text{–}100\%$ mark. The smoothed curves reveal a near-monotonic trajectory, confirming that the overall trend is robust. Two further observations merit emphasis. First, tasks that are intrinsically image-centric (\eg, SLAKE and PathVQA) saturate earlier than concept-heavy challenges such as MedXpertQA, which still shows steady improvements at the upper end of the data scale. Second, the aggregate plot steadily climbs from $52\%$ to $62\%$ accuracy, thereby quantifying a clear data scaling effect for training medical MLLMs. Taken together, these findings highlight the decisive role of large-scale data acquisition while also signaling an eventual plateau, after which further progress will likely depend on high-quality data enrichment or architectural innovation rather than on volume alone.

\begin{figure*}[t]
    \centering
    \includegraphics[width=\textwidth]{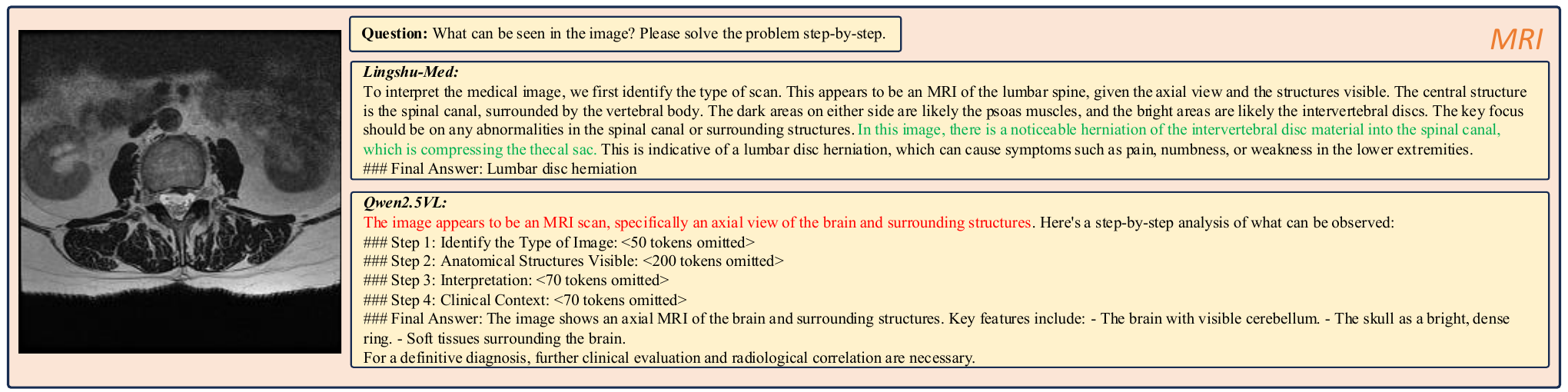}
    \caption{Case studies on \ours and Qwen2.5-VL in visual question answering. The \textcolor{green}{ green parts} are some important reasoning during the generation. The \textcolor{red}{red part} indicates the thoughtless reasoning.}
    \label{fig:modality_vqa_short}
    \vspace{-0.1in}
\end{figure*}

% ================================
\subsection{Case Studies}
\label{sec:case_studies}
% ================================
Fig.~\ref{fig:modality_vqa_short} reveals a dangerous hallucination pattern in Qwen2.5-VL. The model not only misidentifies a lumbar spine MRI as a brain scan but proceeds to hallucinate detailed cerebral anatomy that does not exist in the image, demonstrating how confident-sounding but entirely fabricated descriptions can emerge. In contrast, \ours provides accurate anatomical identification and correct diagnosis of lumbar disc herniation. VQA examples across more modalities and cases of \textit{Report Generation}, \textit{Medical Diagnosis}, \textit{Medical Knowledge}, \textit{Public Health}, and \textit{Medical Consultation} are detailed in Appendix~\S\text{I}.

%% file: sections/7_related_work.tex
% ================================
\section{Related work}
\label{sec:related_work}
% ================================
The rapid development of MLLMs has achieved tremendous improvements in the medical domain, spurring medical applications and healthcare progress~\citep{tian2023role,alSaad2024multimodal}. Early endeavors combine LLMs with specialized medical visual encoders via a projection layer to establish vision-language alignment for medical image understanding and analysis~\citep{li2023llavamed,moor2023medflamingo,zhang2024biomedgpt}. Based on this foundation, subsequent studies adopt similar architectural paradigms for developing medical MLLMs, incorporating various strategies, such as curating more comprehensive datasets~\citep{ikezogwo2023quilt1m,chen2024huatuogptvision,li2025gmaivl,sun2025reasonmed}, designing sophisticated training recipes~\citep{nath2025vilam3,wang2024chatcad}, employing efficient fine-tuning techniques~\citep{lin2025healthgpt}, integrating mixture-of-experts~\citep{he2024gsco}, and leveraging reinforcement learning~\citep{lai2025medr1,pan2025medvlmr1}, to improve performance across various medical tasks. Note that we draw high-level insights from ReasonMed~\citep{sun2025reasonmed} on the effectiveness of long-CoT and the importance of quality control; however, \ours mainly focuses on medical multimodal understanding and reasoning, and thus requires imaging-specific synthesis and verification. Meanwhile, we do not reuse or incorporate any ReasonMed-synthesized data; our dataset and pipeline are independently built. In parallel, widely used proprietary models have integrated medical knowledge, \ie Med-Gemini\footnote{\url{https://research.google/blog/advancing-medical-ai-with-med-gemini/}} and Med-PaLM~\citep{singhal2023large}, and achieve strong performance on diverse medical tasks~\citep{yang2024advancing,aydin2025openai,arora2025healthbench}. 

Beyond general medical MLLMs, many studies explore specialized MLLMs tailored to specific diseases or clinical contexts to better support downstream applications, such as those designed for pathology~\citep{lu2024pathchat,zhao2024biomedparse,seyfioglu2025quiltllava}, radiology~\citep{hyland2024maira1,tanno2024flamingocxr}, ophthalmology~\citep{deng2024ophglm}, pan-cancer analysis~\citep{keyl2024decoding}, specific cancer types~\citep{sammut2022multi,pai2024cancerfm,niu2025multimodal}, etc. Furthermore, building upon the capabilities of medical MLLMs, researchers have initiated the development of medical agents that extend model functionality through integration with external medical tools, such as AgentClinic~\citep{schmidgall2025agentclinic}, Agent Hospital~\citep{li2025agenthospital}, and MedAgent-Pro~\citep{wang2025medagentpro}. In contrast, we focus on developing a multimodal foundation model that accommodates diverse medical modalities and supports tasks such as question answering, diagnosis, report generation, and diverse clinical applications.

%% file: sections/8_conclusion.tex
% ================================
\section{Conclusion}
\label{sec:conclusion}
% ================================

In this work, we introduce \ours, a domain-specialized multimodal foundation model designed for medical AI, addressing key challenges in \emph{medical data construction}, \emph{model training}, and \emph{evaluation}. By unifying large-scale, high-quality medical and general-domain corpora with a comprehensive data curation pipeline, encompassing image captions, QA pairs, and CoT annotations, \ours acquires an expansive and diverse knowledge base. Our multi-stage training paradigm progressively enhances the model's medical understanding and problem-solving skills. Moreover, we explore the potential of incorporating advanced reinforcement learning with verifiable rewards to further boost performance. Beyond model development, we emphasize the importance of evaluation standardization in the medical domain and introduce \oureval, a unified framework that consolidates multiple multimodal and textual benchmarks while enforcing standardized evaluation. This toolkit enables fair, reproducible, and transparent comparisons across models. Extensive experiments confirm the superior performance of \ours. Across a broad spectrum of medical VQA and report generation tasks, \ours consistently surpasses open-source baselines and narrows the gap with proprietary models. Overall, \ours and \oureval provide a high-performing model, a robust evaluation toolkit, and empirically grounded guidelines on data curation, staged training, and evaluation. These contributions represent a step forward in aligning MLLMs with real-world medical applications.

%% file: tables/appdx_medical_multimodal_data_collection.tex
\begin{savenotes}
\begin{table*}[t]
\caption{List of collected medical multimodal datasets. For MIMIC-CXR~\citep{johnson2019mimiccxr} in medical captioning data, we only use its ``Findings'' part as the image captions.
}
    \centering
    \adjustbox{max width=1.0\textwidth}{
    \begin{tabular}{lp{14cm}}
    \toprule
        \textbf{Type of Data} & \textbf{Collected Datasets} \\
        \midrule
        Medical Captioning Data & LLaVA-Med Alignment~\citep{li2023llavamed}, PubMedVision~\citep{chen2024huatuogptvision}, Quilt-LLaVA-Pretrain~\citep{seyfioglu2025quiltllava}, PMC-OA~\citep{liu2023pmcclip}, ROCO~\citep{pelka2018roco}, ROCOv2~\citep{ruckert2024rocov2}, MIMIC-CXR~\citep{johnson2019mimiccxr}, MedICaT~\citep{subramanian2020medicat}, FairVLMed~\citep{luo2024fairclip}, and MedPix-2.0~\citep{siragusa2025medpix20}  \\
        \midrule
        Medical Instruction Data & PathVQA~\citep{he2020pathvqa}, PMC-VQA~\citep{zhang2024pmcvqa}, SLAKE~\citep{liu2021slake}, Quilt-LLaVA Instruct~\citep{seyfioglu2025quiltllava}, VQA-Med-2019~\citep{ben2019vqamed}, PubMedVision~\citep{chen2024huatuogptvision}, LLaVA-Med~\citep{li2023llavamed}, LLaVA-Med-zh\footnote{\url{https://huggingface.co/datasets/BUAADreamer/llava-med-zh-instruct-60k}}, VQA-RAD~\citep{lau2018vqarad}, MIMIC-Ext-MIMIC-CXR-VQA\footnote{\url{https://physionet.org/content/mimic-ext-mimic-cxr-vqa/1.0.0/}}, CheXpert Plus~\citep{chambon2024chexpertplusaugmentinglarge}, MIMIC-CXR~\citep{johnson2019mimiccxr}, and IU-Xray~\citep{demner2016preparing-iu-xray} \\
        \bottomrule
    \end{tabular}}
    \label{tab:data_set_source_multi_modal}
\end{table*}
\end{savenotes}

%% file: tables/appdx_medical_textual_data_collection.tex
\begin{savenotes}
\begin{table*}[t]
\caption{List of collected medical textual instruction datasets.}
    \centering
    \adjustbox{max width=1.0\textwidth}{
    \begin{tabular}{lp{13cm}}
    \toprule
        \textbf{Type of Data} & \textbf{Collected Datasets} \\
        \midrule
        Medical Factoid QA & MedQA~\citep{jin2020medqausmle}, MedQuAD~\citep{ben2019medquad}, ApolloCorpus~\citep{wang2024apollo}, and PMC-LLaMA\footnote{\url{https://huggingface.co/datasets/axiong/pmc\_llama\_instructions}} \\
        \midrule
        Distilled Textual Medical Reasoning Data & MedThoughts\footnote{\url{https://huggingface.co/datasets/hw-hwei/MedThoughts-8K}}, MedReason~\citep{wu2025medreason}, medical-o1-reasoning~\citep{chen2024huatuogpto1}, medical-o1-verifiable~\citep{chen2024huatuogpto1}, and Medical-R1-Distill~\citep{chen2024huatuogpto1}  \\
        \midrule
        Patient-doctor Dialogues & HealthCareMagic\footnote{\url{https://huggingface.co/datasets/lavita/ChatDoctor-HealthCareMagic-100k}}, icliniq\footnote{\url{https://huggingface.co/datasets/lavita/ChatDoctor-iCliniq}}, and HuatuoGPT-II-GPT4~\citep{chen2023huatuogptii} \\
        \midrule
        General Medical Instruction Data &  AlpaCare-MedInstruct\footnote{\url{https://huggingface.co/datasets/lavita/AlpaCare-MedInstruct-52k}} \\
        \bottomrule
    \end{tabular}}
    \label{tab:data_set_source_text}
\end{table*}
\end{savenotes}

%% file: tables/appdx_medical_imaging_data_collection.tex
\begin{savenotes}
\begin{table*}[t]
\caption{List of collected medical imaging datasets of different medical modalities.}
    \centering
    \adjustbox{max width=1.0\textwidth}{
    \begin{tabular}{lp{16cm}}
    \toprule
        \textbf{Modality} & \textbf{Collected Datasets} \\
        \midrule
        X-ray &	COVID19-Radiography~\citep{chowdhury2020covid19}, NIH-Chest-X-Ray~\citep{wang2017chestx}, CheXpert~\citep{irvin2019chexpert}, and Mendeley Digital Knee X-ray\footnote{\url{https://doi.org/10.17632/t9ndx37v5h.1}} \\
        \midrule
        CT &	KIPA22~\citep{HE2021102055}, DeepLesion~\citep{yan2017deeplesion}, and chest-ctscan\footnote{\url{https://tianchi.aliyun.com/dataset/93929}} \\
        \midrule
        MRI & Brain-Tumor-MRI\footnote{\url{https://www.kaggle.com/dsv/2645886}}, BraTS2024\footnote{\url{https://www.synapse.org/Synapse:syn53708249/wiki/}}, LLD-MMRI~\citep{lou2025sdr}, and MAMA-MIA~\citep{garrucho2025} \\
        \midrule
        Ultrasound &	AbdomenUS~\citep{vitale2020improving}, Breast Ultrasound Images\footnote{\url{https://www.kaggle.com/datasets/aryashah2k/breast-ultrasound-images-dataset}}, Annotated Ultrasound Liver images\footnote{\url{https://doi.org/10.5281/zenodo.7272660}}, and RadIMageNet (subset of ultrasound images)~\citep{RadImageNet} \\
        \midrule
        Dermoscopy &	HAM10000~\citep{philipp2018ham10000}, ISIC challenge 2020\footnote{\url{https://kaggle.com/competitions/isic-2024-challenge}}, and Fitzpatrick17k~\citep{groh2021evaluating} \\
        \midrule
        Fundus &	BRSET~\citep{nakayama2023brazilian}, PAPILA~\citep{kovalyk2022papila}, and 2015\&2019 Diabetic Retinopathy Detection\footnote{\url{https://www.kaggle.com/datasets/c7934597/resized-2015-2019-diabetic-retinopathy-detection}} \\
        \midrule
        Histopathology & CPD~\citep{wagner2023semantic}, Breast-Cancer~\citep{ding2023large}, PanNuke~\citep{gamper2020pannuke}, and EBHI-Seg~\citep{shi2023ebhi} \\
        \midrule
        Microscopy &	Blood Cell Image\footnote{\url{https://github.com/Shenggan/BCCD\_Dataset}}, BioMediTech RPE\footnote{\url{https://figshare.com/s/d6fb591f1beb4f8efa6f}}, and mhsma-dataset~\citep{javadi2019novel} \\
        \bottomrule
    \end{tabular}}
    \label{tab:data_set_source_med_image}
\end{table*}
\end{savenotes}

%% file: tables/appdx_patient_doctor_dialogue_clean.tex
\begin{table*}[t]
\caption{The prompt used for model-based response cleaning for patient-doctor dialogue datasets.}
\label{tab:prompt_doctor_patient_dialog_cleaning}
\begin{tcolorbox}[colframe=orange!60, colback=orange!20, fonttitle=\bfseries\large, coltitle=black, boxrule=0.75mm, arc=5mm, auto outer arc, width=\textwidth,toptitle=6pt, bottomtitle=6pt]
\small
\setstretch{1.2} %
Response to patient: \\
\{response\}\\
\\
Revise the above response in a tone of an emphatic health-care agent. You should offer reliable information about medical conditions, but avoid giving explicit diagnosis and prescription. Add suggestions about which kinds of doctors the patient should consult. Include a disclaimer statement saying that it is not a definitive diagnosis. Begin the revised response with ``Revised response:''.
\end{tcolorbox}
\end{table*}

%% file: tables/appdx_OCR_based_instruction_example.tex
\begin{table*}[t]
\centering
\caption{An example of the synthetic OCR-based instruction sample.}
\begin{tabular}{p{0.39\textwidth}|p{0.55\textwidth}}
\hline
\textbf{Input} & \textbf{Output} \\
\hline
\begin{tabular}[t]{@{}|p{0.35\textwidth}|@{}}
\hline 
In the following hybrid combinations, which one belongs to a test cross ( )\\
A. EeFfGg×EeFfGg\\
B. EeFfGg×eeFfGg\\
C. eeffGg×EeFfGg\\
D. eeffggxEeFfGg\\
\hline
\end{tabular}

Explain the question in the photograph
& \begin{tabular}[t]{@{}p{0.55\textwidth}@{}}
Looking at the options:\\
A. EeFfGg×EeFfGg: It is a cross between two heterozygotes.\\
B. EeFfGg×eeFfGg: It is a cross between a heterozygote and an individual that is homozygous recessive for one gene.\\
C. eeffGg×EeFfGg: It is a cross between an individual that is homozygous recessive for two genes and a heterozygote.\\
D. eeffgg×EeFfGg: It is a cross between an individual that is homozygous recessive for all three genes and a heterozygote.\\
The correct answer is D
\end{tabular}\\
\hline
\end{tabular}
\label{tbl:ocr-example}
\end{table*}

%% file: tables/appdx_self_instruct_prompt.tex
\begin{table*}[t]
\caption{The prompt used for synthesizing medical VQA data in our self-instruct based method.}
\label{tab:prompt-self-instruct-vqa-synthesis}
\begin{tcolorbox}[colframe=orange!60, colback=orange!20, fonttitle=\bfseries\large, coltitle=black, boxrule=0.75mm, arc=5mm, auto outer arc, width=\textwidth,toptitle=6pt, bottomtitle=6pt]
\small
\setstretch{1.2} %
Given a medical image and its description, generate a clinically relevant multiple-choice question that could be asked to a medical professional.\\
\\
If the given medical image matches its description, your output should be in JSON format with three keys:\\
- ``question": A medically appropriate question related to the image.\\
- ``answer": Correct answer to the question.\\
- ``wrong answers": A list of wrong but relevant answers that could be given by the medical professional.\\
There should be 1 to 3 options. The format is [``answer1", ``answer2", ...]\\
If the given medical image does not match its description, simply output "Error: the input description does not match the input image."\\
The question should be directly answerable from the image. \\
\\
Below are some examples of questions and answers for other images: \\
\\
\{seed\_example\_1\} \\
\{seed\_example\_2\} \\
\\
Now, use the following image and the image caption to generate the question: \\
Image caption: \{caption\}
\end{tcolorbox}
\end{table*}

%% file: tables/appdx_CoT_reasoning_open.tex
\begin{table*}[t]
\caption{The prompt used for annotating CoT reasoning process for open-ended VQA data.}
\label{tab:cot-annotation-qa}
\begin{tcolorbox}[colframe=orange!60, colback=orange!20, fonttitle=\bfseries\large, coltitle=black, boxrule=0.75mm, arc=5mm, auto outer arc, width=\textwidth,toptitle=6pt, bottomtitle=6pt]
\small
\setstretch{1.2} %
You are a medical expert specializing in interpreting medical images. The user will provide you with the following inputs:\\
- A medical image\\
- A question about the image\\
- The groundtruth answer\\
\\
Your task is to simulate a step-by-step diagnostic reasoning process that leads to the ground-truth answer. However, you must assume you do not know the ground-truth answer in advance, and you should not reference it explicitly during your reasoning. Please do not give empty output. \\
\\
- Carefully analyze both the image and the question.\\
- Reason step by step, as a medical expert would, using visual and textual cues to reach a logical conclusion.\\
- The reasoning should be concise, clear, and clinically sound.\\
- After the reasoning, reveal the correct answer by copying the ground-truth answer.\\
\\
Output in JSON Format:\\
\{\\
	``reasoning'': ``\textless step-by-step reasoning process here\textgreater'',\\
	``answer'': ``\textless groundtruth answer\textgreater''\\
\}\\
\\
Question: \{question\}\\
Groundtruth answer: \{answer\}
\end{tcolorbox}
\end{table*}

%% file: tables/appdx_CoT_reasoning_MCQA.tex
\begin{table*}[t]
\caption{The prompt used for annotating CoT reasoning process for multiple-choice VQA data.}
\label{tab:cot-annotation-mcq}
\begin{tcolorbox}[colframe=orange!60, colback=orange!20, fonttitle=\bfseries\large, coltitle=black, boxrule=0.75mm, arc=5mm, auto outer arc, width=\textwidth,toptitle=6pt, bottomtitle=6pt]
\small
\setstretch{1.2} %
You are a medical expert specializing in interpreting medical images. The user will provide you with the following inputs:\\
- A medical image \\
- A multiple-choice question related to the image \\
- A list of answer options \\
- The groundtruth answer \\
\\
Your task is to simulate a step-by-step diagnostic reasoning process that leads to the ground-truth answer. However, you must assume you do not know the ground-truth answer in advance, and you should not reference it explicitly during your reasoning. Please do not give empty output. \\
\\
- Carefully analyze both the image and the question.\\
- Evaluate each answer option logically and methodically.\\
- Reason step by step, as a medical expert would, using visual and textual cues to reach a logical conclusion.\\
- The reasoning should be concise, clear, and clinically sound.\\
- After the reasoning, reveal the correct answer by copying the ground-truth answer exactly.\\
\\
Output in JSON Format:\\
\{\\
	``reasoning'': ``\textless step-by-step reasoning process here\textgreater'',\\
	``answer'': ``\textless groundtruth answer\textgreater''\\
\}\\
\\
Question:\\
\{question\}\\
\\
Options:\\
\{options\}\\
\\
Groundtruth answer:\\
\{answer\}
\end{tcolorbox}
\end{table*}

%% file: tables/appdx_CoT_reasoning_quality_check.tex
\begin{table*}[t]
\caption{The prompt used for checking the quality of generated CoT reasoning traces.}
\label{tab:cot-quality-check}
\begin{tcolorbox}[colframe=orange!60, colback=orange!20, fonttitle=\bfseries\large, coltitle=black, boxrule=0.75mm, arc=5mm, auto outer arc, width=\textwidth,toptitle=6pt, bottomtitle=6pt]
\small
\setstretch{1.2} %
You are a medical expert. The user will provide you with the following inputs:\\
- A question related to an image.\\
- A step-by-step diagnostic reasoning process that leads to a predicted answer.\\
- The groundtruth answer.\\
\\
Your task is to check whether the predicted answer in the reasoning process is consistent with the ground-truth answer. \\
If the predicted answer in the reasoning process is inconsistent with the ground-truth answer. Please first output ``Inconsistent. '' Then give a short explanation. \\
Otherwise, please first output ``Consistent. '' Then give a short explanation. \\
\\
Question: \{question\}\\
\\
Reasoning process: \{reasoning\}\\
\\
Groundtruth answer: \{answer\}
\end{tcolorbox}
\end{table*}

%% file: tables/appdx_fact_annotation_long_cap.tex
\begin{table*}[t]
\caption{The prompt used for the factual knowledge annotation stage of medical long caption synthesis.}
\label{tab:medical-long-caption-prompts-stage-3}
\centering
\resizebox{.95\textwidth}{!}{
\begin{tcolorbox}[colframe=orange!60, colback=orange!20, fonttitle=\bfseries\large, coltitle=black, boxrule=0.75mm, arc=5mm, auto outer arc, width=\textwidth,toptitle=6pt, bottomtitle=6pt]
\small
\setstretch{1.2} %
You are provided with a biomedical image from a medical dataset, the disease type (or organ name if there is no disease) of the dataset (`Disease or organ'), the medical Knowledge of the disease (`Knowledge'), a coarse caption (`Coarse Caption') of the image and the medical imaging methods (`Modality'). Your task is to answer the following questions based on the image, bounding box (if provided), caption, disease type and disease knowledge, and condense your answers into caption-styled text.\\
\#\#\# question1\\
Give me a detailed description of the image, including type of the image, organs in the image, approximate location of these organs and relavant locations of these organs and any medical devices (if present) visible in the image as detailedly as possible. Note when answering question1:\\
1. You can use all the meta infomation I provided. But never mention that this information has been given. Your should NEVER mention anything about the bounding box like the contour itself and its outline. Assuming you are introducing to someone who can only see the original image.\\
2. If there is some statistical information provided, like the damage area or the relative damage ratio, etc., you should mention it in your response.\\
3. Not all disease knowledge is relevant to this image; only utilize disease knowledge pertinent to the condition depicted in this image for analysis.\\
4. The coarse caption may not explicitly describe the image, for example, there may appear multiple organs in the caption. You should utilize your knowledge to figure out the most ONE organ and ONE disease to give your description.\\
5. Do not explain or emphasize your analysis. Do not suggest or indicate any consequent effects of the disease.\\
\#\#\# question2 Specify the specific location of the disease or the organ in the image and its relative position to other reference objects in the image. Describe what is unusual in this area indicating the disease (color, texture, size and other features). Note when answering question2: \\
1. You can use information from `Coarse Caption' to locate the disease or the organ. If no,  you need to locate the disease or organ by yourself.\\
2. The image may contain multiple bounding boxs, and the contents of these contours may not necessarily represent the affected areas. Therefore, you need to first answer the questions based on the contents within each bounding box. Afterward, analyze the location of the disease based on your answers.\\
3. Do not use phrase ``bounding box'' in your response. The bounding box if provided is only a visual aid for your reference. Do not say something correlates with or matches any of provided information. Assuming you are introducing to someone who cannot see the bounding box and the provided textual information. If you want to mention the bounding box, use the position of the bounding box (e.g. the top-left side of the image). Do not contain phrases ``caption'', ``medical annotation'', ``medical knowledge''.\\
4. Do not say anything that is not needed in your analysis, like introduction of the disease and medical equipments.\\
5. Do not explain or emphasize your analysis.\\
\#\#\# question3 What may be the relationship between the content in the bounding box and other regions (others being cause of the disease/jointly affected by the diseases/one affect the others/relative positional relationships)? Why and is it possible? Note when answering question3:\\
1. Utilize external knowledge, if possible, to choose relationships and give necessary analysis.\\
2. You can only give an explanation to your choice within two sentence.\\
3. Do not summarize what you've said.\\
4. Do not emphasize your analysis.\\
\#\#\# Integrate Information\\
Describe your answers in a descriptive sentence, not in a "Question-Answer" style. Combine and slightly shorten your answers to the above three questions into a coherent text, keeping as much information of your answers as possible.\\
Note when integrating information and outputing your response:\\
1. Don't respond saying you're unable to assist with requests.\\
2. You should only output your combined and shorteded text in the "caption" filed.
\end{tcolorbox}}
\end{table*}

%% file: tables/appdx_doctor_preference_ct_mri_xray_long_cap.tex
\begin{table*}[t]
\caption{The prompts used for the annotation with doctor preference stage of medical long caption synthesis for MRI, X-ray, and CT images.}
\label{tab:medical-long-caption-prompts-stage-4-1}
\textbf{MRI instruction:}
\begin{tcolorbox}[colframe=orange!60, colback=orange!20, fonttitle=\bfseries\large, coltitle=black, boxrule=0.75mm, arc=5mm, auto outer arc, width=\textwidth,toptitle=6pt, bottomtitle=6pt]
\small
\setstretch{1.2} %
Write a description of the MRI image to include the following information, if these information are visually discernible from the image:\\
1. The sequence used (e.g., T1-weighted, T2-weighted, FLAIR, etc.)\\
2. Plane/orientation of the image (axial, coronal, sagittal) \\
3. Describe the anatomical structures visible and their normal or abnormal appearance (e.g., brain lobes, ventricles, midline structures).\\
4. Identify and describe any abnormal findings, shape, location and effects (e.g. compression)\\
5. Describe signal intensity: hyperintense, hypointense, or isointense
\end{tcolorbox}
\textbf{X-ray instruction:}
\begin{tcolorbox}[colframe=orange!60, colback=orange!20, fonttitle=\bfseries\large, coltitle=black, boxrule=0.75mm, arc=5mm, auto outer arc, width=\textwidth,toptitle=6pt, bottomtitle=6pt]
\small
\setstretch{1.2} %
Write a description of the X-ray image to include the following information, if these information are visually discernible from the image:\\
1. Specific area of the body being imaged (e.g. chest area, lungs, knee, femur, etc.) \\
2. Body part or organ being examined, either fully or partially\\
3. Anatomical planes (e.g. Axial, Sagittal, and Coronal), specific X-ray imaging projections (e.g. AP, PA, lateral view, oblique view, etc.), and clearly discernible left vs right side (either clearly labeled in the image, or clearly determined from image). \\
4. Abnormalities that are clearly observed in the image (e.g. fractures, definite osteophytes, effusions, miliary patterns, vascular markings, diaphragm depression, air bronchograms, thickened interlobular septa, mediastinum shifts, deep sulcus sign, etc.), acute trauma signs (e.g. fractures), implants or other foreign bodies.\\
5. Gender and age-range of patient if clearly discernible (e.g. pediatric vs. adult anatomy)\\
6. When 2 (or a series of) images  of the same patient taken across time are provided, describe the clearly discernible difference.\\
7. Identify and describe any abnormal findings, shape, location and effects (e.g. unexpected wires, prostheses, etc.)
\end{tcolorbox}
\textbf{CT instruction:}
\begin{tcolorbox}[colframe=orange!60, colback=orange!20, fonttitle=\bfseries\large, coltitle=black, boxrule=0.75mm, arc=5mm, auto outer arc, width=\textwidth,toptitle=6pt, bottomtitle=6pt]
\small
\setstretch{1.2} %
Write a description of the CT image to include the following information, if these information are visually discernible from the image:\\
1. The plane/orientation of the image (axial, coronal, sagittal).\\
2. Contrast usage: whether contrast material is used (contrast-enhanced) or not (non-contrast).\\
3. Describe the anatomical structures visible and their normal or abnormal appearance (e.g., organs, bones, blood vessels).\\
4. Identify and describe any abnormal findings including their location, size, shape, and potential effects on surrounding structures.\\
4. Assess and describe the density or attenuation of the tissues and structures as hypoattenuating (darker), isoattenuating (similar), or hyperattenuating (brighter) relative to surrounding tissues.
\end{tcolorbox}
\end{table*}

%% file: tables/appdx_doctor_preference_hist_lesion_knee_long_cap.tex
\begin{table*}[t]
\caption{The prompts used for the annotation with doctor preference stage of medical long caption synthesis for histopathology, skin lesion, and knee xray images. }
\label{tab:medical-long-caption-prompts-stage-4-2}
\textbf{Histopathology instruction:}
\begin{tcolorbox}[colframe=orange!60, colback=orange!20, fonttitle=\bfseries\large, coltitle=black, boxrule=0.75mm, arc=5mm, auto outer arc, width=\textwidth,toptitle=6pt, bottomtitle=6pt]
\small
\setstretch{1.2} %
Write a description of the histopathology image to include the following information, if these information are visually discernible from the image:\\
Tissue Type: Identify the tissue or organ from which the sample is taken (e.g., liver, lung, skin).\\
Staining Technique: Specify the staining method used (e.g., Hematoxylin and Eosin (H\&E), PAS, Trichrome).\\
Cellular Morphology: Describe the appearance of the cells, including size, shape, and organization (e.g., normal, dysplastic, neoplastic).\\
Tissue Architecture: Assess the arrangement and structural organization of the tissue (e.g., normal, inflamed, fibrotic).\\
Abnormal Findings: Identify and describe any pathological changes, such as the presence of tumors, necrosis, inflammation, or infections.\\
Distribution and Patterns: Note the distribution of abnormal findings within the tissue (e.g., focal, diffuse) and any specific patterns (e.g., glandular, trabecular).\\
Special Features: Mention any specific histological features, such as mitotic figures, inclusion bodies, or extracellular deposits.
\end{tcolorbox}
\textbf{Skin lesion instruction:}
\begin{tcolorbox}[colframe=orange!60, colback=orange!20, fonttitle=\bfseries\large, coltitle=black, boxrule=0.75mm, arc=5mm, auto outer arc, width=\textwidth,toptitle=6pt, bottomtitle=6pt]
\small
\setstretch{1.2} %
Write a description of the following skin lesion image, and your description should include the following information if they are clearly discernible from the image.\\
1. region: The potential area of the body where the lesion or wound has been examined.\\
2. general skin texture and hair growth.\\
3. lesions: size (if scale is available in the image), shape, definition, color, texture.\\
4. elevation: Description of the lesion or wound relative to the skin surface of the patient.\\
5. skin texture surrounding the lesion (e.g. coarse/thickness/atrophic/erythema/bleeding, etc)
\end{tcolorbox}
\textbf{Knee x-ray instruction:}
\begin{tcolorbox}[colframe=orange!60, colback=orange!20, fonttitle=\bfseries\large, coltitle=black, boxrule=0.75mm, arc=5mm, auto outer arc, width=\textwidth,toptitle=6pt, bottomtitle=6pt]
\small
\setstretch{1.2} %
Write a description of the following knee X-ray image, and your description should include the following information if they are clearly discernible from the image.\\
1. Area of the Body: Identify the knee joint being imaged, specifying whether it is the left or right knee. If both knees are visible, recognize the one being primarily examined.\\
2. Anatomical Structures: Describe the key anatomical structures visible in the X-ray, including the femur, patella, tibia, fibula, and joint space. Note the alignment and position of these bones relative to each other.\\
3. Imaging Projection: Specify the X-ray imaging projection used, such as anteroposterior (AP), lateral, oblique, or skyline view. Note if the image is weight-bearing or non-weight-bearing.\\
4. Bone and Joint Appearance: Assess the appearance of the bones for any signs of fractures, dislocations, or degenerative changes such as osteoarthritis (e.g., joint space narrowing, osteophyte formation, subchondral sclerosis, punched-out erosions). Evaluate the bone density for any signs of osteoporosis, osteopenia, and visible trabecular patterns. \\
5. Soft Tissue Evaluation: Examine the surrounding soft tissue for any swelling, effusion, or calcifications. Where tophi can be seen, include this in the description.\\
6. Other Abnormalities and Pathologies: Identify and describe any other visible abnormalities or pathologies not mentioned above, including visible bone lesions, misalignment, or foreign objects (e.g., surgical hardware, if present).
\end{tcolorbox}
\end{table*}

%% file: tables/appdx_doctor_preference_gast_long_cap.tex
\begin{table*}[t]
\caption{The prompt used for the annotation with doctor preference stage of medical long caption synthesis for gastrointestinal images. }
\label{tab:medical-long-caption-prompts-stage-4-3}
\textbf{Gastrointestinal instruction:}
\begin{tcolorbox}[colframe=orange!60, colback=orange!20, fonttitle=\bfseries\large, coltitle=black, boxrule=0.75mm, arc=5mm, auto outer arc, width=\textwidth,toptitle=6pt, bottomtitle=6pt]
\small
\setstretch{1.2} %
Write a description of the gastrointestinal (GI) tract endoscopic image to include the following information, if these details are visually discernible from the image:\\
1. Region and Specific Area:  Describe the specific region of the upper/lower GI tract being imaged (e.g., esophagus, stomach, small intestine, large intestine, duodenum, appendix, rectum, etc.) and specify the segment or part prominently featured in the image. Where clearly discernible, include the specific GI anatomical landmark(s) in the image, for e.g., distal/proximal esophagus, cardia/fundus/angulus/antrum region of the stomach, the pylorus, Z-line, cecum, splenic/hepatic flexure, transverse/ascending/descending colon, sigmoid, ileocecal valve, or appendiceal orifice.\\
2. Endoscopic GI View: Include the type of view/orientation of the main image, i.e., front-facing, retroflexed, side-viewing, proximal, or distal.\\
3. GI Anatomical Features:  Describe the following structures of the image, namely relating to lumen of the tubular organs along the GI tract, the mucosal/submucosal details (any visible erythema, erosions, ulcers, or masses, and describe the submucosal layers, noting any thickening, edema, or infiltration), as well as other observable normal and abnormal details of folds, vessels and openings.  Detail any abnormal findings, including their size, shape/type, location, and appearance (e.g., polyps, ulcers, strictures, diverticula, inflammation), and discuss their potential impact on the GI tract (e.g., obstruction, compression, perforation).  Signs of resections and dyed sections should also be noted.\\
4. Indentations and compressions: When the image clearly shows indentations like bulges or depressions that are likely to be caused from outside the GI tract, please include these in the description.\\
5. Medical Data and Device: Specify any textual medical data visible in the image, like contrast medium usage, lighting conditions, magnification levels, or timestamp of the image capture. Where medical devices, like endoscopes, probes, endoknives, gastric bands, stents, etc. are clearly visible, include these in your description.
\end{tcolorbox}
\end{table*}

%% file: tables/appdx_doctor_preference_fundus_ultra_long_cap.tex
\begin{table*}[t]
\caption{The prompt used for the annotation with doctor preference stage of medical long caption synthesis for fundus and ultrasound images. }
\label{tab:medical-long-caption-prompts-stage-4-4}
\textbf{Fundus instruction:}
\begin{tcolorbox}[colframe=orange!60, colback=orange!20, fonttitle=\bfseries\large, coltitle=black, boxrule=0.75mm, arc=5mm, auto outer arc, width=\textwidth,toptitle=6pt, bottomtitle=6pt]
\small
\setstretch{1.2} %
Write a description of the fundus image to include the following information, if these details are visually discernible from the image:\\
1. Overall Image Quality: Assess the clarity and quality of the image to ensure that it is suitable for diagnosis (e.g., presence of artifacts, focus, and illumination adequacy).\\
2. Optic Disc: Examine the optic disc for size, shape, color, and cup-to-disc ratio. Note any abnormalities such as swelling (edema), pallor, or unusual cupping that may indicate glaucoma or other optic neuropathies.\\
3. Macula: Evaluate the macula for changes in size, color, or presence of any lesions or deposits like drusen, which may suggest macular degeneration or related conditions.\\
3. Retinal Vessels: Analyze the retinal blood vessels, assessing their caliber, tortuosity, and arteriovenous (AV) crossings. Identify any vessel occlusions, hemorrhages, or microaneurysms that could indicate hypertension, diabetic retinopathy, or vascular disorders.\\
4. Retina and Periphery: Inspect the retina and peripheral areas for any signs of detachment, tears, pigmentary changes, or lesions that may indicate retinal dystrophies or detachment.\\
5. Abnormal Findings: Identify and describe any abnormal findings such as cotton wool spots, hard exudates, retinal hemorrhages, or neovascularization. Mention their location, size, shape, and potential clinical significance.\\
6. Choroid and Sclera: Examine the visibility of the choroid and any atypical features of the sclera, such as thinning or scleral crescents.\\
7. Specific Patterns or Features: Note any specific patterns or features, such as star-shaped macular exudates or the presence of optic disc drusen, which may have diagnostic relevance.
\end{tcolorbox}
\textbf{Ultrasound instruction:}
\begin{tcolorbox}[colframe=orange!60, colback=orange!20, fonttitle=\bfseries\large, coltitle=black, boxrule=0.75mm, arc=5mm, auto outer arc, width=\textwidth,toptitle=6pt, bottomtitle=6pt]
\small
\setstretch{1.2} %
Write a description of the ultrasound image to include the following information, if these information are visually discernible from the image:
1. The orientation or plane of the image if applicable (e.g., Left and right intercostal scan, left and right subcostal margin scan, longitudinal scan, transverse scan, FAST- Right upper quadrant, FAST- Left upper quadrant and FAST- Heart).
2. Echogenicity of tissues: Describe the echogenicity of the anatomical structures (e.g., hyperechoic, hypoechoic, anechoic, isoechoic) compared to surrounding tissues.
3. Anatomical Structures: Identify and describe the normal or abnormal appearance of organs, vessels, and other structures.
4. Abnormal Findings: Point out and describe any visible abnormalities, including their size, shape, location, and potential effects on nearby tissues (e.g., masses, cysts, fluid collections).
5. Doppler Imaging: If Doppler is used, describe findings related to blood flow, such as presence of turbulence, direction, or any abnormalities.
6. Additional Features: Note any additional features such as the presence of artifacts (e.g., shadowing, enhancement) or any other relevant observations.
\end{tcolorbox}
\end{table*}

%% file: tables/appdx_summarization_doctor_prefer_long_cap.tex
\begin{table*}[t]
\caption{The prompt used for the summarization stage of medical long caption synthesis. The variable \{doctor\_preferred\_features\} in the prompt should be replaced by the doctor preferred features in Tables~\ref{tab:medical-long-caption-prompts-stage-5-1} and \ref{tab:medical-long-caption-prompts-stage-5-2} according to the type of the source image.}
\label{tab:medical-long-caption-prompts-stage-5}
\begin{tcolorbox}[colframe=orange!60, colback=orange!20, fonttitle=\bfseries\large, coltitle=black, boxrule=0.75mm, arc=5mm, auto outer arc, width=\textwidth,toptitle=6pt, bottomtitle=6pt]
\small
\setstretch{1.2} %
You are provided with a biomedical image (which may contain multiple bounding boxes emphasizing the related disease or organs), the disease type (or organ name if there is no disease) of the dataset (`Disease or organ'), the medical knowledge of the disease (`Knowledge'), a coarse caption (`Coarse Caption') of the image, and the medical imaging method used to create the image (`Modality'). In addition, detailed captions for many aspects of the medical image (`Detailed Captions') are also provided.  Your task is to combine the above detailed captions into a long caption for the entire image.\\
1. You can also use the coarse caption, disease type, image modality and disease knowledge, which may contain additional useful information. But never mention that this information has been given.\\
2. Caption 0 in `Detailed Captions' is a model-generated description, including some doctor-preferred features: \{doctor\_preferred\_features\}. Caption 1 focus on specific diseases and are annotated by humans, which are more trustworthy. Consequently, if any pieces of information in Caption 0 does not contradict Caption 1 or is not mentioned in Caption 1, please keep them, especially the location, the size and the name of the disease. If there is a contradiction, you should only keep the information from Caption 1. \\
3. `Detailed Captions' may not explain all bounding boxes. Sometimes some small ones are not mentioned. In this case, in addition to combining the above detailed captions, please also provide explanations for these unmentioned smaller bounding boxes.\\
4. You should NEVER mention anything about the bounding box, like the contour itself or its outline. The bounding box, if present, is only a visual aid for your reference.  Assume you are introducing the image to someone who cannot see the bounding box.\\
5. If there is some statistical information provided, like the damage area or the relative damage ratio, etc., you should mention it in your response.\\
6. For some critical information, like the damage area(s) and its position(s), you should explain one by one in the combined caption.\\
7. Not all disease knowledge is relevant to this image; only utilize disease knowledge pertinent to the condition depicted in this image for analysis.\\
8. Do not explain or emphasize your analysis. Do not suggest or indicate any consequent effects of the disease.\\
9. Do not contain phrases ``caption'', ``medical annotation'', ``medical knowledge''.\\
10. Do not say anything that is not needed in your analysis, like introduction of the disease and medical treatments.\\
11. Do not explain or emphasize your analysis.\\
12. For CT MRI, Xray, GI tract or ultrasound image, if it contains multiple organs, please organize your description organ by organ. You should also include other organs information even if they are healthy. For Histopathology image, you should first describe the overall information, then go deep into the specific areas.\\
13. When integrating the information and outputting your response: (1) Don't respond by saying you're unable to assist with the requests. (2) You should only output your combined text in the ``caption'' field.
\end{tcolorbox}
\end{table*}

%% file: tables/appdx_summarization_ct_mri_xray_hist_skin_long_cap.tex
\begin{table*}[t]
\caption{Doctor preferred features in the summarization stage of medical long caption synthesis for MRI, X-ray, CT, histopathology, skin, and knee X-ray images. }
\label{tab:medical-long-caption-prompts-stage-5-1}
\textbf{Doctor preferred features for MRI images:}
\begin{tcolorbox}[colframe=orange!60, colback=orange!20, fonttitle=\bfseries\large, coltitle=black, boxrule=0.75mm, arc=5mm, auto outer arc, width=\textwidth,toptitle=6pt, bottomtitle=6pt]
\small
\setstretch{1.2} %
1. The sequence used; 2. Plane/orientation of the image; 3. The anatomical structures visible and their normal or abnormal appearance; 4. Any abnormal findings, shape, location and effects; 5. Aignal intensity
\end{tcolorbox}
\textbf{Doctor preferred features for X-ray images:}
\begin{tcolorbox}[colframe=orange!60, colback=orange!20, fonttitle=\bfseries\large, coltitle=black, boxrule=0.75mm, arc=5mm, auto outer arc, width=\textwidth,toptitle=6pt, bottomtitle=6pt]
\small
\setstretch{1.2} %
1. Specific area of the body being imaged; 2. Body part or organ being examined; 3. Anatomical planes, specific X-ray imaging projections, and left vs right side; 4. Abnormalities that are clearly observed in the image, acute trauma signs, implants or other foreign bodies; 5. Gender and age-range of patient; 6. Any abnormal findings, shape, location and effects
\end{tcolorbox}
\textbf{Doctor preferred features for CT images:}
\begin{tcolorbox}[colframe=orange!60, colback=orange!20, fonttitle=\bfseries\large, coltitle=black, boxrule=0.75mm, arc=5mm, auto outer arc, width=\textwidth,toptitle=6pt, bottomtitle=6pt]
\small
\setstretch{1.2} %
1. The plane/orientation of the image; 2. Contrast usage; 3. Anatomical structures visible and their normal or abnormal appearance; 4. Any abnormal findings including their location, size, shape, and potential effects on surrounding structures; 4. The density or attenuation of the tissues and structures relative to surrounding tissues.
\end{tcolorbox}
\textbf{Doctor preferred features for histopathology images:}
\begin{tcolorbox}[colframe=orange!60, colback=orange!20, fonttitle=\bfseries\large, coltitle=black, boxrule=0.75mm, arc=5mm, auto outer arc, width=\textwidth,toptitle=6pt, bottomtitle=6pt]
\small
\setstretch{1.2} %
1. Tissue Type; 2. Staining Technique; 3. Cellular Morphology; 4.Tissue Architecture; 5. Abnormal Findings; 6. The distribution of abnormal findings within the tissue; 7. Special Histological Features
\end{tcolorbox}
\textbf{Doctor preferred features for skin images:}
\begin{tcolorbox}[colframe=orange!60, colback=orange!20, fonttitle=\bfseries\large, coltitle=black, boxrule=0.75mm, arc=5mm, auto outer arc, width=\textwidth,toptitle=6pt, bottomtitle=6pt]
\small
\setstretch{1.2} %
1. region: The area of the body where the lesion or wound has been examined; 2. general skin texture and hair growth; 3. lesion size, shape, definition, color, texture; 4. elevation: Description of the lesion or wound relative to the skin surface of the patient; 5. skin texture surrounding the lesion
\end{tcolorbox}
\textbf{Doctor preferred features for knee X-ray images:}
\begin{tcolorbox}[colframe=orange!60, colback=orange!20, fonttitle=\bfseries\large, coltitle=black, boxrule=0.75mm, arc=5mm, auto outer arc, width=\textwidth,toptitle=6pt, bottomtitle=6pt]
\small
\setstretch{1.2} %
1. Area of the Body: the left or right knee, or both, 2. Anatomical Structures: the key anatomical structures visible in the X-ray, including the femur, patella, tibia, fibula, and joint space; 3. Imaging Projection; 4. Bone and Joint Appearance: any signs of fractures, dislocations, or degenerative changes such as osteoarthritis; 5. the bone density; 5. Soft Tissue: the surrounding soft tissue for any swelling, effusion, or calcifications; 6. Abnormalities and Pathologies: any visible abnormalities or pathologies such as fractures, bone lesions, misalignment, or foreign objects
\end{tcolorbox}
\end{table*}

%% file: tables/appdx_summarization_gast_fundus_ultra_long_cap.tex
\begin{table*}[t]
\caption{Doctor preferred features in the summarization stage of medical long caption synthesis for gastrointestinal, fundus, and ultrasound images. }
\label{tab:medical-long-caption-prompts-stage-5-2}
\textbf{Doctor preferred features for gastrointestinal images:}
\begin{tcolorbox}[colframe=orange!60, colback=orange!20, fonttitle=\bfseries\large, coltitle=black, boxrule=0.75mm, arc=5mm, auto outer arc, width=\textwidth,toptitle=6pt, bottomtitle=6pt]
\small
\setstretch{1.2} %
1. The specific region of the GI tract being imaged; 2. The orientation, perspective and angle of the image; 3. Normal anatomical structures visible in the image; 4. Abnormal findings including their size, shape, location, and appearance; 5. Mucosal and Submucosal Details: The mucosal surface for anomalies; 6. Bowel Wall Characteristics: The thickness and consistency of the bowel wall and note any signs of mural stratification or fat stranding; 7. Surrounding Tissues and Organs: Any visible structures adjacent to the GI tract including any abnormalities extending beyond the GI tract itself; 8. Medical Data and Device: Medical devices, instruments, or tools visible in the image, such as endoscopes, probes, or surgical instruments. Textual medical data visible in the image, like contrast medium usage, lighting conditions, magnification levels, or timestamp of the image capture.
\end{tcolorbox}
\textbf{Doctor preferred features for fundus images:}
\begin{tcolorbox}[colframe=orange!60, colback=orange!20, fonttitle=\bfseries\large, coltitle=black, boxrule=0.75mm, arc=5mm, auto outer arc, width=\textwidth,toptitle=6pt, bottomtitle=6pt]
\small
\setstretch{1.2} %
1. Overall Image Quality: The clarity and quality of the image; 2. Optic Disc: The optic disc for size, shape, color, and cup-to-disc ratio; 3. Macula: The macula for changes in size, color, or presence of any lesions or deposits like drusen; 3. Retinal Vessels: The retinal blood vessels, assessing their caliber, tortuosity, and arteriovenous (AV) crossings; 4. Retina and Periphery: The retina and peripheral areas for any signs of detachment, tears, pigmentary changes, or lesions that may indicate retinal dystrophies or detachment. 5. Abnormal Findings: Any abnormal findings such as cotton wool spots, hard exudates, retinal hemorrhages, or neovascularization; 6. Choroid and Sclera: The visibility of the choroid and any atypical features of the sclera; 7. Specific Patterns or Features: Any specific patterns or features, such as star-shaped macular exudates or the presence of optic disc drusen.
\end{tcolorbox}
\textbf{Doctor preferred features for ultrasound images:}
\begin{tcolorbox}[colframe=orange!60, colback=orange!20, fonttitle=\bfseries\large, coltitle=black, boxrule=0.75mm, arc=5mm, auto outer arc, width=\textwidth,toptitle=6pt, bottomtitle=6pt]
\small
\setstretch{1.2} %
1. The orientation or plane of the image if applicable. 2. Echogenicity of tissues. 3. Normal or abnormal appearance of organs, vessels, and other structures. 4. Visible abnormalities, including their size, shape, location, and potential effects on nearby tissues. 5. Doppler Imaging: If Doppler is used, describe findings related to blood flow.
\end{tcolorbox}
\end{table*}

%% file: tables/appdx_data_mixture_4_training_stages.tex
\begin{table*}[t]
\caption{The overview of data mixture for the four training stages. Note all datasets listed below have been rigorously processed through our data pipeline (detailed in \S~II), resulting in cleaned and validated versions rather than raw data.}
    \centering
    \adjustbox{max width=1.0\textwidth}{
    \setlength{\tabcolsep}{1mm}{
    \begin{tabular}{lp{13cm}l}
    \toprule
        \textbf{Stage} & \textbf{Training Data Composition} & \textbf{Amount}  \\
        \midrule
        \multirow{1}{*}{\textbf{Medical Shallow Alignment}} & PMC-OA and ROCO & $\sim$927K \\ \midrule
        \multirow{3}{*}{\textbf{Medical Deep Alignment}}& \textbf{1. Medical Image Captions:}
        ROCOv2, LLaVA-Med, Quilt-LLaVA, PubMedVision, MIMIC-CXR, FairVLMed, MedICaT, MedPix-2.0, and \textit{synthesized long-form medical caption data}
        \newline
        \textbf{2. General Image Captions:}
        LLaVA-1.5 Caption and PixMo  & $\sim$4.1M \\  \midrule
        \multirow{6}{*}{\textbf{Medical Instruction Tuning}} & \textbf{1. Medical Multimodal:} 
        PathVQA, PMC-VQA, SLAKE, Quilt-LLaVA, VQA-Med-2019, MIMIC-Ext-MIMIC-CXR-VQA, PubMedVision, LLaVA-Med (en \& zh), VQA-RAD, CheXpert Plus, MIMIC-CXR, ROCOv2, IU-Xray, and \textit{synthesized medical QA, OCR-based, and CoT data detailed in \S~II-B}
        \newline 
        \textbf{2. General Multimodal:} 
        LLaVA-1.5 instruction (multimodal instruct subset only) and ALLaVA
        \newline 
        \textbf{3. Medical Text:}  MedQuAD, medical-o1-verifiable, medical-o1-reasoning, ApolloCorpus, Medical-R1-Distill, MedThoughts, icliniq, AlpaCare-MedInstruct, HealthCareMagic,  PMC-LLaMA, HuatuoGPT2-GPT4, MedReason, MedQA, and \textit{synthesized medical text QA detailed in \S~II-B} 
        \newline 
        \textbf{4. General Text:} 
        LLaVA-1.5 instructions (textual instruct subset only) and OpenHermes-2.5 & $\sim$7.1M\\  \midrule
        
        \multirow{2}{*}{\textbf{Medical-oriented RL}} & Data curated from MIMIC-Ext-MIMIC-CXR-VQA, PMC-VQA, Kvasir-VQA, Path-VQA, SLAKE, VQA-Med-2019, VQA-RAD and \textit{synthesized QA} datasets followed by rigorous post-processing & $\sim$100K\\
        \bottomrule
    \end{tabular}}}
    \label{tab:data_mix}
\end{table*}

%% file: tables/appdx_LLM_as_a_Judge.tex
\begin{table*}[t]
\caption{The judgment prompt used for GPT-4.1 in the LLM-as-a-Judge module.}
\label{tab:prompt_for_llm_judge}
\begin{tcolorbox}[colframe=orange!60, colback=orange!20, fonttitle=\bfseries\large, coltitle=black, boxrule=0.75mm, arc=5mm, auto outer arc, width=\textwidth,toptitle=6pt, bottomtitle=6pt]
\small
\setstretch{1.2} %
\texttt{Your task is to determine whether the user's answer is correct based on the provided questions and standard answers (for example, if the user expresses a similar meaning to the standard answer, or another interpretation of the standard answer, it is considered correct.) } \\
The question is: \texttt{\{question\}} \\
The standard answer: \texttt{\{answer\}} \\
The user's answer: \texttt{\{response\}} \\
Please strictly follow the following format for output(0 represents correct, 1 represents incorrect): \\
\textless think\textgreater{{your concise think step}}\textless /think\textgreater \\
\textless judge\textgreater{{0/1}}\textless /judge\textgreater \\
For example: \\
\textless think\textgreater The standard answer is right, and the user's answer is right frontal lobe, they express the same meaning, so it is correct.\textless /think\textgreater
\textless judge\textgreater0\textless /judge\textgreater
\end{tcolorbox}
\end{table*}

%% file: tables/appdx_benchmarks_overview.tex
\begin{table*}[t]
    \caption{The statistics of medical evaluation benchmarks utilized in the \oureval framework.}
    \centering
    \adjustbox{max width=1.0\textwidth}{
    \begin{tabular}{lccccccc}
    \toprule
          \textbf{Benchmark}&\textbf{Category} & \textbf{Multiple-Choice} &\textbf{Close-ended} &\textbf{Open-ended}&\textbf{Report} &\textbf{Questions} &\textbf{Imgs}\\
        \midrule
        VQA-RAD & X-ray,CT,MRI & - & 251 & 200 & - & 451 & 203 \\ 
        \midrule
        SLAKE & X-ray,CT,MRI & - & 836 & 1,258 & - & 2,094 & 180 \\ 
        \midrule
        PathVQA & Mixed & - & 3,362 & 3,357 & - & 6,719 & 858 \\
        \midrule
        PMC-VQA & Mixed & 33,430 & - & - & - & 33,430 & 29,021 \\
        \midrule
        OmniMedVQA & Mixed & 88,996 & -  & - & - & 88,996 & 82,059 \\
        \midrule
        MMMU & Mixed & 1,927 & - & - & - & 1,927 & 1,927 \\
        \midrule
        MedXpertQA & Mixed & 4,500 & - & - & - & 4,500 & 2,852 \\
        \midrule
        MMLU & Text & 1,897 & - & - & - & 1,897 & - \\
        \midrule
        PubMedQA &Text & 500 & - & - & - & 500 & - \\
        \midrule
        MedMCQA &Text & 4,183 & - & - & - & 4,183 & - \\
        \midrule
        MedQA-USMLE & Text & 1,273 & - & - & - & 1,273 & - \\
        \midrule
        MedBullets &Text & 616 & - & - & - & 616 & - \\
        \midrule
        SuperGPQA &Text & 2,755 & - & - & - & 2,755 & - \\
        \midrule
        MIMIC-CXR & X-ray & - & - & - & 2,195 & 2,195 & 3,681 \\
        \midrule
        IU-Xray & X-ray & - & - & - & 296 & 296 & 607 \\
        \midrule
        CheXpert Plus & X-ray & - & - & - & 234 & 234 & 234 \\
        \midrule
        Total & Mixed & 140,077 & 4,449 &  4,815& 2,725 & 152,066 & 121,622\\
        \bottomrule
    \end{tabular}}
    \label{tab:benchmarks_overview}
\end{table*}

%% file: tables/appdx_eval_data_quality.tex
\begin{table*}[t]
\caption{The prompt used for evaluating the quality of \ours training data.}
\label{tab:eval_data_quality}
\begin{tcolorbox}[colframe=orange!60, colback=orange!20, fonttitle=\bfseries\large, coltitle=black, boxrule=0.75mm, arc=5mm, auto outer arc, width=\textwidth,toptitle=6pt, bottomtitle=6pt]
\small
\setstretch{1.2} %
You are an expert evaluator of biomedical and clinical texts. Your task is to score the provided text based on three specific dimensions using a scale of 0 to 5.

\#\#\# Input Text:
``\{text\}''

\#\#\# Scoring Criteria:

1. **Logical Complexity (0-5)**: 0=No logic, 5=Highly complex, multi-step expert reasoning.

2. **Medical/Scientific Vocabulary Density (0-5)**: 0=Layman language, 5=Dense academic/clinical jargon.

3. **Terminology Diversity (0-5)**: 0=No terms, 5=Comprehensive coverage across $>$4 distinct medical/scientific categories.

\#\#\# Output Format:

JSON object only.
\{

  ``logical\_complexity\_score'': $<$int$>$,
  
  ``logical\_reasoning'': ``$<$short explanation$>$'',
  
  ``vocabulary\_density\_score'': $<$int$>$,
  
  ``vocabulary\_reasoning'': ``$<$short explanation$>$'',
  
  ``terminology\_diversity\_score'': $<$int$>$,
  
  ``terminology\_reasoning'': ``$<$short explanation$>$''
  
\}
\end{tcolorbox}
\end{table*}

%% file: paper.bib
@IEEEtranBSTCTL{IEEEtran:BSTcontrol,
CTLuse_forced_etal       = "yes",
CTLmax_names_forced_etal = "3",
CTLnames_show_etal       = "10" }

@inproceedings{zhao2024ratescore,
    title = "{R}a{TES}core: A Metric for Radiology Report Generation",
    author = "Zhao, Weike  and
      Wu, Chaoyi  and
      Zhang, Xiaoman  and
      Zhang, Ya  and
      Wang, Yanfeng  and
      Xie, Weidi",
    booktitle = "EMNLP",
    year = "2024",
}

@article{yu2023evaluating,
  title={Evaluating progress in automatic chest x-ray radiology report generation},
  author={Yu, Feiyang and Endo, Mark and Krishnan, Rayan and Pan, Ian and Tsai, Andy and Reis, Eduardo Pontes and Fonseca, Eduardo Kaiser Ururahy Nunes and Lee, Henrique Min Ho and Abad, Zahra Shakeri Hossein and Ng, Andrew Y and others},
  journal={Patterns},
  year={2023},
}

@inproceedings{smit2020chexbert,
    title = "Combining Automatic Labelers and Expert Annotations for Accurate Radiology Report Labeling Using {BERT}",
    author = "Smit, Akshay  and
      Jain, Saahil  and
      Rajpurkar, Pranav  and
      Pareek, Anuj  and
      Ng, Andrew  and
      Lungren, Matthew",
    booktitle = "EMNLP",
    year = "2020",
}

@inproceedings{vedantam2015cider,
  title={Cider: Consensus-based image description evaluation},
  author={Vedantam, Ramakrishna and Lawrence Zitnick, C and Parikh, Devi},
  booktitle={CVPR},
  year={2015},
}

@inproceedings{lin2004rouge,
    title = "{ROUGE}: A Package for Automatic Evaluation of Summaries",
    author = "Lin, Chin-Yew",
    booktitle = "Text Summarization Branches Out",
    year = "2004",
}

@article{zhang2024rexrank,
  title={ReXrank: A Public Leaderboard for AI-Powered Radiology Report Generation},
  author={Zhang, Xiaoman and Zhou, Hong-Yu and Yang, Xiaoli and Banerjee, Oishi and Acosta, Juli{\'a}n N and Miller, Josh and Huang, Ouwen and Rajpurkar, Pranav},
  journal={ArXiv},
  year={2024},
}

@article{jin2019pubmedqa,
  title={Pubmedqa: A dataset for biomedical research question answering},
  author={Jin, Qiao and Dhingra, Bhuwan and Liu, Zhengping and Cohen, William W and Lu, Xinghua},
  journal={ArXiv},
  year={2019},
}

@inproceedings{kwon2023efficient,
author = {Kwon, Woosuk and Li, Zhuohan and Zhuang, Siyuan and Sheng, Ying and Zheng, Lianmin and Yu, Cody Hao and Gonzalez, Joseph and Zhang, Hao and Stoica, Ion},
title = {Efficient Memory Management for Large Language Model Serving with PagedAttention},
year = {2023},
booktitle = {SOSP},
}

@inproceedings{lmms_eval2024,
    title = "{LMM}s-Eval: Reality Check on the Evaluation of Large Multimodal Models",
    author = "Zhang, Kaichen  and
      Li, Bo  and
      Zhang, Peiyuan  and
      Pu, Fanyi  and
      Cahyono, Joshua Adrian  and
      Hu, Kairui  and
      Liu, Shuai  and
      Zhang, Yuanhan  and
      Yang, Jingkang  and
      Li, Chunyuan  and
      Liu, Ziwei",
    booktitle = "Findings of NAACL",
    year = "2025",
}

@inproceedings{duan2024vlmevalkit,
author = {Duan, Haodong and Yang, Junming and Qiao, Yuxuan and Fang, Xinyu and Chen, Lin and Liu, Yuan and Dong, Xiaoyi and Zang, Yuhang and Zhang, Pan and Wang, Jiaqi and Lin, Dahua and Chen, Kai},
title = {VLMEvalKit: An Open-Source ToolKit for Evaluating Large Multi-Modality Models},
year = {2024},
booktitle = {MM},
}

@article{li2025gmaivl,
      title={GMAI-VL \& GMAI-VL-5.5M: A Large Vision-Language Model and A Comprehensive Multimodal Dataset Towards General Medical AI}, 
      author={Tianbin Li and Yanzhou Su and Wei Li and Bin Fu and Zhe Chen and Ziyan Huang and Guoan Wang and Chenglong Ma and Ying Chen and Ming Hu and Yanjun Li and Pengcheng Chen and Xiaowei Hu and Zhongying Deng and Yuanfeng Ji and Jin Ye and Yu Qiao and Junjun He},
      year={2025},
      journal={ArXiv},
}

@article{chen2024huatuogpto1,
      title={HuatuoGPT-o1, Towards Medical Complex Reasoning with LLMs}, 
      author={Junying Chen and Zhenyang Cai and Ke Ji and Xidong Wang and Wanlong Liu and Rongsheng Wang and Jianye Hou and Benyou Wang},
      year={2024},
      journal={ArXiv},
}

@article{chen2023huatuogptii,
      title={HuatuoGPT-II, One-stage Training for Medical Adaption of LLMs}, 
      author={Junying Chen and Xidong Wang and Anningzhe Gao and Feng Jiang and Shunian Chen and Hongbo Zhang and Dingjie Song and Wenya Xie and Chuyi Kong and Jianquan Li and Xiang Wan and Haizhou Li and Benyou Wang},
      year={2023},
      journal={ArXiv},
}

@article{chen2024huatuogptvision,
      title={HuatuoGPT-Vision, Towards Injecting Medical Visual Knowledge into Multimodal LLMs at Scale}, 
      author={Junying Chen and Chi Gui and Ruyi Ouyang and Anningzhe Gao and Shunian Chen and Guiming Hardy Chen and Xidong Wang and Ruifei Zhang and Zhenyang Cai and Ke Ji and Guangjun Yu and Xiang Wan and Benyou Wang},
      year={2024},
      journal={ArXiv},
}

@article{nath2025vilam3,
      title={VILA-M3: Enhancing Vision-Language Models with Medical Expert Knowledge}, 
      author={Vishwesh Nath and Wenqi Li and Dong Yang and Andriy Myronenko and Mingxin Zheng and Yao Lu and Zhijian Liu and Hongxu Yin and Yucheng Tang and Pengfei Guo and Can Zhao and Ziyue Xu and Yufan He and Greg Heinrich and Yee Man Law and Benjamin Simon and Stephanie Harmon and Stephen Aylward and Marc Edgar and Michael Zephyr and Song Han and Pavlo Molchanov and Baris Turkbey and Holger Roth and Daguang Xu},
      year={2025},
      journal={ArXiv},
}

@article{zhang2025biomedclip,
      title={BiomedCLIP: a multimodal biomedical foundation model pretrained from fifteen million scientific image-text pairs}, 
      author={Sheng Zhang and Yanbo Xu and Naoto Usuyama and Hanwen Xu and Jaspreet Bagga and Robert Tinn and Sam Preston and Rajesh Rao and Mu Wei and Naveen Valluri and Cliff Wong and Andrea Tupini and Yu Wang and Matt Mazzola and Swadheen Shukla and Lars Liden and Jianfeng Gao and Angela Crabtree and Brian Piening and Carlo Bifulco and Matthew P. Lungren and Tristan Naumann and Sheng Wang and Hoifung Poon},
      year={2025},
      journal={ArXiv},
}

@inproceedings{li2023llavamed,
    title={{LL}a{VA}-Med: Training a Large Language-and-Vision Assistant for Biomedicine in One Day},
    author={Chunyuan Li and Cliff Wong and Sheng Zhang and Naoto Usuyama and Haotian Liu and Jianwei Yang and Tristan Naumann and Hoifung Poon and Jianfeng Gao},
    booktitle={NeurIPS},
    year={2023},
}

@article{zhang2024biomedgpt,
   title={A generalist vision–language foundation model for diverse biomedical tasks},
   journal={Nature Medicine},
   author={Zhang, Kai and Zhou, Rong and Adhikarla, Eashan and Yan, Zhiling and Liu, Yixin and Yu, Jun and Liu, Zhengliang and Chen, Xun and Davison, Brian D. and Ren, Hui and Huang, Jing and Chen, Chen and Zhou, Yuyin and Fu, Sunyang and Liu, Wei and Liu, Tianming and Li, Xiang and Chen, Yong and He, Lifang and Zou, James and Li, Quanzheng and Liu, Hongfang and Sun, Lichao},
   year={2024},
}

@article{lu2024pathchat,
      title={A multimodal generative AI copilot for human pathology}, 
      author={Ming Y. Lu and Bowen Chen and Drew F. K. Williamson and Richard J. Chen and Kenji Ikamura and Georg Gerber and Ivy Liang and Long Phi Le and Tong Ding and Anil V Parwani and Faisal Mahmood},
      year={2024},
      volume={634},
      journal={Nature},
}

@InProceedings{moor2023medflamingo,
  title = 	 {Med-Flamingo: a Multimodal Medical Few-shot Learner},
  author =       {Moor, Michael and Huang, Qian and Wu, Shirley and Yasunaga, Michihiro and Dalmia, Yash and Leskovec, Jure and Zakka, Cyril and Reis, Eduardo Pontes and Rajpurkar, Pranav},
  booktitle = 	 {Machine Learning for Health Symposium},
  year = 	 {2023},
}

@inproceedings{ikezogwo2023quilt1m,
title={Quilt-1M: One Million Image-Text Pairs for Histopathology},
author={Wisdom Oluchi Ikezogwo and Mehmet Saygin Seyfioglu and Fatemeh Ghezloo and Dylan Stefan Chan Geva and Fatwir Sheikh Mohammed and Pavan Kumar Anand and Ranjay Krishna and Linda Shapiro},
booktitle={NeurIPS},
year={2023},
}

@article{seyfioglu2025quiltllava,
      title={Quilt-LLaVA: Visual Instruction Tuning by Extracting Localized Narratives from Open-Source Histopathology Videos}, 
      author={Mehmet Saygin Seyfioglu and Wisdom O. Ikezogwo and Fatemeh Ghezloo and Ranjay Krishna and Linda Shapiro},
      year={2025},
      journal={ArXiv},
}

@article{chambon2024chexpertplusaugmentinglarge,
      title={CheXpert Plus: Augmenting a Large Chest X-ray Dataset with Text Radiology Reports, Patient Demographics and Additional Image Formats}, 
      author={Pierre Chambon and Jean Delbrouck and Thomas Sounack and Shihcheng Huang and Zhihong Chen and Maya Varma and Steven Truong and Chu Chuong and Curtis Langlotz},
      year={2024},
      journal={ArXiv},
}

@article{demner2016preparing-iu-xray,
  title={Preparing a collection of radiology examinations for distribution and retrieval},
  author={Demner-Fushman, Dina and Kohli, Marc and Rosenman, Marc and Shooshan, Sonya and Rodriguez, Laritza and Antani, Sameer and Thoma, George and McDonald, Clement},
  journal={JAMIA},
  year={2016},
}

@article{wu2023generalist,
      title={Towards Generalist Foundation Model for Radiology by Leveraging Web-scale 2D\&3D Medical Data}, 
      author={Chaoyi Wu and Xiaoman Zhang and Ya Zhang and Yanfeng Wang and Weidi Xie},
      year={2023},
      journal={ArXiv},
}

@article{hyland2024maira1,
      title={MAIRA-1: A specialised large multimodal model for radiology report generation}, 
      author={Stephanie L. Hyland and Shruthi Bannur and Kenza Bouzid and Daniel C. Castro and Mercy Ranjit and Anton Schwaighofer and Fernando Pérez-García and Valentina Salvatelli and Shaury Srivastav and Anja Thieme and Noel Codella and Matthew P. Lungren and Maria Teodora Wetscherek and Ozan Oktay and Javier Alvarez-Valle},
      year={2024},
      journal={ArXiv},
}

@inproceedings{hu2024omnimedvqa,
    author    = {Hu, Yutao and Li, Tianbin and Lu, Quanfeng and Shao, Wenqi and He, Junjun and Qiao, Yu and Luo, Ping},
    title     = {OmniMedVQA: A New Large-Scale Comprehensive Evaluation Benchmark for Medical LVLM},
    booktitle = {CVPR},
    year      = {2024},
}

@inproceedings{liu2021slake,
  author={Liu, Bo and Zhan, Li-Ming and Xu, Li and Ma, Lin and Yang, Yan and Wu, Xiao-Ming},
  booktitle={ISBI}, 
  title={Slake: A Semantically-Labeled Knowledge-Enhanced Dataset For Medical Visual Question Answering}, 
  year={2021},
}

@inproceedings{xie2025medtrinity,
    title={MedTrinity-25M: A Large-scale Multimodal Dataset with Multigranular Annotations for Medicine},
    author={Yunfei Xie and Ce Zhou and Lang Gao and Juncheng Wu and Xianhang Li and Hong-Yu Zhou and Sheng Liu and Lei Xing and James Zou and Cihang Xie and Yuyin Zhou},
    booktitle={ICLR},
    year={2025},
}

@article{pan2025medvlmr1,
      title={MedVLM-R1: Incentivizing Medical Reasoning Capability of Vision-Language Models (VLMs) via Reinforcement Learning}, 
      author={Jiazhen Pan and Che Liu and Junde Wu and Fenglin Liu and Jiayuan Zhu and Hongwei Bran Li and Chen Chen and Cheng Ouyang and Daniel Rueckert},
      year={2025},
      journal={ArXiv},
}

@article{wu2025medreason,
      title={MedReason: Eliciting Factual Medical Reasoning Steps in LLMs via Knowledge Graphs}, 
      author={Juncheng Wu and Wenlong Deng and Xingxuan Li and Sheng Liu and Taomian Mi and Yifan Peng and Ziyang Xu and Yi Liu and Hyunjin Cho and Chang-In Choi and Yihan Cao and Hui Ren and Xiang Li and Xiaoxiao Li and Yuyin Zhou},
      year={2025},
      journal={ArXiv},
}

@article{lai2025medr1,
      title={Med-R1: Reinforcement Learning for Generalizable Medical Reasoning in Vision-Language Models}, 
      author={Yuxiang Lai and Jike Zhong and Ming Li and Shitian Zhao and Xiaofeng Yang},
      year={2025},
      journal={ArXiv},
}

@article{he2020pathvqa,
      title={PathVQA: 30000+ Questions for Medical Visual Question Answering}, 
      author={Xuehai He and Yichen Zhang and Luntian Mou and Eric Xing and Pengtao Xie},
      year={2020},
      journal={ArXiv},
}

@article{zhang2024pmcvqa,
      title={PMC-VQA: Visual Instruction Tuning for Medical Visual Question Answering}, 
      author={Xiaoman Zhang and Chaoyi Wu and Ziheng Zhao and Weixiong Lin and Ya Zhang and Yanfeng Wang and Weidi Xie},
      year={2024},
      journal={ArXiv},
}

@article{zuo2025medxpertqa,
      title={MedXpertQA: Benchmarking Expert-Level Medical Reasoning and Understanding}, 
      author={Yuxin Zuo and Shang Qu and Yifei Li and Zhangren Chen and Xuekai Zhu and Ermo Hua and Kaiyan Zhang and Ning Ding and Bowen Zhou},
      year={2025},
      journal={ArXiv},
}

@article{lau2018vqarad,
  title={A dataset of clinically generated visual questions and answers about radiology images},
  author={Lau, Jason J and Gayen, Soumya and Ben Abacha, Asma and Demner-Fushman, Dina},
  journal={Scientific data},
  volume={5},
  year={2018},
}

@article{meng2025mmeureka,
      title={MM-Eureka: Exploring the Frontiers of Multimodal Reasoning with Rule-based Reinforcement Learning}, 
      author={Fanqing Meng and Lingxiao Du and Zongkai Liu and Zhixiang Zhou and Quanfeng Lu and Daocheng Fu and Tiancheng Han and Botian Shi and Wenhai Wang and Junjun He and Kaipeng Zhang and Ping Luo and Yu Qiao and Qiaosheng Zhang and Wenqi Shao},
      year={2025},
      journal={ArXiv},
}

@article{wang2025vlrethinker,
      title={VL-Rethinker: Incentivizing Self-Reflection of Vision-Language Models with Reinforcement Learning}, 
      author={Haozhe Wang and Chao Qu and Zuming Huang and Wei Chu and Fangzhen Lin and Wenhu Chen},
      year={2025},
      journal={ArXiv},
}

@inproceedings{chu2025sft,
title={{SFT} Memorizes, {RL} Generalizes: A Comparative Study of Foundation Model Post-training},
author={Tianzhe Chu and Yuexiang Zhai and Jihan Yang and Shengbang Tong and Saining Xie and Sergey Levine and Yi Ma},
booktitle={CPAL},
year={2025},
}

@article{tan2025reasonrft,
      title={Reason-RFT: Reinforcement Fine-Tuning for Visual Reasoning}, 
      author={Huajie Tan and Yuheng Ji and Xiaoshuai Hao and Minglan Lin and Pengwei Wang and Zhongyuan Wang and Shanghang Zhang},
      year={2025},
      journal={ArXiv},
}

@InProceedings{yue2024mmmu,
    author    = {Yue, Xiang and Ni, Yuansheng and Zhang, Kai and Zheng, Tianyu and Liu, Ruoqi and Zhang, Ge and Stevens, Samuel and Jiang, Dongfu and Ren, Weiming and Sun, Yuxuan and Wei, Cong and Yu, Botao and Yuan, Ruibin and Sun, Renliang and Yin, Ming and Zheng, Boyuan and Yang, Zhenzhu and Liu, Yibo and Huang, Wenhao and Sun, Huan and Su, Yu and Chen, Wenhu},
    title     = {MMMU: A Massive Multi-discipline Multimodal Understanding and Reasoning Benchmark for Expert AGI},
    booktitle = {CVPR},
    year      = {2024},
}

@article{lin2025healthgpt,
      title={HealthGPT: A Medical Large Vision-Language Model for Unifying Comprehension and Generation via Heterogeneous Knowledge Adaptation}, 
      author={Tianwei Lin and Wenqiao Zhang and Sijing Li and Yuqian Yuan and Binhe Yu and Haoyuan Li and Wanggui He and Hao Jiang and Mengze Li and Xiaohui Song and Siliang Tang and Jun Xiao and Hui Lin and Yueting Zhuang and Beng Chin Ooi},
      year={2025},
      journal={ArXiv},
}

@article{chen2025internvl25,
      title={Expanding Performance Boundaries of Open-Source Multimodal Models with Model, Data, and Test-Time Scaling}, 
      author={Zhe Chen and Weiyun Wang and Yue Cao and Yangzhou Liu and Zhangwei Gao and Erfei Cui and Jinguo Zhu and Shenglong Ye and Hao Tian and Zhaoyang Liu and Lixin Gu and Xuehui Wang and Qingyun Li and Yimin Ren and Zixuan Chen and Jiapeng Luo and Jiahao Wang and Tan Jiang and Bo Wang and Conghui He and Botian Shi and Xingcheng Zhang and Han Lv and Yi Wang and Wenqi Shao and Pei Chu and Zhongying Tu and Tong He and Zhiyong Wu and Huipeng Deng and Jiaye Ge and Kai Chen and Kaipeng Zhang and Limin Wang and Min Dou and Lewei Lu and Xizhou Zhu and Tong Lu and Dahua Lin and Yu Qiao and Jifeng Dai and Wenhai Wang},
      year={2025},
      journal={ArXiv},
}

@article{zhu2025internvl3,
      title={InternVL3: Exploring Advanced Training and Test-Time Recipes for Open-Source Multimodal Models}, 
      author={Jinguo Zhu and Weiyun Wang and Zhe Chen and Zhaoyang Liu and Shenglong Ye and Lixin Gu and Hao Tian and Yuchen Duan and Weijie Su and Jie Shao and Zhangwei Gao and Erfei Cui and Xuehui Wang and Yue Cao and Yangzhou Liu and Xingguang Wei and Hongjie Zhang and Haomin Wang and Weiye Xu and Hao Li and Jiahao Wang and Nianchen Deng and Songze Li and Yinan He and Tan Jiang and Jiapeng Luo and Yi Wang and Conghui He and Botian Shi and Xingcheng Zhang and Wenqi Shao and Junjun He and Yingtong Xiong and Wenwen Qu and Peng Sun and Penglong Jiao and Han Lv and Lijun Wu and Kaipeng Zhang and Huipeng Deng and Jiaye Ge and Kai Chen and Limin Wang and Min Dou and Lewei Lu and Xizhou Zhu and Tong Lu and Dahua Lin and Yu Qiao and Jifeng Dai and Wenhai Wang},
      year={2025},
      journal={ArXiv},
}

@article{deepseekai2025deepseekr1,
      title={DeepSeek-R1: Incentivizing Reasoning Capability in LLMs via Reinforcement Learning}, 
      author={DeepSeek},
      year={2025},
      journal={ArXiv},
}

@article{nakayama2023brazilian,
	author = {Nakayama, Luis Filipe and Restrepo, David and Matos, Jo{\~a}o and Ribeiro, Lucas Zago and Malerbi, Fernando Korn and Celi, Leo Anthony and Regatieri, Caio Saito},
	title = {BRSET: A Brazilian Multilabel Ophthalmological Dataset of Retina Fundus Photos},
	year = {2024},
	publisher = {Cold Spring Harbor Laboratory Press},
	journal = {PLOS Digital Health}
}

@InProceedings{pal2022medmcqa,
  title = 	 {MedMCQA: A Large-scale Multi-Subject Multi-Choice Dataset for Medical domain Question Answering},
  author =       {Pal, Ankit and Umapathi, Logesh Kumar and Sankarasubbu, Malaikannan},
  booktitle = 	 {CHIL},
  year = 	 {2022},
}

@inproceedings{chen2025medbullets,
    title = "Benchmarking Large Language Models on Answering and Explaining Challenging Medical Questions",
    author = "Chen, Hanjie  and
      Fang, Zhouxiang  and
      Singla, Yash  and
      Dredze, Mark",
    booktitle = "NAACL",
    year = "2025",
}

@article{pteam2025supergpqa,
      title={SuperGPQA: Scaling LLM Evaluation across 285 Graduate Disciplines}, 
      author={P Team and Xinrun Du and Yifan Yao and Kaijing Ma and Bingli Wang and Tianyu Zheng and King Zhu and Minghao Liu and Yiming Liang and Xiaolong Jin and Zhenlin Wei and Chujie Zheng and Kaixin Deng and Shawn Gavin and Shian Jia and Sichao Jiang and Yiyan Liao and Rui Li and Qinrui Li and Sirun Li and Yizhi Li and Yunwen Li and David Ma and Yuansheng Ni and Haoran Que and Qiyao Wang and Zhoufutu Wen and Siwei Wu and Tyshawn Hsing and Ming Xu and Zhenzhu Yang and Zekun Moore Wang and Junting Zhou and Yuelin Bai and Xingyuan Bu and Chenglin Cai and Liang Chen and Yifan Chen and Chengtuo Cheng and Tianhao Cheng and Keyi Ding and Siming Huang and Yun Huang and Yaoru Li and Yizhe Li and Zhaoqun Li and Tianhao Liang and Chengdong Lin and Hongquan Lin and Yinghao Ma and Tianyang Pang and Zhongyuan Peng and Zifan Peng and Qige Qi and Shi Qiu and Xingwei Qu and Shanghaoran Quan and Yizhou Tan and Zili Wang and Chenqing Wang and Hao Wang and Yiya Wang and Yubo Wang and Jiajun Xu and Kexin Yang and Ruibin Yuan and Yuanhao Yue and Tianyang Zhan and Chun Zhang and Jinyang Zhang and Xiyue Zhang and Xingjian Zhang and Yue Zhang and Yongchi Zhao and Xiangyu Zheng and Chenghua Zhong and Yang Gao and Zhoujun Li and Dayiheng Liu and Qian Liu and Tianyu Liu and Shiwen Ni and Junran Peng and Yujia Qin and Wenbo Su and Guoyin Wang and Shi Wang and Jian Yang and Min Yang and Meng Cao and Xiang Yue and Zhaoxiang Zhang and Wangchunshu Zhou and Jiaheng Liu and Qunshu Lin and Wenhao Huang and Ge Zhang},
      year={2025},
      journal={ArXiv},
}

@inproceedings{hendrycks2021mmlu,
title={Measuring Massive Multitask Language Understanding},
author={Dan Hendrycks and Collin Burns and Steven Basart and Andy Zou and Mantas Mazeika and Dawn Song and Jacob Steinhardt},
booktitle={ICLR},
year={2021},
}

@article{jin2020medqausmle,
      title={What Disease does this Patient Have? A Large-scale Open Domain Question Answering Dataset from Medical Exams}, 
      author={Di Jin and Eileen Pan and Nassim Oufattole and Wei-Hung Weng and Hanyi Fang and Peter Szolovits},
      year={2021},
      volume={11},
      journal={Applied Sciences},
}

@article{bai2025qwen25vl,
      title={Qwen2.5-VL Technical Report}, 
      author={Shuai Bai and Keqin Chen and Xuejing Liu and Jialin Wang and Wenbin Ge and Sibo Song and Kai Dang and Peng Wang and Shijie Wang and Jun Tang and Humen Zhong and Yuanzhi Zhu and Mingkun Yang and Zhaohai Li and Jianqiang Wan and Pengfei Wang and Wei Ding and Zheren Fu and Yiheng Xu and Jiabo Ye and Xi Zhang and Tianbao Xie and Zesen Cheng and Hang Zhang and Zhibo Yang and Haiyang Xu and Junyang Lin},
      year={2025},
      journal={ArXiv},
}

@InProceedings{pelka2018roco,
author="Pelka, Obioma
and Koitka, Sven
and R{\"u}ckert, Johannes
and Nensa, Felix
and Friedrich, Christoph M.",
title="Radiology Objects in COntext (ROCO): A Multimodal Image Dataset",
booktitle="LABELS",
year="2018",
}

@article{ruckert2024rocov2,
   title={ROCOv2: Radiology Objects in COntext Version 2, an Updated Multimodal Image Dataset},
   journal={Scientific Data},
   author={Rückert, Johannes and Bloch, Louise and Brüngel, Raphael and Idrissi-Yaghir, Ahmad and Schäfer, Henning and Schmidt, Cynthia S. and Koitka, Sven and Pelka, Obioma and Abacha, Asma Ben and G. Seco de Herrera, Alba and Müller, Henning and Horn, Peter A. and Nensa, Felix and Friedrich, Christoph M.},
   year={2024},
}

@inproceedings{yeo2025demystifying,
title={Demystifying Long Chain-of-Thought Reasoning in {LLM}s},
author={Edward Yeo and Yuxuan Tong and Xinyao Niu and Graham Neubig and Xiang Yue},
booktitle={ICLR Workshop},
year={2025},
}

@article{johnson2019mimiccxr,
  title={MIMIC-CXR, a de-identified publicly available database of chest radiographs with free-text reports},
  author={Johnson, Alistair E. W. and Pollard, Tom J. and Berkowitz, Seth J. and Greenbaum, Nathaniel R. and Lungren, Matthew P. and Deng, Chih-ying and Mark, Roger G. and Horng, Steven},
  journal={Scientific data},
  year={2019},
}

@article{philipp2018ham10000,
  title={The HAM10000 dataset, a large collection of multi-source dermatoscopic images of common pigmented skin lesions},
  author={Philipp, Tschandl and Cliff, Rosendahl and Harald, Kittler },
  journal={Scientific data},
  volume={5},
  year={2018},
}

@InProceedings{liu2023pmcclip,
author="Lin, Weixiong
and Zhao, Ziheng
and Zhang, Xiaoman
and Wu, Chaoyi
and Zhang, Ya
and Wang, Yanfeng
and Xie, Weidi",
title="PMC-CLIP: Contrastive Language-Image Pre-training Using Biomedical Documents",
booktitle="MICCAI",
year="2023",
}

@inproceedings{subramanian2020medicat,
    title = "{M}ed{IC}a{T}: A Dataset of Medical Images, Captions, and Textual References",
    author = "Subramanian, Sanjay  and
      Wang, Lucy Lu  and
      Bogin, Ben  and
      Mehta, Sachin  and
      van Zuylen, Madeleine  and
      Parasa, Sravanthi  and
      Singh, Sameer  and
      Gardner, Matt  and
      Hajishirzi, Hannaneh",
    booktitle = "Findings of EMNLP",
    year = "2020",
}

@inproceedings{luo2024fairclip,
  title={Fairclip: Harnessing fairness in vision-language learning},
  author={Luo, Yan and Shi, Min and Khan, Muhammad Osama and Afzal, Muhammad Muneeb and Huang, Hao and Yuan, Shuaihang and Tian, Yu and Song, Luo and Kouhana, Ava and Elze, Tobias and others},
  booktitle={CVPR},
  year={2024},
}

@article{siragusa2025medpix20,
      title={MedPix 2.0: A Comprehensive Multimodal Biomedical Data set for Advanced AI Applications}, 
      author={Irene Siragusa and Salvatore Contino and Massimo Ciura and Rosario Alicata and Roberto Pirrone},
      year={2025},
      journal={ArXiv},
}

@InProceedings{liu2024llava15,
    author    = {Liu, Haotian and Li, Chunyuan and Li, Yuheng and Lee, Yong Jae},
    title     = {Improved Baselines with Visual Instruction Tuning},
    booktitle = {CVPR},
    year      = {2024},
}

@article{deitke2024pixmo,
      title={Molmo and PixMo: Open Weights and Open Data for State-of-the-Art Vision-Language Models}, 
      author={Matt Deitke and Christopher Clark and Sangho Lee and Rohun Tripathi and Yue Yang and Jae Sung Park and Mohammadreza Salehi and Niklas Muennighoff and Kyle Lo and Luca Soldaini and Jiasen Lu and Taira Anderson and Erin Bransom and Kiana Ehsani and Huong Ngo and YenSung Chen and Ajay Patel and Mark Yatskar and Chris Callison-Burch and Andrew Head and Rose Hendrix and Favyen Bastani and Eli VanderBilt and Nathan Lambert and Yvonne Chou and Arnavi Chheda and Jenna Sparks and Sam Skjonsberg and Michael Schmitz and Aaron Sarnat and Byron Bischoff and Pete Walsh and Chris Newell and Piper Wolters and Tanmay Gupta and Kuo-Hao Zeng and Jon Borchardt and Dirk Groeneveld and Crystal Nam and Sophie Lebrecht and Caitlin Wittlif and Carissa Schoenick and Oscar Michel and Ranjay Krishna and Luca Weihs and Noah A. Smith and Hannaneh Hajishirzi and Ross Girshick and Ali Farhadi and Aniruddha Kembhavi},
      year={2024},
      journal={ArXiv},
}

@inproceedings{ben2019vqamed,
  title={Vqa-med: Overview of the medical visual question answering task at imageclef 2019},
  author={Ben Abacha, Asma and Hasan, Sadid A and Datla, Vivek V and Demner-Fushman, Dina and M{\"u}ller, Henning},
  booktitle={CLEF},
  year={2019},
}

@article{chen2024allava,
      title={ALLaVA: Harnessing GPT4V-Synthesized Data for Lite Vision-Language Models}, 
      author={Guiming Hardy Chen and Shunian Chen and Ruifei Zhang and Junying Chen and Xiangbo Wu and Zhiyi Zhang and Zhihong Chen and Jianquan Li and Xiang Wan and Benyou Wang},
      year={2024},
      journal={ArXiv},
}

@article{chowdhury2020covid19,
  author={Chowdhury, Muhammad E. H. and Rahman, Tawsifur and Khandakar, Amith and Mazhar, Rashid and Kadir, Muhammad Abdul and Mahbub, Zaid Bin and Islam, Khandakar Reajul and Khan, Muhammad Salman and Iqbal, Atif and Emadi, Nasser Al and Reaz, Mamun Bin Ibne and Islam, Mohammad Tariqul},
  journal={IEEE Access}, 
  title={Can AI Help in Screening Viral and COVID-19 Pneumonia?}, 
  year={2020},
}

@article{openai2024gpt4o,
      title={GPT-4o System Card}, 
      author={OpenAI},
      year={2024},
      journal={ArXiv},
}

@article{gemini2025gemini,
      title={Gemini: A Family of Highly Capable Multimodal Models}, 
      author={Google DeepMind},
      year={2025},
      journal={ArXiv},
}

@article{yan2023multimodal,
      title={Multimodal ChatGPT for Medical Applications: an Experimental Study of GPT4V}, 
      author={Zhiling Yan and Kai Zhang and Rong Zhou and Lifang He and Xiang Li and Lichao Sun},
      year={2023},
      journal={ArXiv},
}

@article{lee2023benefits,
  author = {Peter Lee  and Sebastien Bubeck  and Joseph Petro },
  title = {Benefits, Limits, and Risks of GPT-4 as an AI Chatbot for Medicine},
  journal = {NEJM},
  volume = {388},
  year = {2023},
}

@article{shao2024deepseekmath,
      title={DeepSeekMath: Pushing the Limits of Mathematical Reasoning in Open Language Models}, 
      author={Zhihong Shao and Peiyi Wang and Qihao Zhu and Runxin Xu and Junxiao Song and Xiao Bi and Haowei Zhang and Mingchuan Zhang and Y. K. Li and Y. Wu and Daya Guo},
      year={2024},
      journal={ArXiv},
}

@inproceedings{wang2017chestx,
  title={Chestx-ray8: Hospital-scale chest x-ray database and benchmarks on weakly-supervised classification and localization of common thorax diseases},
  author={Wang, Xiaosong and Peng, Yifan and Lu, Le and Lu, Zhiyong and Bagheri, Mohammadhadi and Summers, Ronald M},
  booktitle={CVPR},
  year={2017},
}

@inproceedings{groh2021evaluating,
  title={Evaluating deep neural networks trained on clinical images in dermatology with the fitzpatrick 17k dataset},
  author={Groh, Matthew and Harris, Caleb and Soenksen, Luis and Lau, Felix and Han, Rachel and Kim, Aerin and Koochek, Arash and Badri, Omar},
  booktitle={CVPR},
  year={2021},
}

@article{vitale2020improving,
  title={Improving realism in patient-specific abdominal ultrasound simulation using CycleGANs},
  author={Vitale, Santiago and Orlando, José Ignacio and Iarussi, Emmanuel and Larrabide, Ignacio},
  journal={IJCARS},
  volume={15},
  year={2020},
}

@article{ben2019medquad,
  title={A question-entailment approach to question answering},
  author={Ben Abacha, Asma and Demner-Fushman, Dina},
  journal={BMC Bioinformatics},
  volume={20},
  year={2019},
}

@article{mullappilly2024bimedix2,
      title={BiMediX2: Bio-Medical EXpert LMM for Diverse Medical Modalities}, 
      author={Sahal Shaji Mullappilly and Mohammed Irfan Kurpath and Sara Pieri and Saeed Yahya Alseiari and Shanavas Cholakkal and Khaled Aldahmani and Fahad Khan and Rao Anwer and Salman Khan and Timothy Baldwin and Hisham Cholakkal},
      year={2024},
      journal={ArXiv},
}

@article{sun2025reasonmed,
      title={ReasonMed: A 370K Multi-Agent Generated Dataset for Advancing Medical Reasoning}, 
      author={Yu Sun and Xingyu Qian and Weiwen Xu and Hao Zhang and Chenghao Xiao and Long Li and Yu Rong and Wenbing Huang and Qifeng Bai and Tingyang Xu},
      year={2025},
      volume={abs/2506.09513},
      journal={ArXiv},
}

@article{wang2024apollo,
   title={Apollo: Lightweight Multilingual Medical LLMs towards Democratizing Medical AI to 6B People},
   author={Xidong Wang and Nuo Chen and Junyin Chen and Yan Hu and Yidong Wang and Xiangbo Wu and Anningzhe Gao and Xiang Wan and Haizhou Li and Benyou Wang},
   year={2024},
   journal={ArXiv},
}

@article{li2023chatdoctor,
  title={ChatDoctor: A Medical Chat Model Fine-Tuned on a Large Language Model Meta-AI (LLaMA) Using Medical Domain Knowledge},
  author={Li, Yunxiang and Li, Zihan and Zhang, Kai and Dan, Ruilong and Jiang, Steve and Zhang, You},
  journal={Cureus},
  volume={15},
  year={2023},
}

@article{sellergren2025medgemma,
      title={MedGemma Technical Report}, 
      author={Andrew Sellergren and Sahar Kazemzadeh and Tiam Jaroensri and Atilla Kiraly and Madeleine Traverse and Timo Kohlberger and Shawn Xu and Fayaz Jamil and Cían Hughes and Charles Lau and Justin Chen and Fereshteh Mahvar and Liron Yatziv and Tiffany Chen and Bram Sterling and Stefanie Anna Baby and Susanna Maria Baby and Jeremy Lai and Samuel Schmidgall and Lu Yang and Kejia Chen and Per Bjornsson and Shashir Reddy and Ryan Brush and Kenneth Philbrick and Mercy Asiedu and Ines Mezerreg and Howard Hu and Howard Yang and Richa Tiwari and Sunny Jansen and Preeti Singh and Yun Liu and Shekoofeh Azizi and Aishwarya Kamath and Johan Ferret and Shreya Pathak and Nino Vieillard and Ramona Merhej and Sarah Perrin and Tatiana Matejovicova and Alexandre Ramé and Morgane Riviere and Louis Rouillard and Thomas Mesnard and Geoffrey Cideron and Jean-bastien Grill and Sabela Ramos and Edouard Yvinec and Michelle Casbon and Elena Buchatskaya and Jean-Baptiste Alayrac and Dmitry Lepikhin and Vlad Feinberg and Sebastian Borgeaud and Alek Andreev and Cassidy Hardin and Robert Dadashi and Léonard Hussenot and Armand Joulin and Olivier Bachem and Yossi Matias and Katherine Chou and Avinatan Hassidim and Kavi Goel and Clement Farabet and Joelle Barral and Tris Warkentin and Jonathon Shlens and David Fleet and Victor Cotruta and Omar Sanseviero and Gus Martins and Phoebe Kirk and Anand Rao and Shravya Shetty and David F. Steiner and Can Kirmizibayrak and Rory Pilgrim and Daniel Golden and Lin Yang},
      year={2025},
      volume={abs/2507.05201},
      journal={ArXiv},
}

@article{he2024gsco,
      title={GSCo: Towards Generalizable AI in Medicine via Generalist-Specialist Collaboration}, 
      author={Sunan He and Yuxiang Nie and Hongmei Wang and Shu Yang and Yihui Wang and Zhiyuan Cai and Zhixuan Chen and Yingxue Xu and Luyang Luo and Huiling Xiang and Xi Lin and Mingxiang Wu and Yifan Peng and George Shih and Ziyang Xu and Xian Wu and Qiong Wang and Ronald Cheong Kin Chan and Varut Vardhanabhuti and Winnie Chiu Wing Chu and Yefeng Zheng and Pranav Rajpurkar and Kang Zhang and Hao Chen},
      year={2024},
      journal={ArXiv},
}

@article{kovalyk2022papila,
  title={PAPILA: Dataset with fundus images and clinical data of both eyes of the same patient for glaucoma assessment},
  author={Kovalyk, Oleksandr and Morales-S{\'a}nchez, Juan and Verd{\'u}-Monedero, Rafael and Sell{\'e}s-Navarro, Inmaculada and Palaz{\'o}n-Cabanes, Ana and Sancho-G{\'o}mez, Jos{\'e}-Luis},
  journal={Scientific Data},
  volume={9},
  year={2022},
}

@article{javadi2019novel,
  title={A novel deep learning method for automatic assessment of human sperm images},
  author={Javadi, Soroush and Mirroshandel, Seyed Abolghasem},
  journal={Computers in Biology and Medicine},
  year={2019},
}

@article{RadImageNet,
author = {Mei, Xueyan and Liu, Zelong and Robson, Philip M. and Marinelli, Brett and Huang, Mingqian and Doshi, Amish and Jacobi, Adam and Cao, Chendi and Link, Katherine E. and Yang, Thomas and Wang, Ying and Greenspan, Hayit and Deyer, Timothy and Fayad, Zahi A. and Yang, Yang},
title = {RadImageNet: An Open Radiologic Deep Learning Research Dataset for Effective Transfer Learning},
journal = {Radiology: Artificial Intelligence},
volume = {0},
year = {2022},
}

@article{DBLP:journals/corr/llama3,
  author       = {Abhimanyu Dubey and
                  Abhinav Jauhri and
                  Abhinav Pandey and
                  Abhishek Kadian and
                  Ahmad Al{-}Dahle and
                  Aiesha Letman and
                  Akhil Mathur and
                  Alan Schelten and
                  Amy Yang and
                  Angela Fan and
                  Anirudh Goyal and
                  Anthony Hartshorn and
                  Aobo Yang and
                  Archi Mitra and
                  Archie Sravankumar and
                  Artem Korenev and
                  Arthur Hinsvark and
                  Arun Rao and
                  Aston Zhang and
                  Aur{\'{e}}lien Rodriguez and
                  Austen Gregerson and
                  Ava Spataru and
                  Baptiste Rozi{\`{e}}re and
                  Bethany Biron and
                  Binh Tang and
                  Bobbie Chern and
                  Charlotte Caucheteux and
                  Chaya Nayak and
                  Chloe Bi and
                  Chris Marra and
                  Chris McConnell and
                  Christian Keller and
                  Christophe Touret and
                  Chunyang Wu and
                  Corinne Wong and
                  Cristian Canton Ferrer and
                  Cyrus Nikolaidis and
                  Damien Allonsius and
                  Daniel Song and
                  Danielle Pintz and
                  Danny Livshits and
                  David Esiobu and
                  Dhruv Choudhary and
                  Dhruv Mahajan and
                  Diego Garcia{-}Olano and
                  Diego Perino and
                  Dieuwke Hupkes and
                  Egor Lakomkin and
                  Ehab AlBadawy and
                  Elina Lobanova and
                  Emily Dinan and
                  Eric Michael Smith and
                  Filip Radenovic and
                  Frank Zhang and
                  Gabriel Synnaeve and
                  Gabrielle Lee and
                  Georgia Lewis Anderson and
                  Graeme Nail and
                  Gr{\'{e}}goire Mialon and
                  Guan Pang and
                  Guillem Cucurell and
                  Hailey Nguyen and
                  Hannah Korevaar and
                  Hu Xu and
                  Hugo Touvron and
                  Iliyan Zarov and
                  Imanol Arrieta Ibarra and
                  Isabel M. Kloumann and
                  Ishan Misra and
                  Ivan Evtimov and
                  Jade Copet and
                  Jaewon Lee and
                  Jan Geffert and
                  Jana Vranes and
                  Jason Park and
                  Jay Mahadeokar and
                  Jeet Shah and
                  Jelmer van der Linde and
                  Jennifer Billock and
                  Jenny Hong and
                  Jenya Lee and
                  Jeremy Fu and
                  Jianfeng Chi and
                  Jianyu Huang and
                  Jiawen Liu and
                  Jie Wang and
                  Jiecao Yu and
                  Joanna Bitton and
                  Joe Spisak and
                  Jongsoo Park and
                  Joseph Rocca and
                  Joshua Johnstun and
                  Joshua Saxe and
                  Junteng Jia and
                  Kalyan Vasuden Alwala and
                  Kartikeya Upasani and
                  Kate Plawiak and
                  Ke Li and
                  Kenneth Heafield and
                  Kevin Stone and
                  et al.},
  title        = {The Llama 3 Herd of Models},
  year         = {2024},
  journal    = {ArXiv},
}

@article{aydin2025openai,
      title={OpenAI ChatGPT interprets Radiological Images: GPT-4 as a Medical Doctor for a Fast Check-Up}, 
      author={Omer Aydin and Enis Karaarslan},
      year={2025},
      journal={ArXiv},
}

@article{arora2025healthbench,
      title={HealthBench: Evaluating Large Language Models Towards Improved Human Health}, 
      author={Rahul K. Arora and Jason Wei and Rebecca Soskin Hicks and Preston Bowman and Joaquin Quiñonero-Candela and Foivos Tsimpourlas and Michael Sharman and Meghan Shah and Andrea Vallone and Alex Beutel and Johannes Heidecke and Karan Singhal},
      year={2025},
      journal={ArXiv},
}

@article{yang2024advancing,
      title={Advancing Multimodal Medical Capabilities of Gemini}, 
      author={Lin Yang and Shawn Xu and Andrew Sellergren and Timo Kohlberger and Yuchen Zhou and Ira Ktena and Atilla Kiraly and Faruk Ahmed and Farhad Hormozdiari and Tiam Jaroensri and Eric Wang and Ellery Wulczyn and Fayaz Jamil and Theo Guidroz and Chuck Lau and Siyuan Qiao and Yun Liu and Akshay Goel and Kendall Park and Arnav Agharwal and Nick George and Yang Wang and Ryutaro Tanno and David G. T. Barrett and Wei-Hung Weng and S. Sara Mahdavi and Khaled Saab and Tao Tu and Sreenivasa Raju Kalidindi and Mozziyar Etemadi and Jorge Cuadros and Gregory Sorensen and Yossi Matias and Katherine Chou and Greg Corrado and Joelle Barral and Shravya Shetty and David Fleet and S. M. Ali Eslami and Daniel Tse and Shruthi Prabhakara and Cory McLean and Dave Steiner and Rory Pilgrim and Christopher Kelly and Shekoofeh Azizi and Daniel Golden},
      year={2024},
      journal={ArXiv},
}

@article{singhal2023large,
      title={Large language models encode clinical knowledge}, 
      author={Singhal, Karan and Azizi, Shekoofeh and Tu, Tao and Mahdavi, S. Sara and Wei, Jason and Chung, Hyung Won and Scales, Nathan and Tanwani, Ajay and Cole-Lewis, Heather and Pfohl, Stephen and Payne, Perry and Seneviratne, Martin and Gamble, Paul and Kelly, Chris and Babiker, Abubakr and Schärli, Nathanael and Chowdhery, Aakanksha and Mansfield, Philip and Demner-Fushman, Dina and Agüera y Arcas, Blaise and Webster, Dale and Corrado, Greg S. and Matias, Yossi and Chou, Katherine and Gottweis, Juraj and Tomasev, Nenad and Liu, Yun and Rajkomar, Alvin and Barral, Joelle and Semturs, Christopher and Karthikesalingam, Alan and Natarajan, Vivek},
      year={2023},
      journal={Nature},
}

@article{li2025agenthospital,
      title={Agent Hospital: A Simulacrum of Hospital with Evolvable Medical Agents}, 
      author={Junkai Li and Yunghwei Lai and Weitao Li and Jingyi Ren and Meng Zhang and Xinhui Kang and Siyu Wang and Peng Li and Ya-Qin Zhang and Weizhi Ma and Yang Liu},
      year={2025},
      journal={ArXiv},
}

@article{schmidgall2025agentclinic,
      title={AgentClinic: a multimodal agent benchmark to evaluate AI in simulated clinical environments}, 
      author={Samuel Schmidgall and Rojin Ziaei and Carl Harris and Eduardo Reis and Jeffrey Jopling and Michael Moor},
      year={2025},
      journal={ArXiv},
}

@article{wang2025medagentpro,
      title={MedAgent-Pro: Towards Evidence-based Multi-modal Medical Diagnosis via Reasoning Agentic Workflow}, 
      author={Ziyue Wang and Junde Wu and Linghan Cai and Chang Han Low and Xihong Yang and Qiaxuan Li and Yueming Jin},
      year={2025},
      journal={ArXiv},
}

@article{keyl2024decoding,
      title={Decoding pan-cancer treatment outcomes using multimodal real-world data and explainable artificial intelligence}, 
      author={Keyl, Julius and Keyl, Philipp and Montavon, Grégoire and Hosch, René and Brehmer, Alexander and Mochmann, Liliana and Jurmeister, Philipp and Dernbach, Gabriel and Kim, Moon and Koitka, Sven and Bauer, Sebastian and Bechrakis, Nikolaos and Forsting, Michael and Führer-Sakel, Dagmar and Glas, Martin and Grünwald, Viktor and Hadaschik, Boris and Haubold, Johannes and Herrmann, Ken and Kasper, Stefan and Kimmig, Rainer and Lang, Stephan and Rassaf, Tienush and Roesch, Alexander and Schadendorf, Dirk and Siveke, Jens T. and Stuschke, Martin and Sure, Ulrich and Totzeck, Matthias and Welt, Anja and Wiesweg, Marcel and Baba, Hideo A. and Nensa, Felix and Egger, Jan and Müller, Klaus-Robert and Schuler, Martin and Klauschen, Frederick and Kleesiek, Jens},
      year={2025},
      volume={6},
      journal={Nature Cancer},
}

@article{alSaad2024multimodal,
author="AlSaad, Rawan
and Abd-alrazaq, Alaa
and Boughorbel, Sabri
and Ahmed, Arfan
and Renault, Max-Antoine
and Damseh, Rafat
and Sheikh, Javaid",
title="Multimodal Large Language Models in Health Care: Applications, Challenges, and Future Outlook",
journal="J Med Internet Res",
year="2024",
volume="26",
}

@article{tian2023role,
  title={The role of large language models in medical image processing: a narrative review},
  author={Tian, Dianzhe and Jiang, Shitao and Zhang, Lei and Lu, Xin and Xu, Yiyao},
  journal={Quantitative Imaging in Medicine and Surgery},
  volume={14},
  year={2023},
}

@article{pai2024cancerfm,
      title={Foundation model for cancer imaging biomarkers}, 
      author={Pai, Suraj and Bontempi, Dennis and Hadzic, Ibrahim and Prudente, Vasco and Sokač, Mateo and Chaunzwa, Tafadzwa and Bernatz, Simon and Hosny, Ahmed and Mak, Raymond and Birkbak, Nicolai and Aerts, Hugo},
      year={2024},
      journal={Nature Machine Intelligence},
}

@article{sammut2022multi,
      title={Multi-omic machine learning predictor of breast cancer therapy response}, 
      author={Sammut, Stephen-John and Crispin-Ortuzar, Mireia and Chin, Suet-Feung and Provenzano, Elena and Bardwell, Helen A. and Ma, Wenxin and Cope, Wei and Dariush, Ali and Dawson, Sarah-Jane and Abraham, Jean E. and Dunn, Janet and Hiller, Louise and Thomas, Jeremy and Cameron, David A. and Bartlett, John M. S. and Hayward, Larry and Pharoah, Paul D. and Markowetz, Florian and Rueda, Oscar M. and Earl, Helena M. and Caldas, Carlos},
      year={2022},
      volume={601},
      journal={Nature},
}

@article{deng2024ophglm,
author = {Deng, Zhuo and Gao, Weihao and Chen, Chucheng and Niu, Zhiyuan and Gong, Zheng and Zhang, Ruiheng and Cao, Zhenjie and Li, Fang and Ma, Zhaoyi and Wei, Wenbin and Ma, Lan},
title = {OphGLM: An ophthalmology large language-and-vision assistant},
year = {2024},
volume = {157},
journal = {Artif. Intell. Med.},
}

@article{wang2024chatcad,
      title={Interactive computer-aided diagnosis on medical image using large language models}, 
      author={Wang, Sheng and Zhao, Zihao and Ouyang, Xi and Liu, Tianming and Wang, Qian and Shen, Dinggang},
      year={2024},
      journal={Communications Engineering},
}

@article{zhao2024biomedparse,
   title={A foundation model for joint segmentation, detection and recognition of biomedical objects across nine modalities},
   volume={22},
   journal={Nature Methods},
   author={Zhao, Theodore and Gu, Yu and Yang, Jianwei and Usuyama, Naoto and Lee, Ho Hin and Kiblawi, Sid and Naumann, Tristan and Gao, Jianfeng and Crabtree, Angela and Abel, Jacob and Moung-Wen, Christine and Piening, Brian and Bifulco, Carlo and Wei, Mu and Poon, Hoifung and Wang, Sheng},
   year={2024},
}

@article{tanno2024flamingocxr,
      title={Collaboration between clinicians and vision–language models in radiology report generation}, 
      author={Tanno, Ryutaro and Barrett, David G. T. and Sellergren, Andrew and Ghaisas, Sumedh and Dathathri, Sumanth and See, Abigail and Welbl, Johannes and Lau, Charles and Tu, Tao and Azizi, Shekoofeh and Singhal, Karan and Schaekermann, Mike and May, Rhys and Lee, Roy and Man, SiWai and Mahdavi, Sara and Ahmed, Zahra and Matias, Yossi and Barral, Joelle and Eslami, S. M. Ali and Belgrave, Danielle and Liu, Yun and Kalidindi, Sreenivasa Raju and Shetty, Shravya and Natarajan, Vivek and Kohli, Pushmeet and Huang, Po-Sen and Karthikesalingam, Alan and Ktena, Ira},
      year={2025},
      volume={31},
      journal={Nature Medicine},
}

@article{niu2025multimodal,
      title={Medical multimodal multitask foundation model for lung cancer screening}, 
      author={Niu, Chuang and Lyu, Qing and Carothers, Christopher D. and Kaviani, Parisa and Tan, Josh and Yan, Pingkun and Kalra, Mannudeep K. and Whitlow, Christopher T. and Wang, Ge},
      year={2025},
      volume={16},
      journal={Nature Communications},
}

@article{PACHECO2020106221,
title = {PAD-UFES-20: A skin lesion dataset composed of patient data and clinical images collected from smartphones},
journal = {Data in Brief},
volume = {32},
year = {2020},
author = {Andre G.C. Pacheco and Gustavo R. Lima and Amanda S. Salomão and Breno Krohling and Igor P. Biral and Gabriel G. {de Angelo} and Fábio C.R. {Alves Jr} and José G.M. Esgario and Alana C. Simora and Pedro B.C. Castro and Felipe B. Rodrigues and Patricia H.L. Frasson and Renato A. Krohling and Helder Knidel and Maria C.S. Santos and Rachel B. {do Espírito Santo} and Telma L.S.G. Macedo and Tania R.P. Canuto and Luíz F.S. {de Barros}},
}

@inproceedings{irvin2019chexpert,
  title={Chexpert: A large chest radiograph dataset with uncertainty labels and expert comparison},
  author={Irvin, Jeremy and Rajpurkar, Pranav and Ko, Michael and Yu, Yifan and Ciurea-Ilcus, Silviana and Chute, Chris and Marklund, Henrik and Haghgoo, Behzad and Ball, Robyn and Shpanskaya, Katie and others},
  booktitle={AAAI},
  volume={33},
  year={2019},
}

@article{rajpurkar2017chexnet,
  title={Chexnet: Radiologist-level pneumonia detection on chest x-rays with deep learning},
  author={Rajpurkar, Pranav and Irvin, Jeremy and Zhu, Kaylie and Yang, Brandon and Mehta, Hershel and Duan, Tony and Ding, Daisy and Bagul, Aarti and Langlotz, Curtis and Shpanskaya, Katie and others},
  journal={ArXiv},
  year={2017},
}

@article{HE2021102055,
title = {Meta grayscale adaptive network for 3D integrated renal structures segmentation},
journal = {Medical Image Analysis},
volume = {71},
year = {2021},
author = {Yuting He and Guanyu Yang and Jian Yang and Rongjun Ge and Youyong Kong and Xiaomei Zhu and Shaobo Zhang and Pengfei Shao and Huazhong Shu and Jean-Louis Dillenseger and Jean-Louis Coatrieux and Shuo Li},
}

@article{yan2017deeplesion,
  title={Deeplesion: Automated deep mining, categorization and detection of significant radiology image findings using large-scale clinical lesion annotations},
  author={Yan, Ke and Wang, Xiaosong and Lu, Le and Summers, Ronald M},
  journal={ArXiv},
  year={2017},
}

@article{de20242024,
  title={The 2024 Brain Tumor Segmentation (BraTS) challenge: glioma segmentation on post-treatment MRI},
  author={de Verdier, Maria Correia and Saluja, Rachit and Gagnon, Louis and LaBella, Dominic and Baid, Ujjwall and Tahon, Nourel Hoda and Foltyn-Dumitru, Martha and Zhang, Jikai and Alafif, Maram and Baig, Saif and others},
  journal={ArXiv},
  year={2024}
}

@article{lou2025sdr,
  title={SDR-Former: A Siamese Dual-Resolution Transformer for Liver Lesion Classification Using 3D Multi-Phase Imaging},
  author={Lou, Meng and Ying, Hanning and Liu, Xiaoqing and Zhou, Hong-Yu and Zhang, Yuqin and Yu, Yizhou},
  journal={Neural Networks},
  year={2025},
}

@article{garrucho2025,
  title={A large-scale multicenter breast cancer DCE-MRI benchmark dataset with expert segmentations},
  author={Garrucho, Lidia and Kushibar, Kaisar and Reidel, Claire-Anne and Joshi, Smriti and Osuala, Richard and Tsirikoglou, Apostolia and Bobowicz, Maciej and Riego, Javier del and Catanese, Alessandro and Gwoździewicz, Katarzyna and Cosaka, Maria-Laura and Abo-Elhoda, Pasant M and Tantawy, Sara W and Sakrana, Shorouq S and Shawky-Abdelfatah, Norhan O and Salem, Amr Muhammad Abdo and Kozana, Androniki and Divjak, Eugen and Ivanac, Gordana and Nikiforaki, Katerina and Klontzas, Michail E and García-Dosdá, Rosa and Gulsun-Akpinar, Meltem and Lafcı, Oğuz and Mann, Ritse and Martín-Isla, Carlos and Prior, Fred and Marias, Kostas and Starmans, Martijn P A and Strand, Fredrik and Díaz, Oliver and Igual, Laura and Lekadir, Karim},
  journal = {Scientific Data},
  year = {2025},
}

@article{wagner2023semantic,
  title={Semantic modeling of cell damage prediction: a machine learning approach at human-level performance in dermatology},
  author={Wagner, Patrick and Springenberg, Maximilian and Kr{\"o}ger, Marius and Moritz, Rose KC and Schleusener, Johannes and Meinke, Martina C and Ma, Jackie},
  journal={Scientific Reports},
  volume={13},
  year={2023},
}

@article{gamper2020pannuke,
  title={Pannuke dataset extension, insights and baselines},
  author={Gamper, Jevgenij and Koohbanani, Navid Alemi and Benes, Ksenija and Graham, Simon and Jahanifar, Mostafa and Khurram, Syed Ali and Azam, Ayesha and Hewitt, Katherine and Rajpoot, Nasir},
  journal={ArXiv},
  year={2020},
}

@article{shi2023ebhi,
  title={EBHI-Seg: A novel enteroscope biopsy histopathological hematoxylin and eosin image dataset for image segmentation tasks},
  author={Shi, Liyu and Li, Xiaoyan and Hu, Weiming and Chen, Haoyuan and Chen, Jing and Fan, Zizhen and Gao, Minghe and Jing, Yujie and Lu, Guotao and Ma, Deguo and others},
  journal={Frontiers in Medicine},
  volume={10},
  year={2023},
}

@article{ding2023large,
  title={A large-scale synthetic pathological dataset for deep learning-enabled segmentation of breast cancer},
  author={Ding, Kexin and Zhou, Mu and Wang, He and Gevaert, Olivier and Metaxas, Dimitris and Zhang, Shaoting},
  journal={Scientific Data},
  year={2023},
}

@article{zhang2025videollama,
  title={Videollama 3: Frontier multimodal foundation models for image and video understanding},
  author={Zhang, Boqiang and Li, Kehan and Cheng, Zesen and Hu, Zhiqiang and Yuan, Yuqian and Chen, Guanzheng and Leng, Sicong and Jiang, Yuming and Zhang, Hang and Li, Xin and others},
  journal={arXiv preprint arXiv:2501.13106},
  year={2025}
}

@article{landis1977,
  author  = {Landis, J. Richard and Koch, Gary G.},
  title   = {The Measurement of Observer Agreement for Categorical Data},
  journal = {Biometrics},
  volume  = {33},
  number  = {1},
  pages   = {159--174},
  year    = {1977},
  doi     = {10.2307/2529310},
  publisher = {JSTOR}
}
